\documentclass[journal]{IEEEtran}

\usepackage[american]{babel}

\usepackage{graphicx}
\graphicspath{{figure/}{photo/}}
\usepackage[caption=false]{subfig}

\usepackage{amsmath, amssymb}
\usepackage{mathrsfs}
\usepackage[bold]{hhtensor}
\usepackage{multirow} 
\usepackage[breaklinks=true,colorlinks,bookmarks=false]{hyperref}
\newlength{\tempheight}
\newlength{\tempwidth}
\usepackage{flushend}
\newcommand{\rowname}[1]
{\rotatebox{90}{\makebox[\tempheight][c]{\textbf{#1}}}}

\newcommand{\columnname}[1]
{\makebox[\tempwidth][c]{\textbf{#1}}}

\DeclareMathOperator*{\argmax}{argmax}
\DeclareMathOperator*{\argmin}{argmin}

\begin{document}
\title{Deep Label Distribution Learning \\With Label Ambiguity}
\author{Bin-Bin Gao, Chao Xing, Chen-Wei Xie, Jianxin Wu,~\IEEEmembership{Member,~IEEE,} and Xin Geng,~\IEEEmembership{Member,~IEEE}
\thanks{This work was supported in part
by the National Natural Science Foundation of China under Grant 61422203,
Grant 61622203, and Grant 61232007, in part by the Jiangsu Natural Science 
Funds for Distinguished Young Scholar under Grant BK20140022, in
part by the Collaborative Innovation Center of Novel Software Technology
and Industrialization, and in part by the Collaborative Innovation Center
of Wireless Communications Technology. (\emph{Corresponding Author: Jianxin Wu.})}
\thanks{B.-B. Gao, C.-W. Xie and J. Wu are with the National Key Laboratory for Novel Software Technology, Nanjing University, Nanjing 210023, China (e-mail: gaobb@lamda.nju.edu.cn; xiecw@lamda.nju.edu.cn; wujx@lamda.
nju.edu.cn).}
\thanks{C. Xing and X. Geng are with the MOE Key Laboratory of Computer
Network and Information Integration, School of Computer Science
and Engineering, Southeast University, Nanjing 211189, China (e-mail:
xingchao@seu.edu.cn; xgeng@seu.edu.cn).}
}

\markboth{ACCEPTED BY IEEE TIP}{ACCEPTED BY IEEE TIP}

\maketitle

\begin{abstract}
Convolutional Neural Networks (ConvNets) have achieved excellent recognition performance in various visual recognition tasks. A large labeled training set is one of the most important factors for its success. However, it is difficult to collect sufficient training images with precise labels in some domains such as apparent age estimation, head pose estimation, multi-label classification and semantic segmentation. Fortunately, there is ambiguous information among labels, which makes these tasks different from traditional classification. Based on this observation, we convert the label of each image into a discrete label distribution, and learn the label distribution by minimizing a Kullback-Leibler divergence between the predicted and ground-truth label distributions using deep ConvNets. The proposed DLDL (Deep Label Distribution Learning) method effectively utilizes the label ambiguity in both feature learning and classifier learning, which help prevent the network from over-fitting even when the training set is small. Experimental results show that the proposed approach produces significantly better results than state-of-the-art methods for age estimation and head pose estimation. At the same time, it also improves recognition performance for multi-label classification and semantic segmentation tasks.
\end{abstract}

\IEEEpeerreviewmaketitle

\section{Introduction}
\IEEEPARstart{C}{onvolutional} Neural Networks~(ConvNets) have achieved state-of-the-art performance on various visual recognition tasks such as image classification~\cite{krizhevsky2012imagenet}, object detection~\cite{girshick2014rich} and semantic segmentation~\cite{long2015fully}. The availability of a huge set of training images is one of the most important factors for their success. However, it is difficult to collect sufficient training images with unambiguous labels in domains such as age estimation~\cite{geng2013facial}, head pose estimation~\cite{kong2015head}, multi-label classification and semantic segmentation. Therefore, exploiting deep learning methods with limited samples and ambiguous labels has become an attractive yet challenging topic.

Why is it difficult to collect a large and accurately labeled training set? Firstly, it is difficult (even for domain experts) to provide exact labels to some tasks. For example, the pixels close to object boundaries are very difficult to label for annotators in semantic segmentation. In addition, pixel labeling is a time-consuming task that may limit the amount of training samples. Another example is that people's apparent age and head pose is difficult to describe with an accurate number. Secondly, it is very hard to gather complete and sufficient data. For example, it is difficult to build an age dataset covering people from 1 to 85 years old, and ensure that every age in this range has enough associated images. Similar difficulties arise in head pose estimation, where head poses are usually collected at a small set of angles with a 10$^\circ$ or 15$^\circ$ increment. Thus, the publicly available age, head pose and semantic segmentation datasets are small scale compared to those in image classification tasks.

\begin{figure*}
  \centering
  \subfloat[Age estimation]{
   \begin{tabular}{c}
    \includegraphics[height=0.28\columnwidth,keepaspectratio]{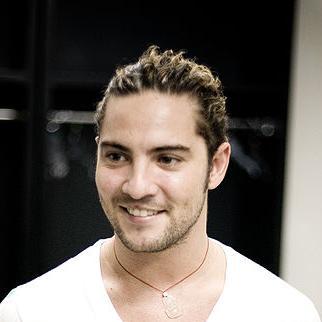} \\
    \includegraphics[height=0.32\columnwidth,keepaspectratio]{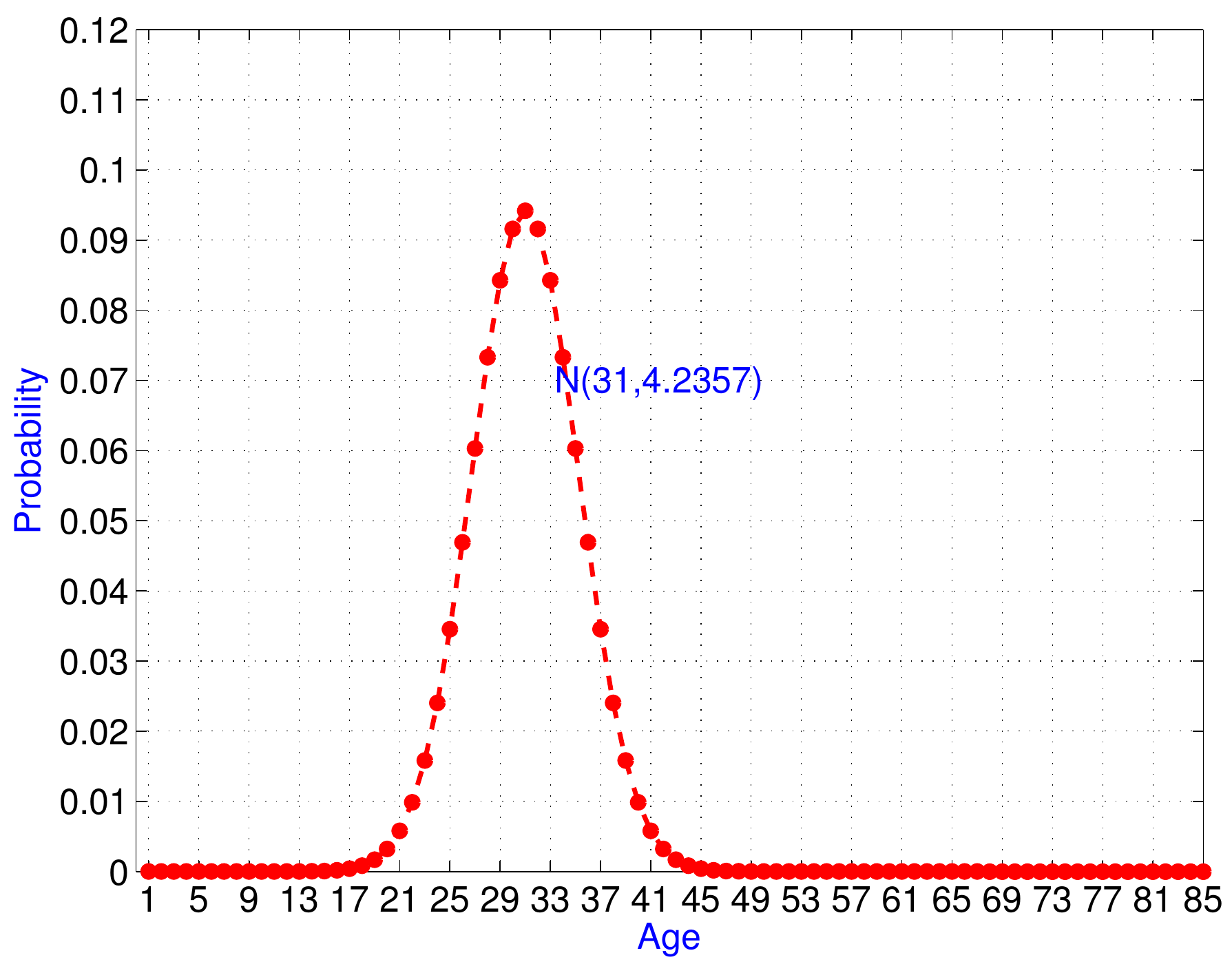} \label{fig:age-ld}
   \end{tabular}
  }
  \subfloat[Head pose estimation] {
   \begin{tabular}{c}
    \includegraphics[height=0.28\columnwidth,keepaspectratio]{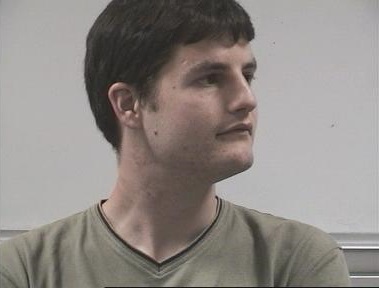} \\
    \includegraphics[height=0.32\columnwidth,keepaspectratio]{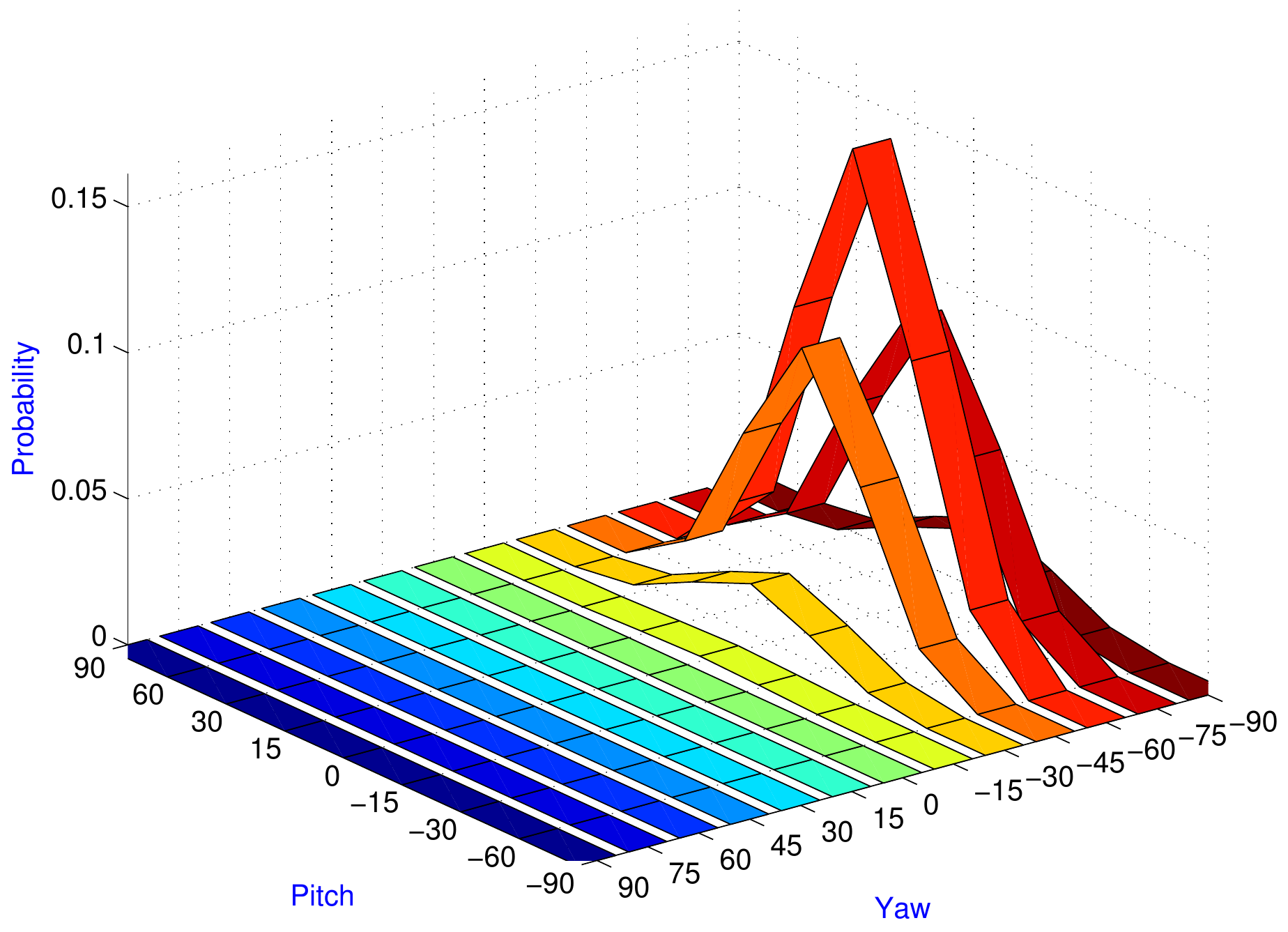} \label{fig:pose-ld}
   \end{tabular}
  }
  \subfloat[Multi-label classification] {
   \begin{tabular}{c}
    \includegraphics[height=0.28\columnwidth,keepaspectratio]{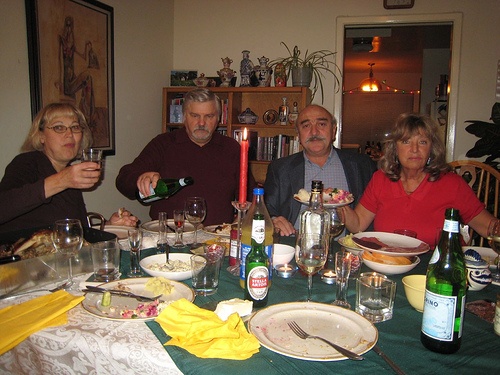} \\
    \includegraphics[height=0.32\columnwidth,keepaspectratio]{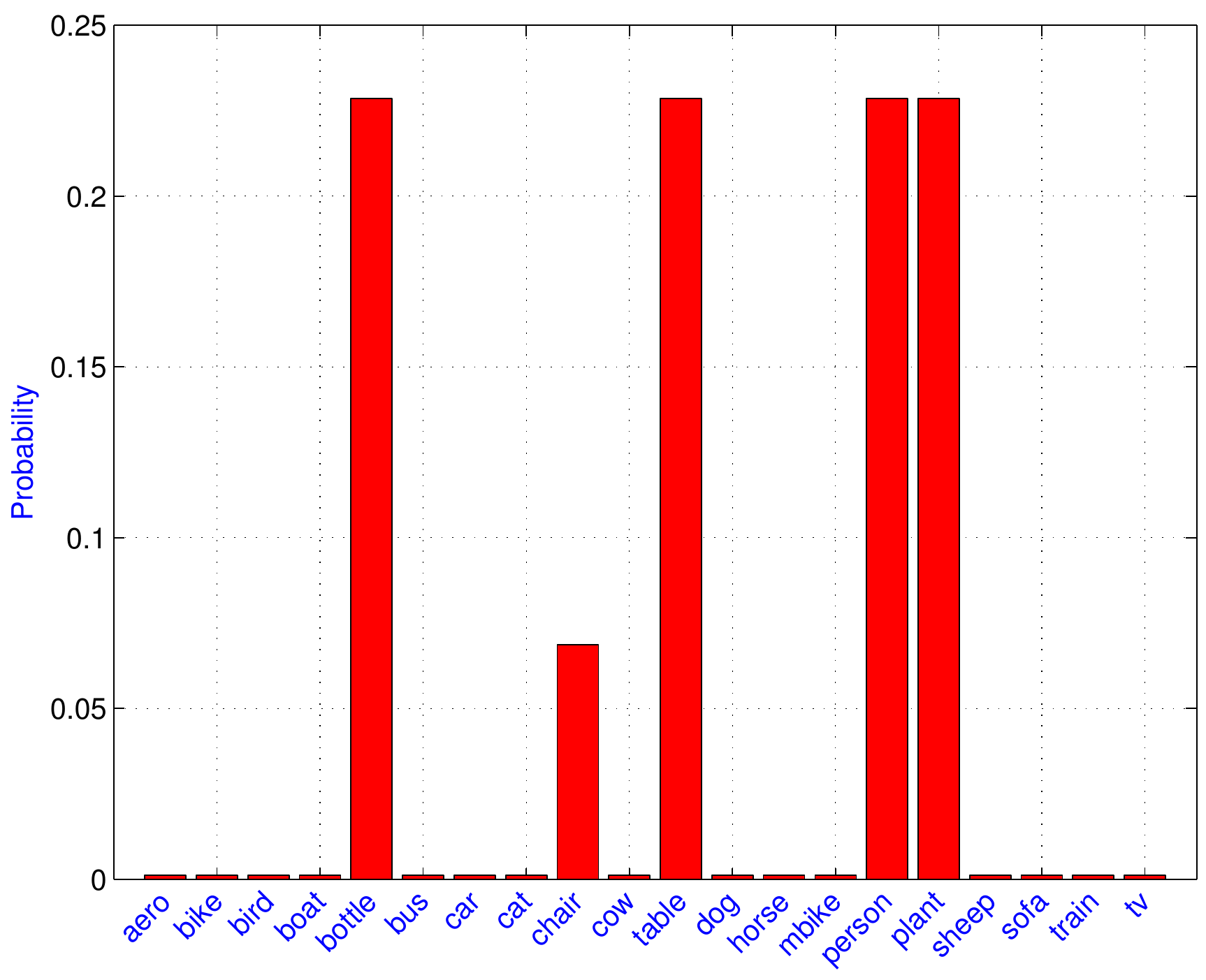} \label{fig:ml-ld}
   \end{tabular}
  }
  \subfloat[Semantic segmentation] {
   \begin{tabular}{c}
    \includegraphics[height=0.28\columnwidth,keepaspectratio]{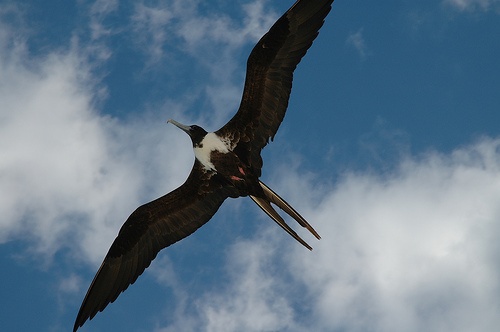} \\
    \includegraphics[height=0.32\columnwidth,keepaspectratio]{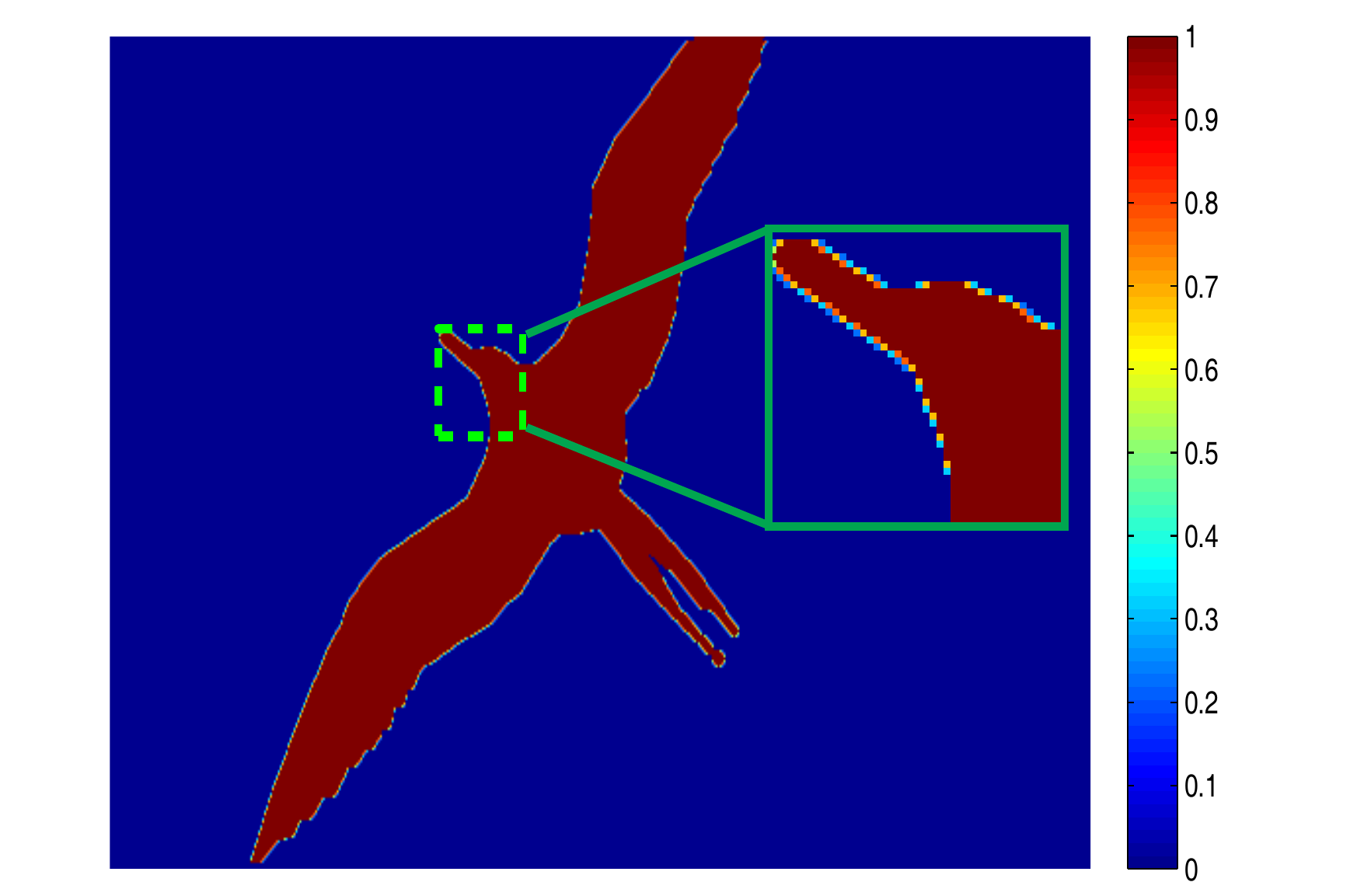} \label{fig:ss-ld}
   \end{tabular}
  }
  \caption{Different label distributions for different recognition tasks. The first row shows four images, with the first two images coming from \emph{ChaLearn 2015} and \emph{Pointing'04} and the last two images coming from the PASCAL \emph{VOC2007} classification task and the PASCAL \emph{VOC2011} segmentation challenge. The second row shows their corresponding label distributions~(best viewed in color).} \label{fig:eg-ld}
\end{figure*}

These aforementioned small datasets have a common characteristic, \emph{i.e.}, label ambiguity, which refers to the uncertainty among the ground-truth labels. On one hand, label ambiguity is unavoidable in some applications. We usually predict another person's age in a way like ``around 25", which indicates using not only 25, but also neighboring ages to describe the face. And, different people may have different guesses towards the same face. Similar situations also hold for other types of tasks. The labels of pixels at object boundaries are difficult to annotate because of the inherent ambiguity of these pixels in semantic segmentation. On the other hand, label ambiguity can also happen if we are not confident in the labels we provide for an image. In the multi-label classification task, some objects are clearly visible but difficult to recognize. This type of objects are annotated as \texttt{Difficult} in the PASCAL Visual Object Classes~(VOC) classification challenge~\cite{everingham2010pascal},~\emph{e.g.}, the chair in the third image of the first row in Fig.~\ref{fig:eg-ld}. 

There are two main types of labeling methods: single-label recognition~(SLR) and multi-label recognition~(MLR). SLR assumes one image or pixel has one label and MLR assumes that one image or pixel may be assigned multiple labels. Both SLR and MLR aim to answer the question of which labels can be used to describe an image or pixel, but they can not describe the label ambiguity associated with it. Label ambiguity will help improve recognition performance if it can be reasonably
exploited. In order to utilize label correlation~(which may be considered as a consequence of label ambiguity in some applications), Geng~\emph{et al.} proposed a label distribution learning~(LDL) approach for age estimation~\cite{geng2013facial} and head pose estimation~\cite{geng2014head}. Recently, some improvements of  LDL have been proposed. Xing~\emph{et al.} proposed two algorithms named LDLogitBoost and AOSO-LDLogitBoost to learn general models to relax the maximum entropy model in traditional LDL methods~\cite{xing2016logistic}. Furthermore, He~\emph{et al.} generated age label distributions through weighted linear combination of the input image's label and its context-neighboring samples~\cite{he35data}. However, these methods are suboptimal because they only utilize the correlation of neighboring labels in classifier learning, but not in learning the visual representations.

Deep ConvNets have natural advantages in feature learning. Existing ConvNet frameworks can be viewed as classification and regression models based on different optimization objective functions. In many cases, the softmax loss and $\ell_2$ loss are used in deep ConvNet models for classification~\cite{he2015deep} and regression problems~\cite{belagiannis2015robust}, respectively. The softmax loss maximizes the estimated probability of the ground-truth class without considering other classes, and the $\ell_2$ loss minimizes the squared difference between the estimated values of the network and the ground-truth. These methods have achieved satisfactory performance in some domains such as image classification, human pose estimation and object detection. However, existing deep learning methods cannot utilize the label ambiguity information. Moreover, a well-known fact is that learning a good ConvNet requires a lot of images.

In order to solve the issues mentioned above, we convert both traditional SLR and MLR problems to \emph{label distribution learning} problems. Every instance is assigned a discrete label distribution $\vec y$ according to its ground-truth. The label distribution can naturally describe the ambiguous information among all possible labels. Through deep label distribution learning, the training instances associated with each class label is significantly increased without actually increase the number of the total training examples. Fig.~\ref{fig:eg-ld} intuitively shows four examples of label distribution for different recognition tasks. Then, we utilize a deep ConvNet to learn the label distribution in both feature learning and classifier learning. Since we learn label distribution with deep ConvNets, we call our method DLDL: Deep Label Distribution Learning. The benefits of DLDL are summarized as follows:
\begin{itemize}
\item DLDL is an end-to-end learning framework which utilizes the label ambiguity in both feature learning and classifier learning;
\item DLDL not only achieves more robust performance than existing classification and regression methods, but also effectively relaxes the requirement for large amount of training images, \emph{e.g.}, a training face image with ground-truth label 25 is also useful for predicting faces at age 24 or 26;
\item DLDL~(only single model without ensemble) achieves better performance than the state-of-the-art methods on age and head pose estimation tasks. DLDL also improves the performance for multi-label classification and semantic segmentation.
\end{itemize}

The rest of this paper is organized as follows. We first review the related work in Section~\ref{sec:rw}. Then, Section~\ref{sec:pa} proposes the DLDL framework, including the DLDL problem definition, DLDL theory, label distribution construction and training details. After that, the experiments are reported in Section~\ref{sec:ex}. Finally, Section~\ref{sec:dis} presents discussions and the conclusion is given in Section~\ref{sec:co}.

\section{Related Work}\label{sec:rw}

In the past two decades, many efforts have been devoted to visual recognition, including at least image classification, object detection, semantic segmentation, and facial attribute~(apparent age and head pose) estimation. These works can be divided into two streams. Earlier research was mainly based on hand-crafted features, while more recent ones are usually deep learning methods. In this section, we briefly review these related approaches.

Methods based on hand-crafted features usually include two stages. The first stage is feature extraction. The second stage learns models for recognition, detection or estimation using these features. SVM, random forest~\cite{fanelli2011real} and neural networks have commonly been used during the learning stage. In addition, Geng~\emph{et al.} proposed the label distribution learning approach to utilize the correlation among adjacent labels, which further improved performance on age estimation~\cite{geng2013facial} and head pose estimation~\cite{geng2014head}.

Although important progresses have been made with these features, the hand-crafted features render them suboptimal for particular tasks such as age or head pose estimation. More recently, learning feature representation has shown great advantages. For example, Lu~\emph{et al.}~\cite{lu2015cost} tried to learn cost-sensitive local binary features for age estimation.

Deep learning has substantially improved upon the state-of-the-art in image classification~\cite{he2015deep}, object detection~\cite{girshick2014rich}, semantic segmentation~\cite{long2015fully} and many other vision tasks. In many cases, the softmax loss is used in deep models for classification~\cite{he2015deep}. Besides classification, deep ConvNets have also been trained for regression tasks such as head pose estimation~\cite{ahn2015real} and facial landmark detection~\cite{sun2013deep}. In regression problems, the training procedure usually optimizes a squared $\ell_2$ loss function. Satisfactory performance has also been obtained by using Tukey's biweight function in human pose estimation~\cite{belagiannis2015robust}. In terms of model architecture, deep ConvNet models which use deeper architecture and smaller convolution filters (\emph{e.g.}, VGG-Nets~\cite{simonyan2015very} and VGG-Face~\cite{Parkhi15}) are very powerful. Nevertheless, these deep learning methods do not make use of the presence of label ambiguity in the training set, and usually require a large amount of training data.

A latest approach, in Inception-v3~\cite{szegedy2015rethinking}, is based on label smoothing~(LS). Instead of only using the ground-truth label, they utilize a mixture of the ground-truth label and a uniform distribution to regularize the classifier. However, LS is limited to the uniform distribution among labels rather than mining labels' ambiguous information. We believe that label ambiguity is too important to ignore. If we make good use of the ambiguity, we expect the required number of training images for some tasks could be effectively reduced.

In this paper, we focus on how to exploit the label ambiguity in deep ConvNets. Age and head pose estimation from still face images are suitable applications of the proposed research. In addition, we also extend our works to multi-label classification and semantic segmentation.

\section{The Proposed DLDL Approach} \label{sec:pa}

In this section, we firstly give the definition of the DLDL problem. Then, we present the DLDL theory. Next, we propose the construction methods of label distribution for different recognition tasks. Finally, we briefly introduce the DLDL architecture and training details.

\subsection{The deep label distribution learning problem}

 Given an input image, we are interested in estimating a category output $y$~(\emph{e.g.}, age or head pose angles). For two input images $X^1$ and $X^2$ with ground-truth labels $y^1$ and $y^2$, $X^1$ and $X^2$ are supposed to be similar to each other if the correlation of $y^1$ and $y^2$ is strong, and vice versa. For example, the correlation between faces aged 32 and 33 should be stronger than that between faces aged 32 and 64, in terms of facial details that reflect the age (\emph{e.g.}, skin smoothness). In other words, we expect high correlation among input images with similar outputs.
The label distribution learning approach~\cite{geng2013facial,geng2014head} exploited such correlations in the machine learning phase, but used features that are extracted ignoring these correlations. The proposed DLDL approach, however, is an end-to-end deep learning method which utilizes such correlation information in both feature learning and classifier learning. We will also extend DLDL to handle other types of label ambiguity beyond correlation.

To fulfill this goal, instead of outputting a single value $y$ for an input $X$, DLDL quantizes the range of possible $y$ values into several \emph{labels}. For example, in age estimation, it is reasonable to assume that $0 < y \le 85$, and it is a common practice to estimate integer values for ages. Thus, we can define the set $L=\{1,2,\dots,85\}$ as the ordered label set for age estimation. The task of DLDL is then to predict a label distribution $\vec{y} \in \mathbb{R}^{85}$, where $y_i$ is the estimated probability that $X$ should be predicted to be $i$ years old. By estimating an entire label distribution, the deep learning machine is forced to take care of the ambiguity among labels.

Specifically, the input space of our framework is $\mathscr{X} = \mathbb{R}^{h\times w\times d}$, where $h$, $w$ and $d$ are the height, width, and number of channels of the input image, respectively. DLDL predicts a \emph{label distribution} vector $\vec{y} \in \mathbb{R}^{|\mathscr{Y}|}$, where $\mathscr{Y}=\{l_1,l_2,\ldots,l_C\}$ is the label set defined for a specific task (\emph{e.g.}, the $L$ above). We assume $\mathscr{Y}$ is complete, \emph{i.e.}, any possible $y$ value has a corresponding member in $\mathscr{Y}$. A training data set with $N$ instances is then denoted as $D = \{(X^1,\vec y^1), \cdots, (X^N,\vec y^N)\}$. We use boldface lowercase letters like $\vec y$ to denote vectors, and the $i$-th element of $\vec y$ is denoted as $y_i$. The goal of DLDL is to directly learn a conditional probability mass function $\vec {\hat y} = p(\vec {y}|X;\vec \theta)$ from $D$, where $\vec \theta$ is the parameters in the framework.

\subsection{Deep label distribution learning} \label{sec:udldl}

Given an instance ${X}$ with label distribution $\vec y$, we assume that $\vec x = \phi (X;\vec \theta)$ is the activation of the last fully connected layer in a deep ConvNet. We use a softmax function to turn these activations into a probability distribution, that is,
\begin{equation}
	\hat y_j  = \frac{\exp(x_j)}{\sum_t \exp(x_t)} \,. \label{eq-sm}
\end{equation}
Given a training data set $D$, the goal of DLDL is to find $\vec \theta$ to generate a distribution $\vec{\hat y}$ that is \emph{similar} to $\vec{y}$.

There are different criteria to measure the similarity or distance between two distributions. For example, if the Kullback-Leibler~(KL) divergence is used as the measurement of the similarity between the ground-truth and predicted label distribution, then the best parameter $\vec \theta ^*$ is determined by
\begin{equation}
	\vec{\theta ^*} = \mathop{\argmin}_{\vec \theta} { \sum_k {y_k} \ln \frac{{y_k}}{\hat y_k}}
	= \mathop{\argmin}_{\vec \theta} -{ \sum_k {y_k} \ln{\hat y_k}}  \,.                  \label{eq-mld}
\end{equation}
Thus, we can define the loss function as:
\begin{equation}
	T =  -\sum_k {y_k}\ln{\hat y_k} \,.                      \label{eq-of}
\end{equation}
Stochastic gradient descent is used to minimize the objective function Eq.~\ref{eq-of}. For any $k$ and $j$,
\begin{equation}
	\frac { \partial T}{ \partial \hat y_k} =  -\frac{{y_k}}{\hat y_k} \,,   \label{eq-ld0}
\end{equation}
and the derivative of softmax (Eq.~\ref{eq-sm}) is well known, as
\begin{align}
	\frac { \partial \hat y_k}{ \partial  x_j} =  \hat y_k \left(\delta_{\{k=j\}} - \hat y_j\right) \,, \label{eq-ld1}
\end{align}
where $\delta_{\{k=j\}}$ is 1 if $k=j$, and 0 otherwise.
According to the chain rule, for any fixed $j$, we have
\begin{equation}
	\frac { \partial T}{ \partial x_j} = \sum_k \frac { \partial T}{ \partial \hat y_k} \frac { \partial \hat y_k}{ \partial  x_j}
	= -{y_j} + \hat y_j\sum_k {y_k} = -y_j+\hat y_j \,.
\end{equation}
Thus, the derivative of $T$ with respect to $\vec \theta$ is
\begin{equation}
	\frac { \partial T}{ \partial \vec \theta}
	= \left( \vec {\hat y} - \vec y \right) \frac{ \partial \vec{x}}{ \partial \vec \theta} \,.
\end{equation}

Once $\vec \theta$ is learned, the label distribution $\vec{ \hat{y}}$ of any new instance $X$ can be generated by a forward run of the network. If the expected class label is a single one, DLDL outputs $l_{i^*} \in \mathscr{Y}$, where
\begin{equation}
i^* = \mathop{\argmax}_{i} \hat y_i \,.  \label{eq-max}
\end{equation}
Prediction with multiple labels is also allowed, which could be a set $\{l_i|\hat y_i > \xi\}$ where $\xi \in [0,1]$ is a predefined threshold. If the expected output is a real number, DLDL predicts the expectation of $\hat y_i$, as
\begin{equation}
\sum_i \hat y_il_{i} \,,  \label{eq-exp}
\end{equation}
where $l_i \in \mathscr{Y}$. This indicates that DLDL is suitable for both classification and regression tasks.

\subsection{Label distribution construction} \label{sec:ldc}
The ground-truth label distribution $\vec y$ is not available in most existing datasets, which must be generated under proper assumptions. A desirable label distribution $\vec y = (y_1,y_2,\ldots,y_C)$ must satisfy some basic principles: (1) $\vec y$ should be a probability distribution. Thus, we have $y_i \in [0,1]$ and $\sum_{i=1}^{C}y_i=1$. (2) The probability values $y_i$ should have difference among all possible labels associated with an image. In other words, a less ambiguous category must be assigned high probability and those more ambiguous labels must have low probabilities. In this section, we propose the way to construct label distributions for age estimation, head pose estimation, multi-label classification and semantic segmentation.

For age estimation, we assume that the probabilities should concentrate around the ground-truth age $y$. Thus, we quantize $y$ to get $\vec{y}$ using a normal distribution. For example, the apparent age of a face is labeled by hundreds of users. The ground-truth (including a mean $\mu$ and a standard deviation $\sigma$) is calculated from all the votes. For this problem, we find the range of the target $y$ (\emph{e.g.}, $0<y \le 85$), quantize it into a complete and ordered label set $L=\{l_1,l_2,\dots,l_C\}$, where $C$ is the label set size and $l_i \in \mathbb{R}$ are all possible predictions for $y$. A label distribution $\vec{y}$ is then $(y_1,y_2,\dots,y_C)$, where $y_i$ is the probability that $y=l_i$ (\emph{i.e.}, $y_i=\Pr(y=l_i)$ for $1\le i \le C$). Since we use equal step size in quantizing $y$, the normal p.d.f. (probability density function) is a natural choice to generate the ground-truth $\vec{y}$ from $\mu$ and $\sigma$:
\begin{equation}
	y_j = \frac{p(l_j|\mu,\sigma)} {\sum_k p(l_k|\mu,\sigma)} \,,  \label{eq-ageld}
\end{equation}
where
$
	p(l_j|\mu,\sigma) = \frac{1}{\sqrt{2\pi}\sigma} \exp\left( -\frac{(l_j-\mu)^2}{2 \sigma^2} \right)
$.
Fig.~\ref{fig:age-ld} shows a face and its corresponding label distribution. For problems where $\sigma$ is unknown, we will show that a reasonably chosen $\sigma$ also works well in DLDL.

For head pose estimation, we need to jointly estimate pitch and yaw angles. Thus, learning joint distribution is also necessary in DLDL. Suppose the label set is $L = \{\vec {l}_{jk}|j =1, \cdots, n_1, k = 1, \cdots, n_2\}$, where $\vec {l}_{jk}$ is a pair of values. That is, we want to learn the joint distribution of two variables. Then, the label distribution $\vec y$ can be represented by an $n_1 \times n_2$ matrix, whose $(j,k)$-th element is ${y_{jk}}$. For example, when we use two angles (pitch and yaw) to describe a head pose, $\vec l_{jk}$ is a pair of pitch and yaw angles. Given an instance $X$ with ground-truth mean $\vec{\mu}$ and covariance matrix $\Sigma$, we calculate its label distribution as
\begin{equation}
	y_{jk} = \frac{ p(\vec l_{jk})} {\sum_j \sum_k p(\vec l_{jk})} \,,  \label{eq-mld1}
\end{equation}
where
$
	p(l_{jk}) = \frac{1}{2 \pi {|\Sigma|}^{\frac{1}{2}}}\exp\left(-\frac{1}{2}(\vec { {l}}_{jk}-\vec {\mu})^{\rm T} \Sigma ^{-1} (\vec { {l}}_{jk}-\vec {\mu})\right)   
$.
In the above, we assume $\Sigma = \left(
\begin{array}{cc}
\sigma^2 & 0 \\
0 &\sigma^2 \\
\end{array}
\right) $, that is, the covariance matrix is diagonal. Fig.~\ref{fig:pose-ld} shows a joint label distribution with head pose $\text{pitch}=0^\circ$ and $\text{yaw}=60^\circ$.

For multi-label classification, a multi-label image always contains at least one object of the class of interest. There are usually multiple labels for an image. These labels are grouped into three different levels, including \texttt{Positive}, \texttt{Negative} and \texttt{Difficult} in the PASCAL VOC dataset~\cite{everingham2010pascal}. A label is \texttt{Positive} means an image contains objects from that category, and \texttt{Negative} otherwise. \texttt{Difficult} indicates that an object is clearly visible but difficult to recognize. Existing multi-label methods often view \texttt{Difficult} as \texttt{Negative}, which leads to the loss of useful information. It is not reasonable either if we simply treat \texttt{Difficult} as \texttt{Positive}. Therefore, a nature choice is to use label ambiguity. We define different probabilities for different types of labels, as
\begin{equation}
	  p_P > p_D > p_N,
\end{equation}
for \texttt{Positive}, \texttt{Difficult} and \texttt{Negative} labels, respectively. Furthermore, an $\ell_1$ normalization is applied to ensure $\sum_{i=1}^{C}y_i=1$:
\begin{equation}
	y_{j} = \frac{p(l_{j})} {\sum_k p(l_{k})} \,, \label{eq-mlldn}
\end{equation}
where $p(l_{k})$ equals $p_P$, $p_D$ or $p_N$ if the label $l_k$ is \texttt{Positive}, \texttt{Difficult} or \texttt{Negative}, respectively. The label distribution is shown for a multi-label image in Fig.~\ref{fig:ml-ld}.

For semantic segmentation, we need to label a pixel as belonging to one class if it is a pixel inside an object of that class, or as the background otherwise. Let $y^\prime_{ijk}$ denote the annotation of the $(i,j)$-th pixel, where $k = \{0,1,\ldots,C\}$~(assuming there are $C$ categories and 0 for background). Fully Convolutional Networks~(FCN) have been an effective solution to this task. In FCN~\cite{long2015fully}, a ground-truth label $l$ means that $y^\prime_{ijl} = 1$ and $y^\prime_{ijk}=0$ for all $k \neq l$. However, it is very difficult to specify ground-truth labels for pixels close to object boundaries, because labels of these pixels are inherently ambiguous. We propose a mechanism to describe the label ambiguity in the boundaries. Considering a Gaussian kernel matrix $f_{K\times K}$, we replace the original label distribution $y^\prime$ with $y^{\prime\prime}$, as
\begin{equation}
	y^{\prime\prime}_{ijk} = \sum_{i^\prime=1}^{K}\sum_{j^\prime=1}^{K}f_{i^\prime j^\prime} \times y^\prime_{i^\prime + (i-1)S-P,j^\prime+(j-1)S-P,k}\,. \label{eq-ssld1}
\end{equation}
where $f_{i^\prime j^\prime}\geq 0$, $\sum_{i^\prime=1}^{K}\sum_{j^\prime=1}^{K}f_{i^\prime j^\prime}=1$, $K$ is the kernel size, $P$ and $S$ are padding and stride sizes. In our experiment, we set $K=5$, $P=2$ and $S=1$, and the generated label distribution is
\begin{equation}
	y_{ijk} = \frac{y^{\prime\prime}_{ijk}}{\sum_k y^{\prime\prime}_{ijk}}\,. \label{eq-ssld2}
\end{equation}
Fig.~\ref{fig:ss-ld} gives the semantic label distribution for a bird image which shows that the ambiguity is encoded in the label distributions.

\subsection{The DLDL architecture and training details}

We use a deep ConvNet and a training set $D$ to learn a $\hat{\vec{y}}$ as the estimation of $\vec{y}$. The structure of our network is based on popular deep models such as ZF-Net~\cite{zeiler2014visualizing} and VGG-Nets~\cite{simonyan2015very}. The ZF-Net consists five convolution layers, followed by three fully connected layers.  The VGG-Nets architecture includes 16 or 19 layers. We modify the last fully connected layer's output based on the task and replace the original softmax loss function with the KL loss function. In addition, we use the parameter ReLU~\cite{he2015delving} for ZF-Net. In our network, the input is an order three tensor $X_{h\times w \times d}$ and the output $\hat{\vec{y}}$  may be a vector~(age estimation and multi-label classification), a matrix~(head pose estimation) or a tensor~(semantic segmentation).

In this paper, we train the deep models in two ways:

\textbf{Training from scratch.} For ZF-Net, the initialization is performed randomly, based on a Gaussian distribution with zero mean and 0.01 standard deviation, and biases are initialized to zero. The coefficient of the parameter ReLU is initialized to 0.25. The dropout is applied to the last two fully connected layers with rate 0.5. The coefficient of weight decay is set to $0.0005$. Optimization is done by Stochastic Gradient Descent (SGD) using mini-batches of 128 and the momentum coefficient is 0.9. The initial learning rate is set to 0.01. The total number of epochs is about 20.

\textbf{Fine-tuning.} Three pre-trained models including VGG-Nets~(16-layers and 19-layers) and VGG-Face~(16-layers) are used to fine-tune for different tasks. We remove these pre-trained models' classification layer and loss layer, and put in our label distribution layer which is initialized by the Gaussian distribution $N(0,0.01)$ and the KL loss layer. The learning rates of the convolutional layers, the first two fully-connected layers and the label distribution layer are initialized as 0.001, 0.001 and 0.01, respectively. We fine-tune all layers by back propagation through the whole net using mini-batches of 32. The total number of epochs is about 10 for age estimation and 20 for multi-label classification.

\section{Experiments}\label{sec:ex}
We evaluate DLDL on four tasks,~\emph{i.e.}, age estimation, head pose estimation, multi-label classification and semantic segmentation. Our implementation is based on MatConvNet~\cite{vedaldi15matconvnet}.\footnote{\url{http://www.vlfeat.org/matconvnet/}} All our experiments are carried out on a NVIDIA K40 GPU with 12GB of onboard memory.

\subsection{Age estimation} \label{subsec:ex-age}
\textbf{Datasets.} Two age estimation datasets are used in our experiments. The first is \emph{Morph}~\cite{ricanek2006morph}, which is one of the largest publicly available age datasets. There are 55,134 face images from more than 13,000 subjects. Ages range from 16 to 77. Since no \texttt{TRAIN/TEST} split is provided, 10-fold cross-validation is used for \emph{Morph}.

The second dataset is from the apparent age estimation competition, the first competition track of the ICCV ChaLearn LAP 2015 workshop~\cite{escalera2015chalearn}. Compared with \emph{Morph}, this dataset~(\emph{ChaLearn}) consists of images collected in the wild, without any position, illumination or quality restriction. The only condition is that each image contains only one face. The dataset has 4,699 images, and is split into 2,476 training~(\texttt{TRAIN}), 1,136 validation~(\texttt{VAL}) and 1,087 testing~(\texttt{TEST}) images. The apparent age (\emph{i.e.}, how old does this person look like) of each image is labeled by multiple individuals. The age of face images range from 3 to 85. For each image, its mean age and the corresponding standard deviation are given. Since the ground-truth for \texttt{TEST} images are not published, we train on the \texttt{TRAIN} split and evaluate on the \texttt{VAL} split of \emph{ChaLearn} images.

\textbf{Baselines.}
To demonstrate the effectiveness of DLDL, we firstly consider two related methods as baselines: ConvNet+LS~(KL) and ConvNet+LD~($\alpha$-div). The former uses label smoothing~(LS)~\cite{szegedy2015rethinking} as ground-truth and KL divergence as loss function. The latter uses label distribution~(LD) as ground-truth and $\alpha$ divergence~\cite{minka2005divergence} as loss function, which is
\begin{equation}
	T =  -2\sum_k (\sqrt{y_k}-\sqrt{\hat y_k})^2 \,.                      \label{eq-of2}
\end{equation}
In addition, we also compare DLDL with the following baseline methods:
\begin{itemize}
\item \textbf{BFGS-LDL} Geng \emph{et al.} proposed the label distribution learning approach~(IIS-LLD) for age and head pose estimation. They used traditional image features. To further improve IIS-LLD, Geng \emph{et al.}~\cite{geng2016label} proposed a BFGS-LDL algorithm by using the effective quasi-Newton optimization method BFGS.
\item \textbf{C-ConvNet} Classification ConvNets have obtained very competitive performance in various computer vision tasks. ZF-Net~\cite{zeiler2014visualizing} and VGG-Net are popular models which use the softmax loss. We replace the ImageNet-specific 1000-way classification in these modes with the label set $\mathscr{Y}$.
\item \textbf{R-ConvNet} ConvNets are also successively trained for regression tasks. In R-ConvNet, the ground-truth label $y$ (age and pose angle) is projected into the range $[-1,1]$ by the mapping $\frac{2(y-\min)}{\max-\min}-1$, where $\max$ and $\min$ are the maximum and minimum values in the training label set. During prediction, the R-ConvNet regression result is reverse mapped to get $\hat y$. To speed up convergence, the last fully connected layer is followed a hyperbolic tangent activation function $f(x) = \tanh(x)$, which maps $[-\infty,+\infty]$ to $[-1,+1]$~\cite{ahn2015real}. The squared $\ell_2$, $\ell_1$ and $\epsilon$-ins loss functions are used in R-ConvNet.
\end{itemize}

\begin{figure}
	\centering
	\subfloat[Input]
	{\includegraphics[height= 0.24\columnwidth]{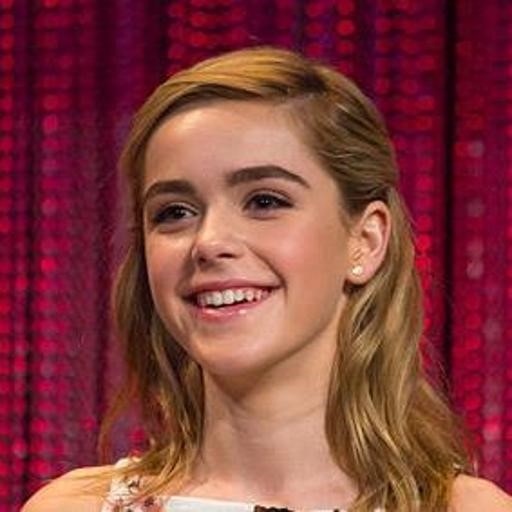}
	}
	\subfloat[Detection]
	{\includegraphics[height= 0.24\columnwidth]{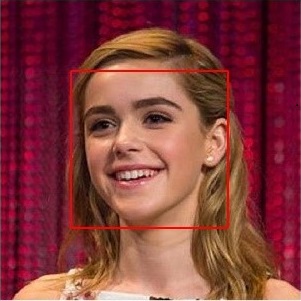}
	}
	\subfloat[Facial points]
	{\includegraphics[height= 0.24\columnwidth]{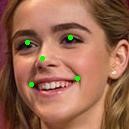}
	}
	\subfloat[Alignment]
	{\includegraphics[height= 0.24\columnwidth]{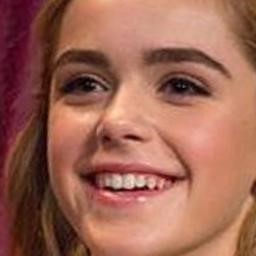}
	}
	\caption{ The face image pre-processing pipeline.} \label{fig:age-pre}
\end{figure}

\textbf{Implementation details.} We use the same preprocessing pipeline for all compared methods, including face detection, facial key points detection and face alignment, as shown in Fig~\ref{fig:age-pre}. We employ the DPM model~\cite{mathias2014face} to detect the main facial region. Then, the detected face is fed into cascaded convolution networks~\cite{sun2013deep} to get the five facial key points, including the left/right eye centers, nose tip and left/right mouth corners. Finally, based on these facial points, we align the face to the upright pose. Data augmentation are only applied to the training images for \emph{ChaLearn}. For one color input training image, we generate its gray-scale version, and left-right flip both color and gray-scale versions. Thus, every training image turns into 4 images.

We define $\mathscr{Y} = \{1,2,\dots,85\}$ for both datasets. The label distribution of each image is generated using Eq.~\ref{eq-ageld}. The mean $\mu$ is provided in both \emph{Morph} and \emph{ChaLearn}. The standard deviation $\sigma$, however, is provided in \emph{ChaLearn} but not in \emph{Morph}. We simply set $\sigma=2$ in \emph{Morph}. Experiments for different methods are conducted under the same data splits.

\begin{table}
	\centering
	\caption{Comparisons of different methods for age estimation. }\label{table:age}
	\small
	\begin{tabular}{|l|ccc|}
		\hline
		 \multirow{2}{*}{Description} & Morph &\multicolumn{2}{c|}{ChaLearn}  \\
		    & MAE    &MAE  &$\epsilon$-error        \\
		\hline
		\hline
		IIS-LDL~\cite{geng2013facial}  &5.67$\pm$0.15  &-  &- \\
		CPNN~\cite{geng2013facial}     &4.87$\pm$0.31  &-  &- \\
        ST+CSHOR~\cite{chang2015learning}\rlap{\textsuperscript{1}} &3.82  &-  &-\\
		M-S ConvNets~\cite{yi2015age}       &3.63 & - &-  \\
		ConvNets~\cite{huerta2015deep}\rlap{\textsuperscript{1}}      &3.31 & - & - \\
        VGG~(softmax, Exp)~\cite{rothe2015dex}\rlap{\textsuperscript{3}}   &-     &6.08 &0.51\\
        VGG~(softmax, Exp)~\cite{rothe2015dex}\rlap{\textsuperscript{2,3}}   &-     &\textbf{3.22} &\textbf{0.28}\\
        VGG~(softmax, Exp)~\cite{rothe2016deep}\rlap{\textsuperscript{2,3}}  &2.68  &3.25 &\textbf{0.28}\\
		\hline
		\hline
		BFGS-LDL~(KL, Max)  &3.94$\pm$0.05  &7.81 &0.57 \\
        BFGS-LDL~(KL, Exp)  &3.85$\pm$0.05  &6.79 &0.53 \\
		C-ConvNet~(softmax, Max)   &3.02$\pm$0.05  &9.48 &0.63 \\
        C-ConvNet~(softmax, Exp)   &2.86$\pm$0.05  &7.95 &0.58 \\
		R-ConvNet~($\ell_2$)      &3.17$\pm$0.04  &5.94 &0.50 \\
	    R-ConvNet~($\ell_1$)      &2.88$\pm$0.03  &5.62 &0.47 \\
	    R-ConvNet~($\epsilon$-ins) &2.89$\pm$0.04  &5.71 &0.48 \\
        ConvNet+LS~(KL, Max)   &2.96$\pm$0.13  &8.64 &0.59 \\
        ConvNet+LS~(KL, Exp)   &5.02$\pm$0.13  &11.58 &0.77 \\
        ConvNet+LD~($\alpha$-div, Max) &2.57$\pm$0.04 & $5.95$ &0.47\\
		ConvNet+LD~($\alpha$-div, Exp) &2.57$\pm$0.04 & $5.69$ &0.46\\
		\hline
		\hline
	    DLDL~(KL, Max)         &2.51$\pm$0.03 & $5.49$ &0.44   \\
        DLDL~(KL, Exp)        &2.52$\pm$0.03 & $5.34$ &0.44   \\
        \hline
        DLDL+VGG-Face~(KL, Max)\rlap{\textsuperscript{3}}    &\textbf{2.42$\pm$0.01} &3.62 &0.32\\
		DLDL+VGG-Face~(KL, Exp)\rlap{\textsuperscript{3}}    &2.43$\pm$0.01 &3.51 &0.31\\ \hline
	\end{tabular}\\
    \leftline{\scriptsize\rlap{\textsuperscript{1}}~Used 80\% of Morph images for training and 20\% for evaluation;}
    \leftline{\scriptsize\rlap{\textsuperscript{2}}~Used additional external face images~(\emph{i.e.}, IMDB-WIKI);}
    \leftline{\scriptsize\rlap{\textsuperscript{3}}~Used pre-trained model~(\emph{i.e.}, VGG-Nets or VGG-Face).}%
\end{table}

\textbf{Evaluation criteria.} Mean Absolute Error (MAE) and Cumulative Score (CS) are used to evaluate the performance of age estimation.  MAE is the average difference between the predicted and the real age:
\begin{equation}
MAE = \frac{1}{N}\sum_{n=1}^N {|\hat{l}_n- l_n|},   \label{eq-mae}
\end{equation}
where $\hat{l}_n$ and $l_n$ are the estimated and ground-truth age of the $n$-th testing image, respectively. CS is defined as the accuracy rate of \emph{correct estimation}:
\begin{equation}
CS_g = \frac{C_g}{N}\times 100\%,   \label{eq-cs}
\end{equation}
where $C_g$ is the number of \emph{correct estimation}, \emph{i.e.}, testing images that satisfy $|\hat{l}_n-l_n| \le g$. In our experiment, $g\in \{1,2,\dots,30\}$. In addition, a special measurement (named $\epsilon$-error) is defined by the \emph{ChaLearn} competition, computed as
\begin{equation}
\epsilon = \frac{1}{N}\sum_{n=1}^N \left(1 - \exp\left(-\frac{(\hat{l}_n- l_n)^2}{2\sigma_n^2}\right)\right) \,.  \label{eq-me}
\end{equation}

\textbf{Results.} Table~\ref{table:age} lists results on both datasets. The upper part shows results in the literature. The middle part shows the baseline results. The lower part shows the results of the proposed approach. The first term in the parenthesis behind each method is the loss function corresponding to the method. \texttt{Max} or \texttt{Exp} represent predicting according to Eq.~\ref{eq-max} or~\ref{eq-exp}, respectively. Since cross-validation is used in \emph{Morph}, we also provide its standard deviations.

\begin{figure*}
	\centering
	\subfloat[ChaLearn]
	{\includegraphics[width= 0.33\textwidth]{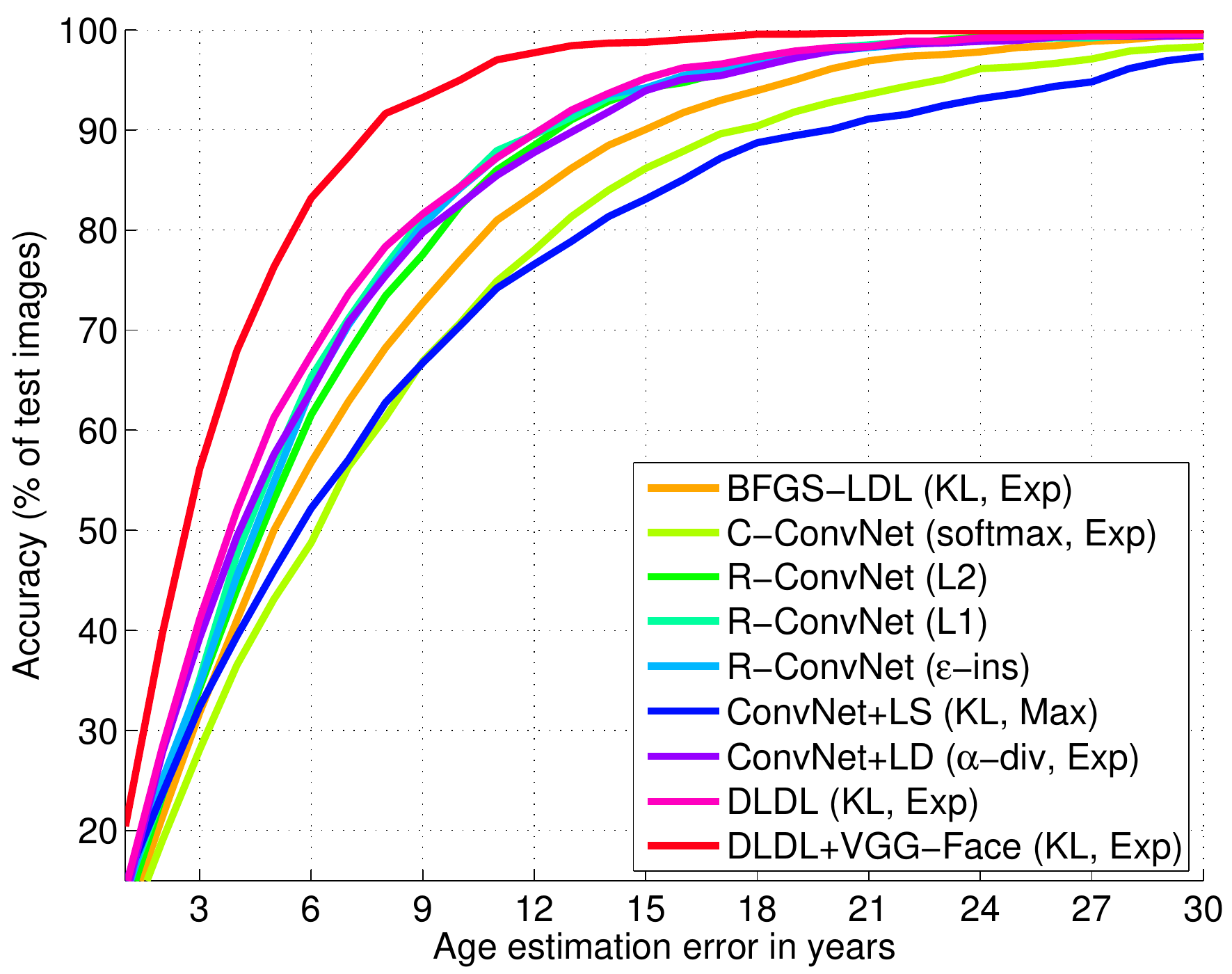}\label{fig:chalearn-cs}}
	\subfloat[Morph]
	{\includegraphics[width= 0.33\textwidth]{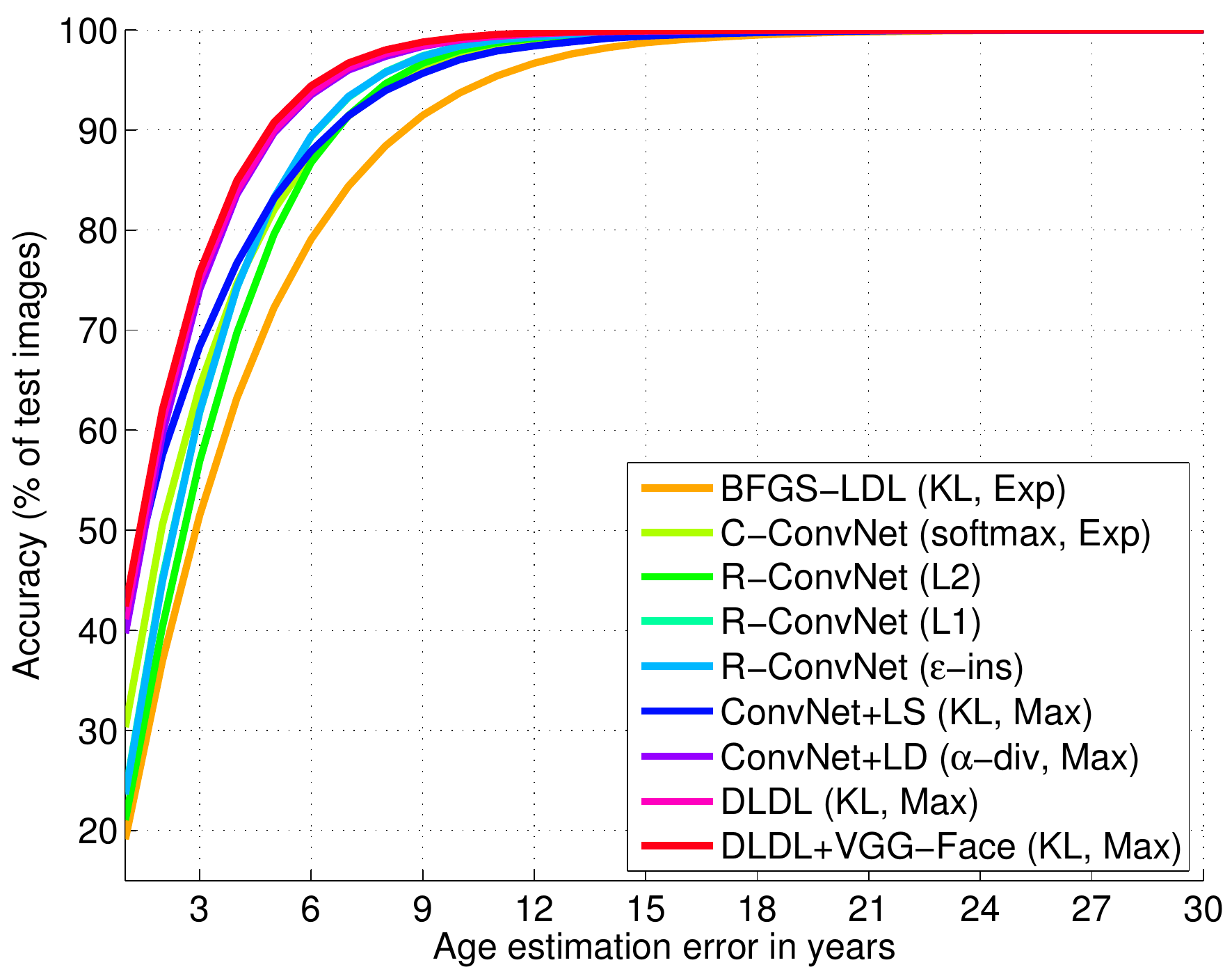}\label{fig:morph-cs}}
    \subfloat[AFLW]
	{\includegraphics[width= 0.33\textwidth]{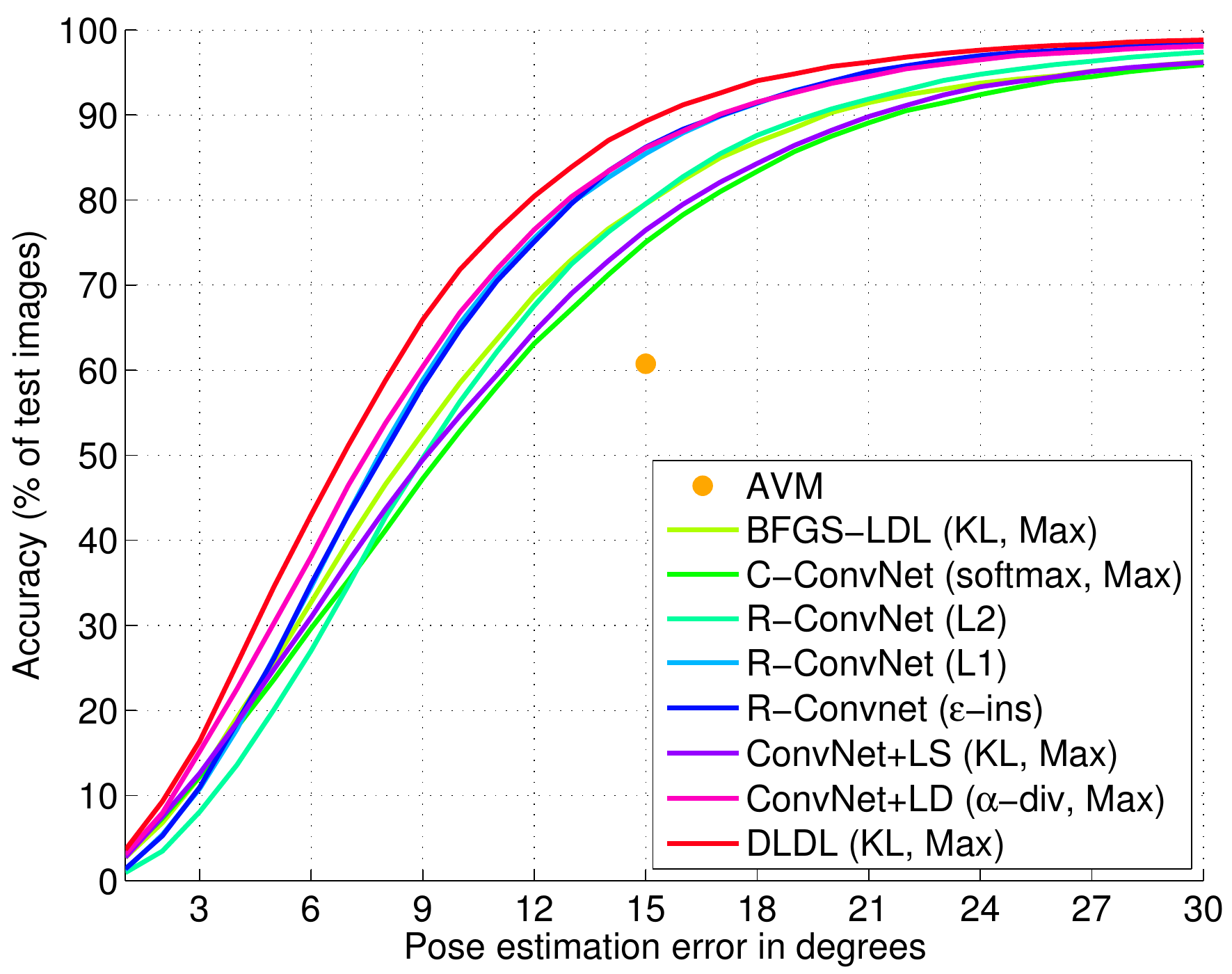}\label{fig:aflw-cs}}
	\centering\caption{Comparisons of CS curves on the \emph{ChaLearn}, \emph{Morph} and \emph{AFLW} validation sets. Note that the CS cures are plotted using better estimation based on Table~\ref{table:age} for those methods involving Max~(Eq.~\ref{eq-max}) and Exp~(Eq.~\ref{eq-exp}) (higher is better, best viewed in color).} \label{fig:cs}
\end{figure*}
\begin{figure*}
    \captionsetup[subfigure]{labelformat=empty}
    \centering
    \subfloat[{40}] {\includegraphics[width= 0.185\columnwidth,keepaspectratio]{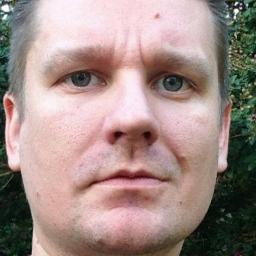}}
    \subfloat[{19}] {\includegraphics[width= 0.185\columnwidth,keepaspectratio]{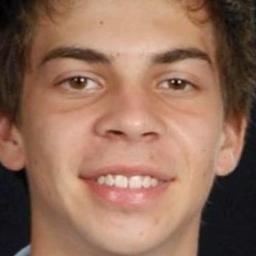}}
    \subfloat[{62}] {\includegraphics[width= 0.185\columnwidth,keepaspectratio]{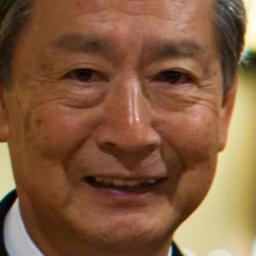}}
    \subfloat[{23}] {\includegraphics[width= 0.185\columnwidth,keepaspectratio]{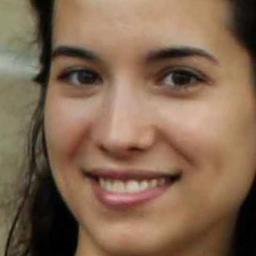}}
    \subfloat[{38}] {\includegraphics[width= 0.185\columnwidth,keepaspectratio]{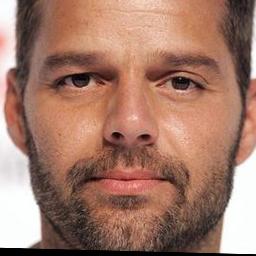}}
    \subfloat[{24}] {\includegraphics[width= 0.185\columnwidth,keepaspectratio]{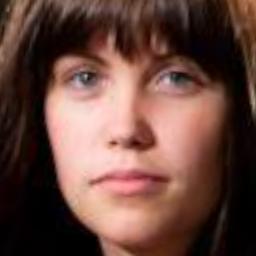}}
    \subfloat[{26}] {\includegraphics[width= 0.185\columnwidth,keepaspectratio]{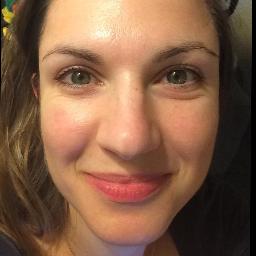}}
    \quad \vrule \quad
    \subfloat[{66}] {\includegraphics[width= 0.185\columnwidth,keepaspectratio]{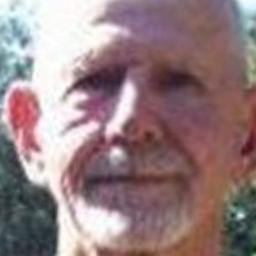}}
    \subfloat[{52}] {\includegraphics[width= 0.185\columnwidth,keepaspectratio]{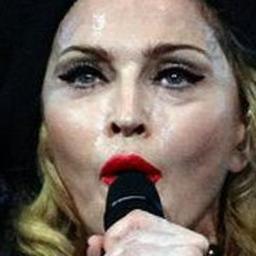}}
    \subfloat[{22}] {\includegraphics[width= 0.185\columnwidth,keepaspectratio]{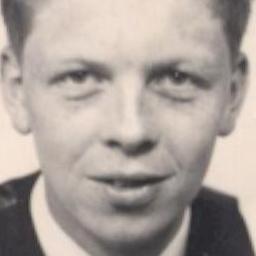}}
    \\ \vspace{-9pt}
    \subfloat[\color{red}{39.69}] {\includegraphics[width= 0.185\columnwidth,keepaspectratio]{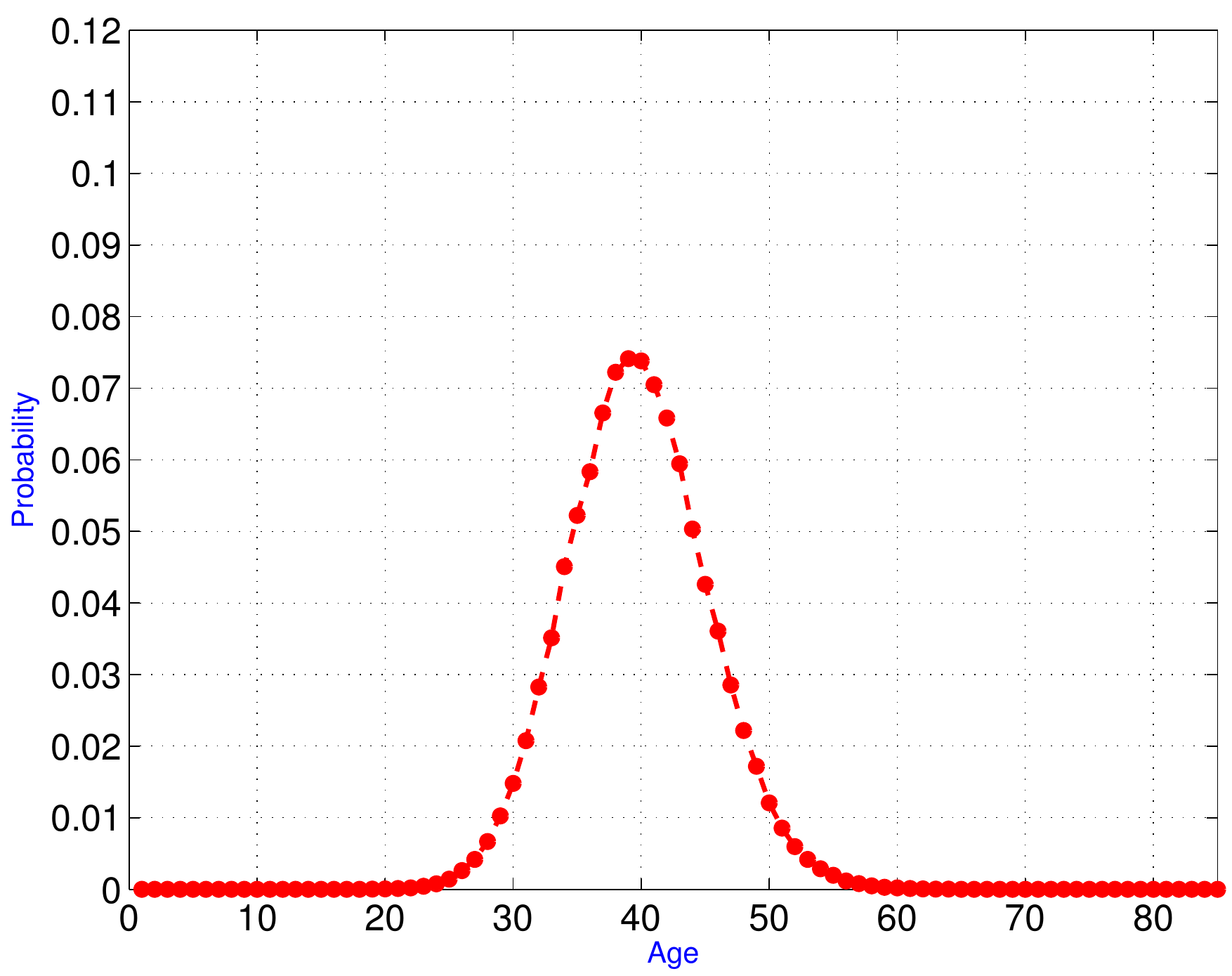}}
    \subfloat[\color{red}{19.29}] {\includegraphics[width= 0.185\columnwidth,keepaspectratio]{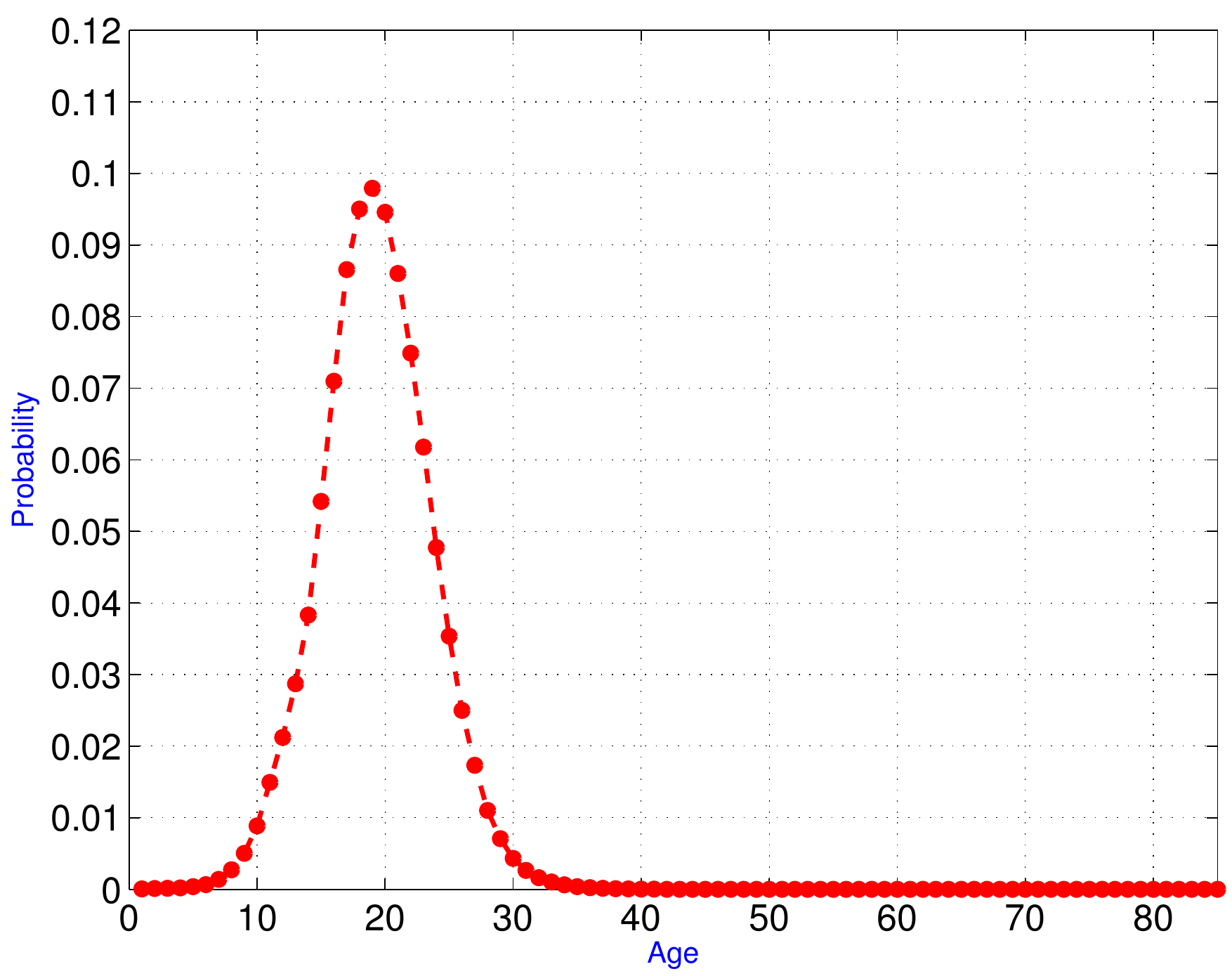}}
    \subfloat[\color{red}{61.61}] {\includegraphics[width= 0.185\columnwidth,keepaspectratio]{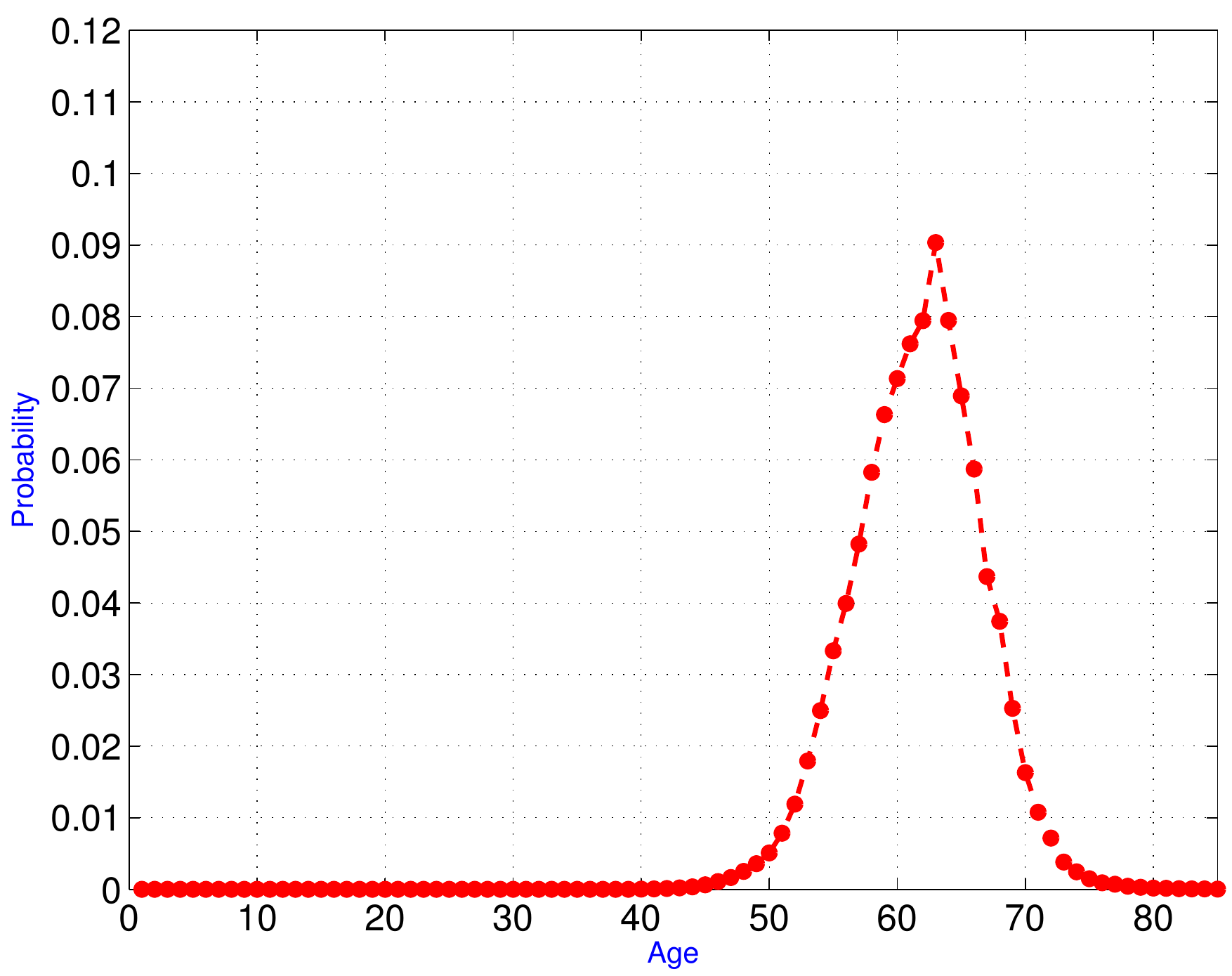}}
    \subfloat[\color{red}{22.94}] {\includegraphics[width= 0.185\columnwidth,keepaspectratio]{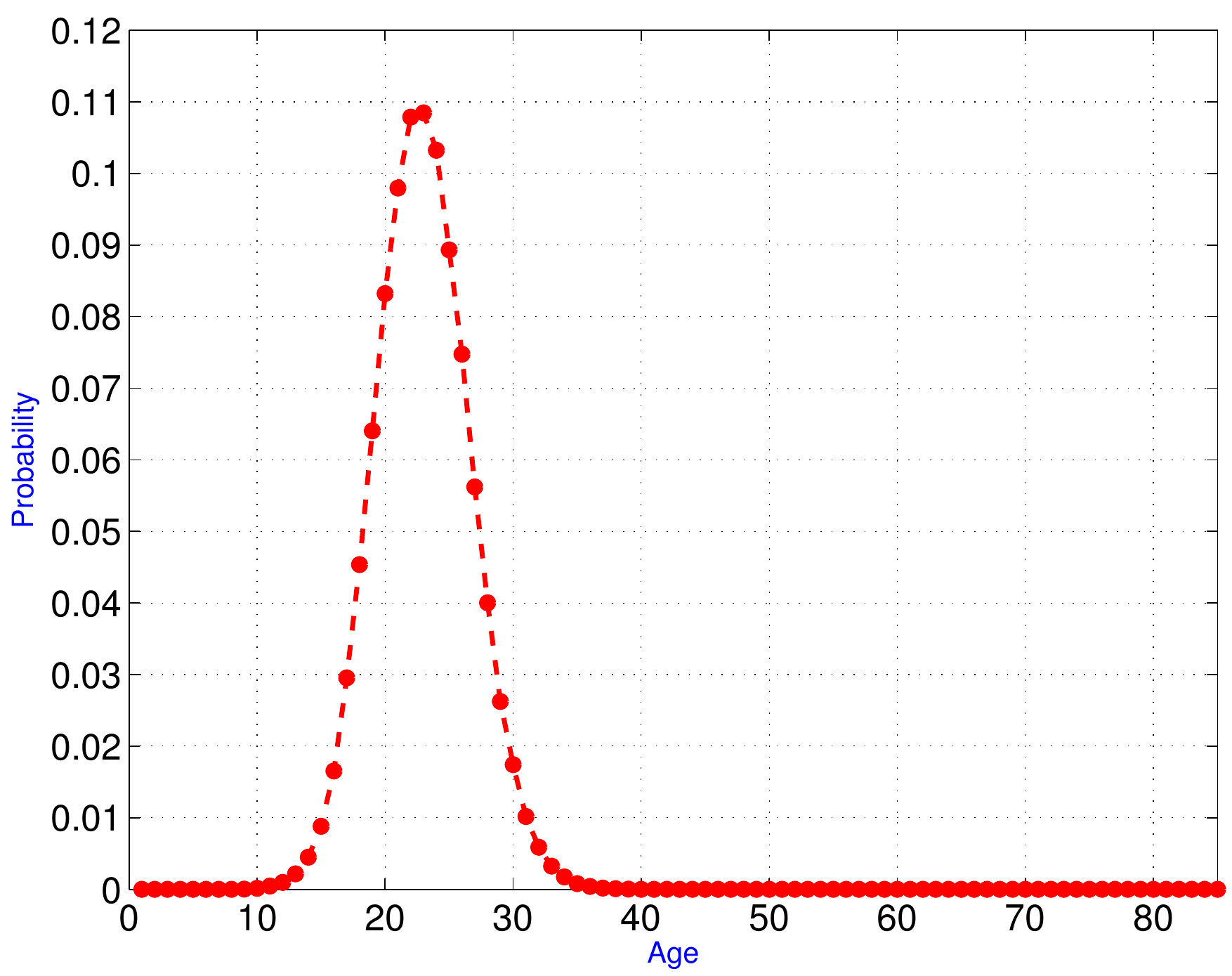}}
    \subfloat[\color{red}{37.87}] {\includegraphics[width= 0.185\columnwidth,keepaspectratio]{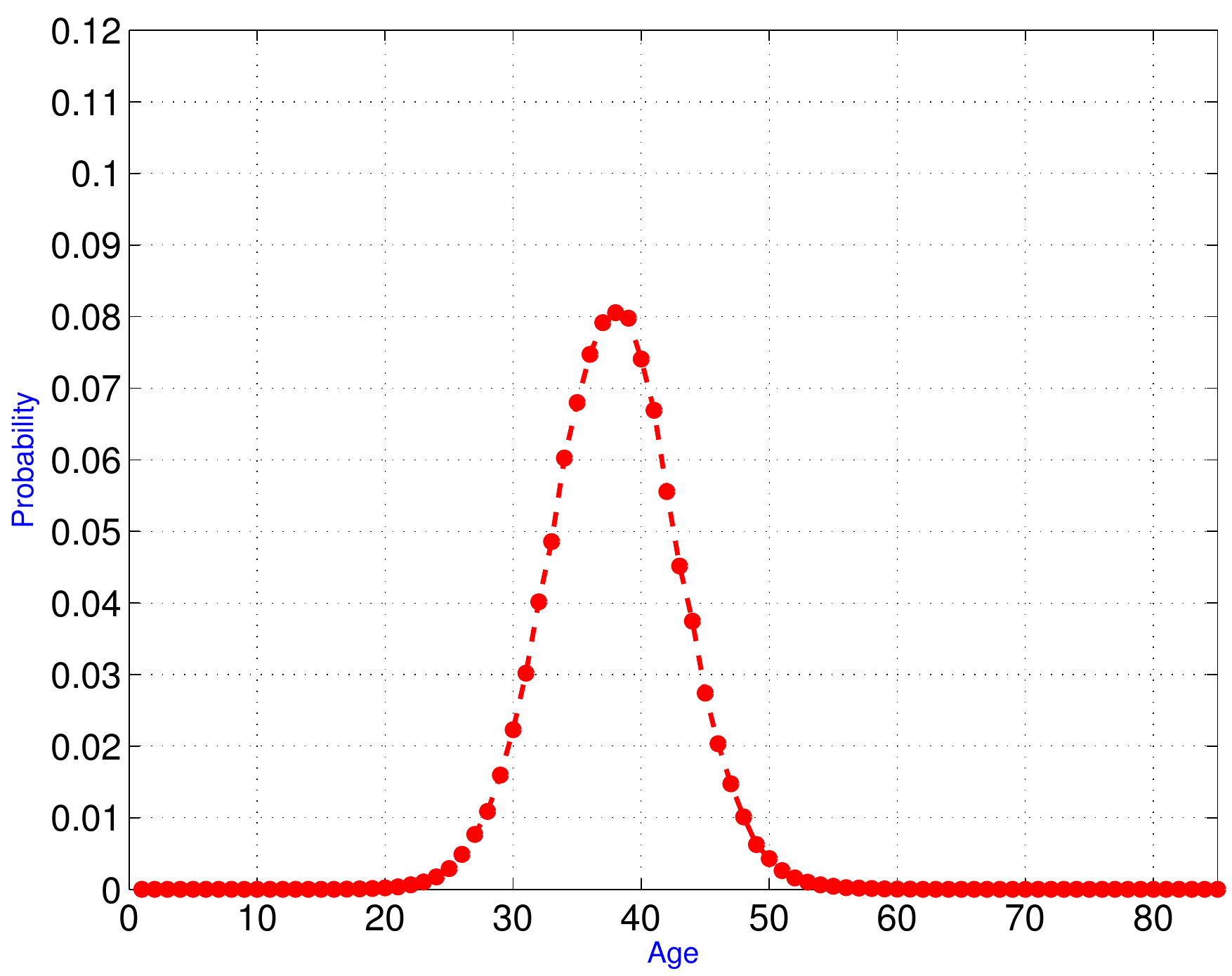}}
    \subfloat[\color{red}{24.27}] {\includegraphics[width= 0.185\columnwidth,keepaspectratio]{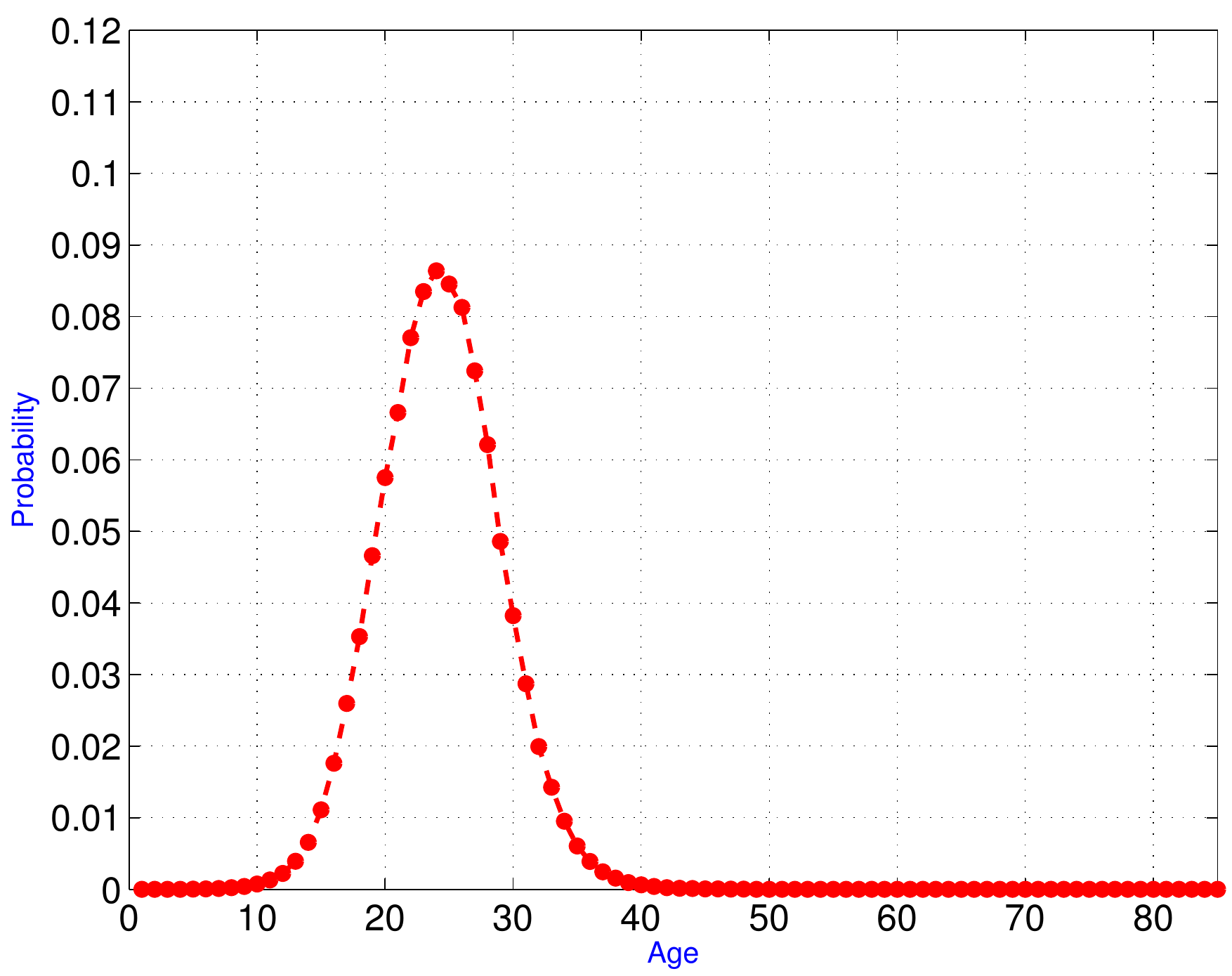}}
    \subfloat[\color{red}{25.40}] {\includegraphics[width= 0.185\columnwidth,keepaspectratio]{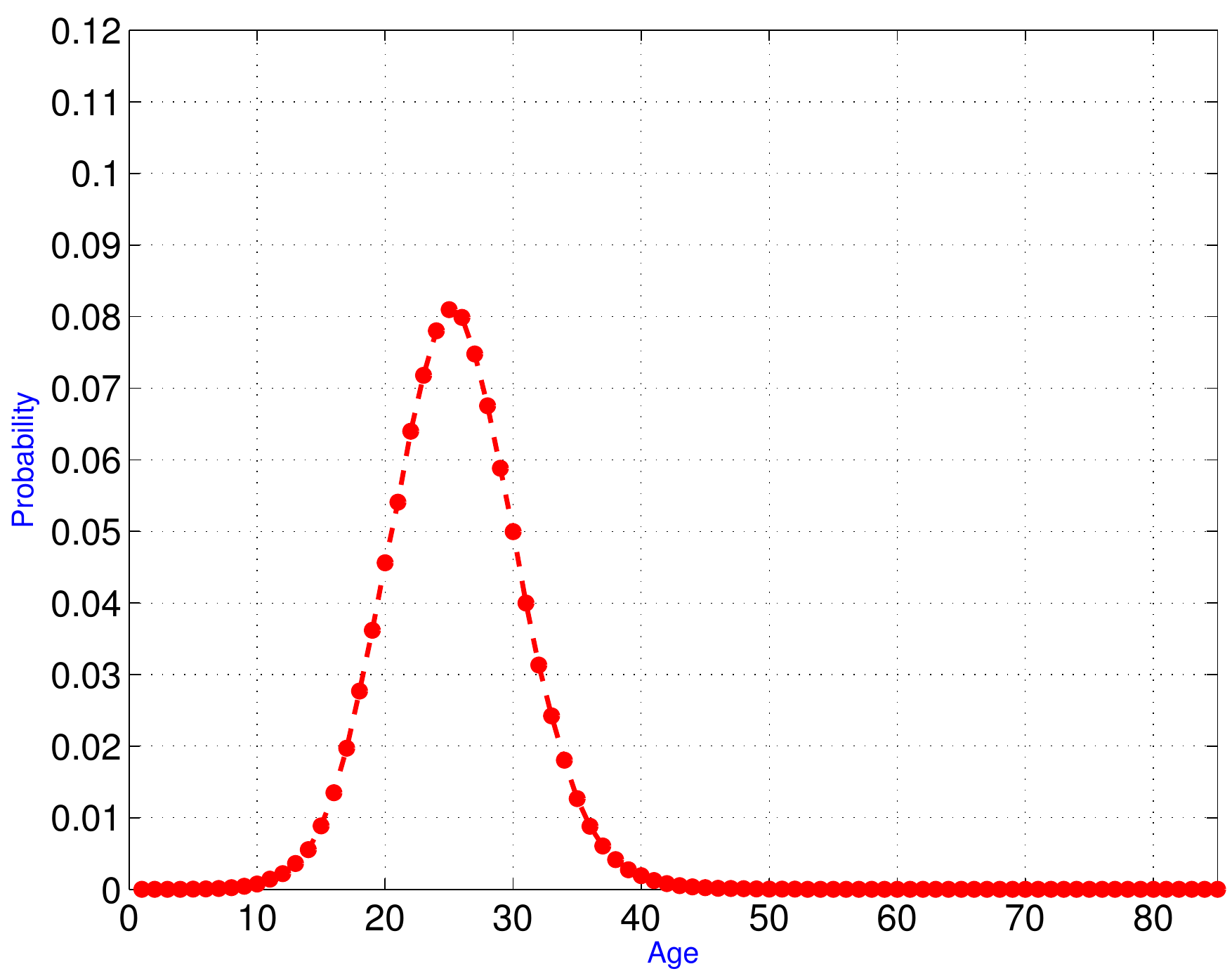}}
    \quad \vrule \quad
    \subfloat[\color{blue}{60.17}]{\includegraphics[width= 0.185\columnwidth,keepaspectratio]{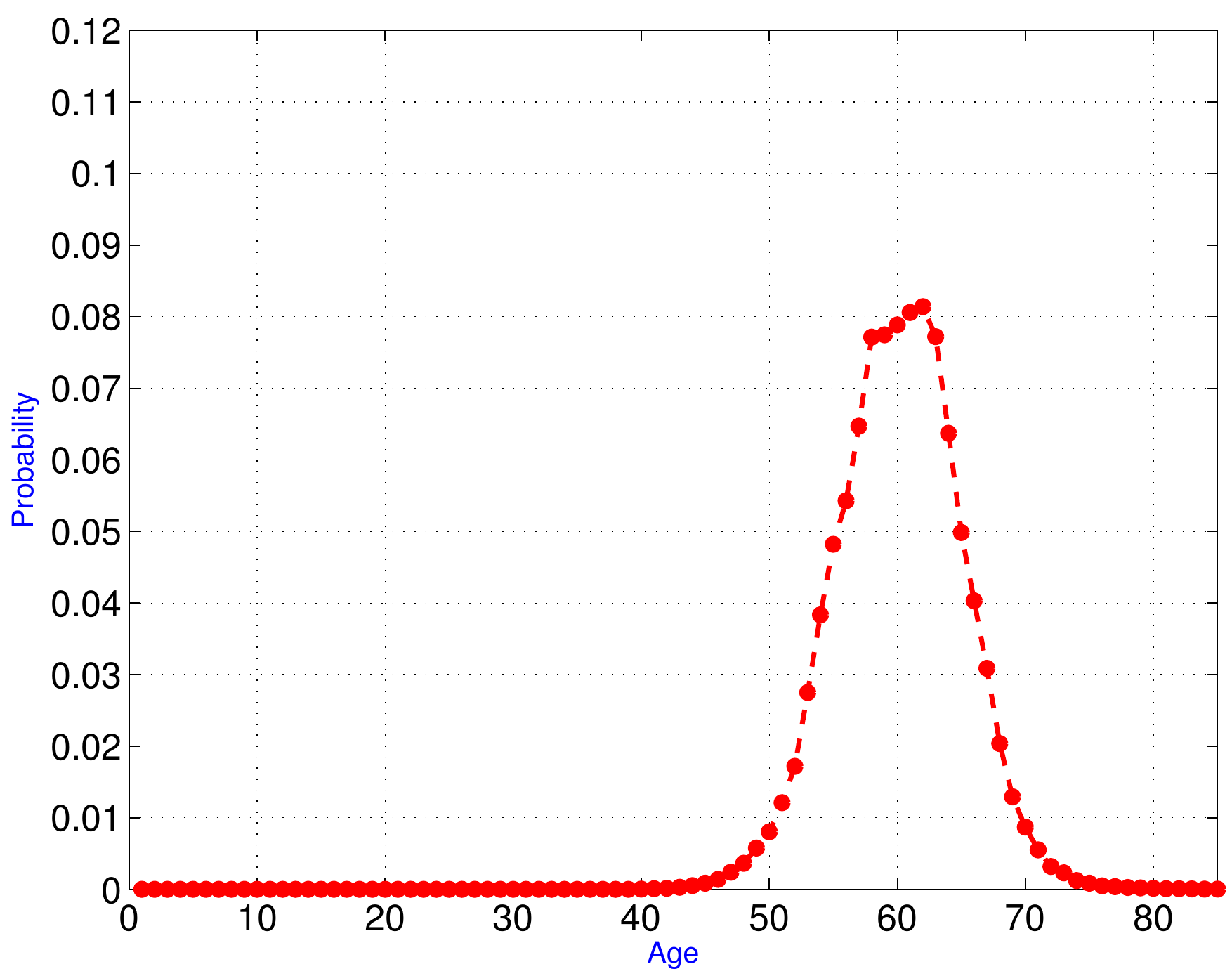}}
    \subfloat[\color{blue}{35.06}]{\includegraphics[width= 0.185\columnwidth,keepaspectratio]{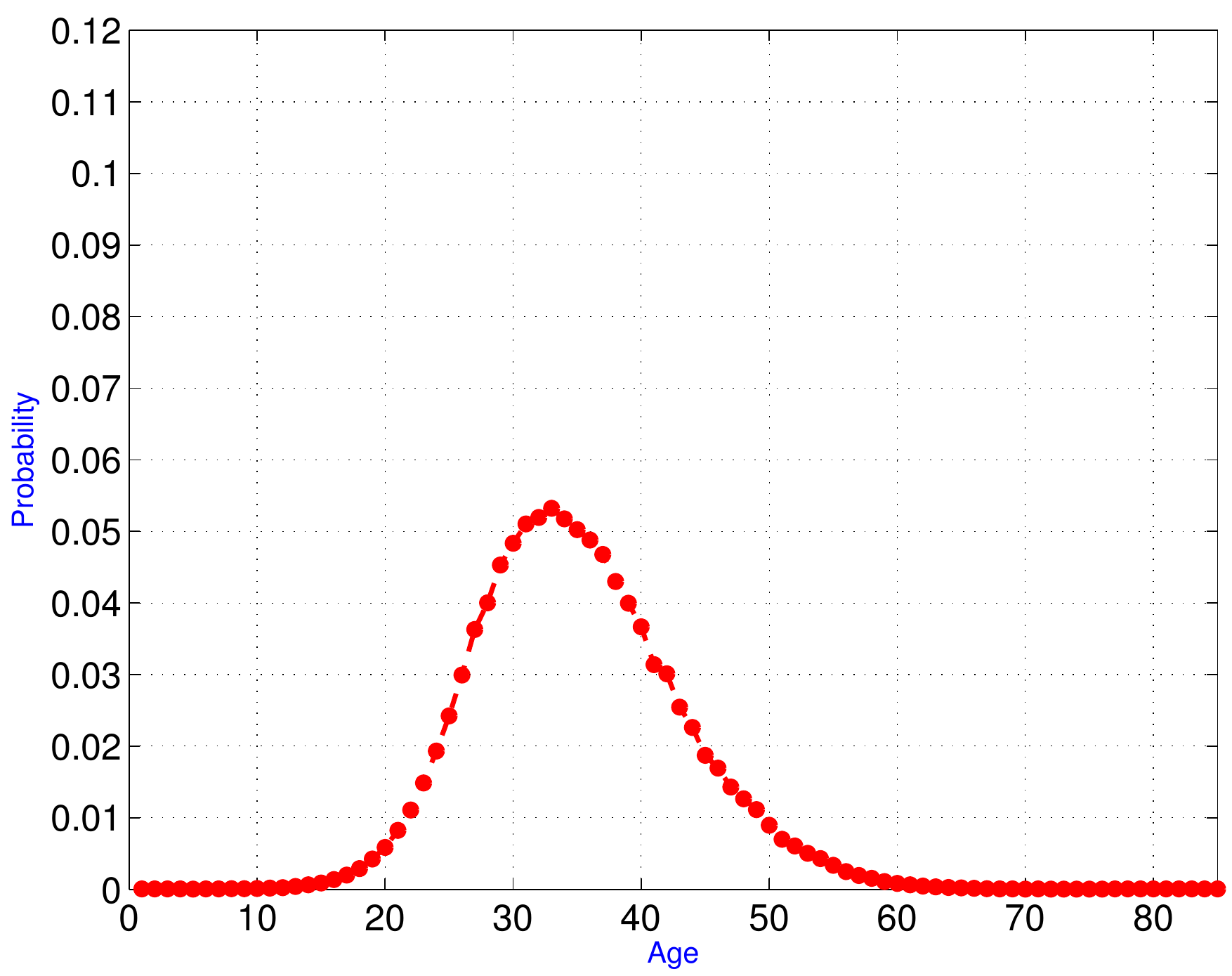}}
    \subfloat[\color{blue}{28.55}]{\includegraphics[width= 0.185\columnwidth,keepaspectratio]{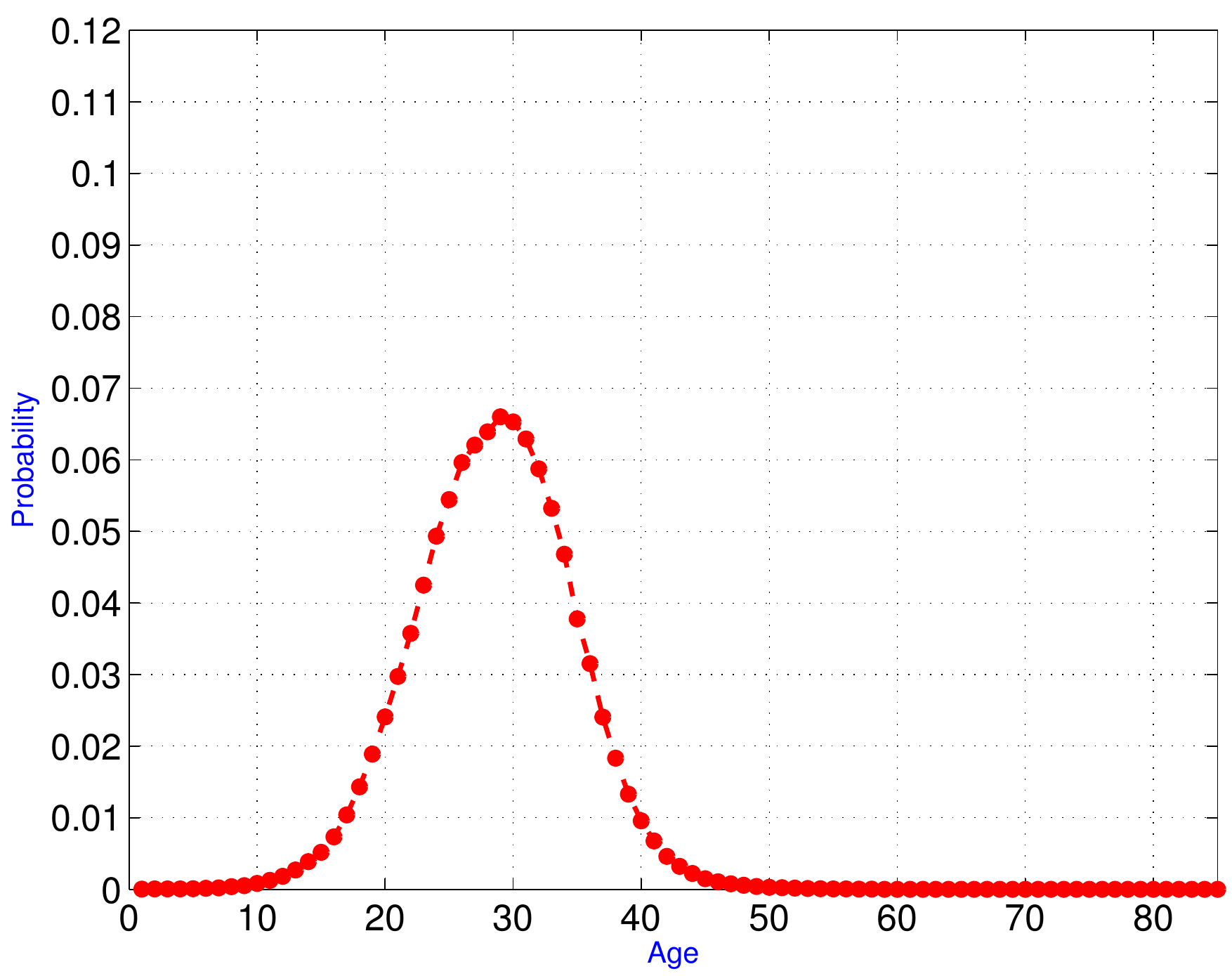}}
 \caption{Examples of face images and DLDL results. The first row shows ten cropped and aligned faces from the apparent age estimation challenge and their corresponding ground-truth apparent ages. The second row shows their predicted label distributions and predicted ages. The left seven columns show good age estimations and the right three columns are failure cases.}\label{fig:eg-age}
\end{figure*}

From Table~\ref{table:age}, we can see that DLDL consistently outperforms baselines and other published methods. The difference between DLDL~(KL, Max) and its competitor C-ConvNet~(softmax, Max) is 0.51 on \emph{Morph}. This gap is more than 6 times the sum of their standard deviations (0.03+0.05), showing statistically significant differences. The advantage of DLDL over R-ConvNet, C-ConvNet and ConvNet+LS suggests that learning label distribution is advantageous in deep end-to-end models. DLDL has much better results than BFGS-LDL, which shows that the learned deep features are more powerful than manually designed ones. Compared to ConvNet+LD~($\alpha$-div), DLDL~(KL) achieves lower MAE on both datasets. It indicates that KL-divergence is better than $\alpha$-divergence for measuring the similarity of two distributions in this context.

We find that C-ConvNet and R-ConvNet are not stable. The R-ConvNet~($\ell_1$) method, although being the second best method for \emph{ChaLearn}, is inferior to C-ConvNet~(softmax, Exp) for \emph{Morph}. In addition, we also find that Eq.~\ref{eq-exp} is better than Eq.~\ref{eq-max} in many cases, which suggests that Eq.~\ref{eq-exp} is more suitable than Eq.~\ref{eq-max} for age estimation.

\textbf{Fine-tuning DLDL.} Instead of training DLDL from scratch, we also fine-tune the network of VGG-Face~\cite{Parkhi15}. On the small scale \emph{ChaLearn} dataset, the MAE of DLDL is reduced from 5.34 to 3.51, yielding a significant improvement. The $\epsilon$-error of DLDL is reduced from 0.44 to 0.31, which is close to the best competition result 0.28~\cite{rothe2015dex} on the validation set. In~\cite{rothe2016deep}, external training images (260,282 additional external training images with real age annotation) were used. DLDL only uses the \emph{ChaLearn} dataset's 2,476 training images and is the best among \emph{ChaLearn} teams that do not use external data~\cite{escalera2015chalearn}. In the competition, the best external-data-free $\epsilon$-error is 0.48, which is worse than DLDL's. However, the idea in~\cite{rothe2016deep} to use external data is useful for further reducing DLDL's estimation error.

Fig.~\ref{fig:chalearn-cs} and Fig.~\ref{fig:morph-cs} show the CS curves on \emph{ChaLearn} and \emph{Morph} datasets. At every error level, our DLDL fine-tuned VGG-Face always achieves the best accuracy among all methods. It is noteworthy that the CS curves of DLDL~(KL, Max) and ConvNet~($\alpha$-div, Max) are very close to that of the DLDL+VGG-Face~(KL, Max) on \emph{Morph} even without lots of external data and very deep model. This observation supports the idea that using DLDL can achieve competitive performance even with limited training samples.

In Fig.~\ref{fig:eg-age}, we show some examples of face images from the \emph{ChaLearn} validation set and predicted label distributions by DLDL~(KL, Exp). In many cases, our solution is able to accurately predict the apparent age of faces. Failures may come from two causes. The first is the failure to detect or align the face. The second is some extreme conditions of face images such as occlusion, low resolution, heavy makeup and old photos.

\subsection{Head pose estimation}
\textbf{Datasets.} We use three datasets in head pose estimation: \emph{Pointing'04}~\cite{gourier2004estimating}, \emph{BJUT-3D}~\cite{baocai2009bjut} and Annotated Facial Landmarks in the Wild~(\emph{AFLW})~\cite{tugraz:icg:lrs}. In them, head pose is determined by two angles: pitch and yaw. \emph{Pointing'04} discretizes the pitch into 9 angles $\{0^\circ, \allowbreak \pm 15^\circ, \allowbreak \pm 30^\circ, \allowbreak \pm 60^\circ, \allowbreak \pm 90^\circ\}$ and the yaw into 13 angles $\{0^\circ, \allowbreak \pm 15^\circ, \allowbreak \pm 30^\circ, \allowbreak \pm 45^\circ, \allowbreak \pm 60^\circ, \pm 75^\circ, \pm 90^\circ\}$. When the pitch angel is $+90^\circ$ or $-90^\circ$, the yaw angle is always set to $0^\circ$. Thus, there are 93 poses in total. The head images are taken from 15 different human subjects in two different time periods, resulting in $15 \times 2 \times 93 = 2,790$ images.

\emph{BJUT-3D} contains 500 3D faces (250 male and 250 female people), acquired by a CyberWare Laser Scanner in an engineered environment. 9 pitch angles $\{0^\circ, \allowbreak \pm 10^\circ, \allowbreak \pm 20^\circ, \allowbreak \pm 30^\circ, \allowbreak \pm 40^\circ\}$ and 13 yaw angles $\{0^\circ, \allowbreak \pm 10^\circ, \allowbreak \pm 20^\circ, \allowbreak \pm 30^\circ, \pm 40^\circ, \pm 50^\circ, \pm 60^\circ\}$ are used. There are in total 93 poses in this dataset, similar to that in \emph{Pointing'04}. Therefore, $500 \times 93 = 46,500$ face images are obtained.

\begin{table*}[h]
	\centering
	\caption{Comparisons of different methods for head pose estimation on the \emph{Pointing'04} dataset. } \label{table:pose-po}
	\footnotesize
	\begin{tabular}{|l|l|ccc|ccc|}
		\hline
		\multirow{2}{*}{Methods} &\multirow{2}{*}{Description}  &\multicolumn{3}{c|}{MAE (lower is better)} &\multicolumn{3}{c|}{Acc (higher is better)}\\
		& & Pitch & Yaw & Pitch+Yaw  &Pitch & Yaw & Pitch+Yaw\\ \hline \hline
		\multirow{1}{*}{}
		&LDL-wJ~\cite{geng2014head} &2.69$\pm$0.15  &4.24$\pm$0.17 &6.45$\pm$0.29 &86.24$\pm$0.97  &73.30$\pm$1.36 &64.27$\pm$1.82  \\
		\hline \hline
		\multirow{7}{*}{Baselines}
		&BFGS-LDL~(KL)      &1.99$\pm$0.19   &4.00$\pm$0.20 	&5.68$\pm$0.13 	&88.78$\pm$0.11 &74.37$\pm$0.13 &66.42$\pm$0.11\\
		&C-ConvNet~(softmax)  &5.28$\pm$0.65 	&6.02$\pm$0.44 	&10.56$\pm$0.74 &73.15$\pm$2.74 &62.90$\pm$1.81 &42.97$\pm$1.67 \\
		&R-ConvNet~($\ell_2$)  &6.11$\pm$0.33   &6.61$\pm$0.17  &10.13$\pm$0.26 &- &- &-  \\
		&R-ConvNet~($\ell_1$)  &5.94$\pm$0.71   &5.90$\pm$0.39  &9.43$\pm$0.79  &- &- &-  \\
		&R-ConvNet~($\epsilon$-ins)  &5.77$\pm$0.45 &6.66$\pm$0.19 &9.04$\pm$0.40 &- &- &-  \\
        &ConvNet+LS~(KL)  &5.23$\pm$0.39 	&5.87$\pm$0.53 	&10.42$\pm$0.66 &72.62$\pm$1.01 &62.90$\pm$2.76 &41.83$\pm$2.20 \\
        &ConvNet+LD~($\alpha$-div)
		&1.94$\pm$0.20 	&3.68$\pm$0.16  &5.34$\pm$0.17	&90.00$\pm$0.77 	&76.27$\pm$0.82	 &69.00$\pm$0.89 \\
		\hline \hline
		\multirow{1}{*}{Ours}
		&DLDL~(KL)
		&\textbf{1.69$\pm$0.32} 	&\textbf{3.16$\pm$0.07}  &\textbf{4.64$\pm$0.24} 	&\textbf{91.65$\pm$1.13} 	&\textbf{79.57$\pm$0.57}	 &\textbf{73.15$\pm$0.72} \\ \hline
	\end{tabular}
\end{table*}

\begin{table*}
	\centering
	\caption{Comparisons of different methods for head pose estimation on the \emph{BJUT-3D} dataset.} \label{table:pose-bj}
	\footnotesize
	\begin{tabular}{|l|l|ccc|ccc|}
		\hline
		\multirow{2}{*}{Methods} &\multirow{2}{*}{Description}  &\multicolumn{3}{c|}{MAE (lower is better)} &\multicolumn{3}{c|}{Acc (higher is better)}\\
		& & Pitch & Yaw & Pitch+Yaw  &Pitch & Yaw & Pitch+Yaw\\
        \hline \hline
		\multirow{7}{*}{Baselines}
		&BFGS-LDL~(KL)     &0.19$\pm$0.02 &0.33$\pm$0.04 &0.51$\pm$0.05 &98.15$\pm$0.19 &96.69$\pm$0.38 &94.95$\pm$0.54\\
		&C-ConvNet~(Softmax) &0.06$\pm$0.01 &0.09$\pm$0.02 &0.14$\pm$0.03 &99.45$\pm$0.09 &99.16$\pm$0.16 &98.64$\pm$0.23  \\
		&R-ConvNet~($\ell_2$) &1.83$\pm$0.01 &2.17$\pm$0.03 &3.15$\pm$0.03 &- &- &- \\
        &R-ConvNet~($\ell_1$) &1.25$\pm$0.06 &1.37$\pm$0.09 &2.11$\pm$0.09 &- &- &- \\
        &R-ConvNet~($\epsilon$-ins) &1.21$\pm$0.07 &1.42$\pm$0.07 &2.09$\pm$0.10 &- &- &-\\
        &ConvNet+LS~(KL) &0.05$\pm$0.01 &0.08$\pm$0.01 &0.12$\pm$0.01 &99.55$\pm$0.06 &99.28$\pm$0.08 &98.86$\pm$0.10\\
        &ConvNet+LD~($\alpha$-div) &0.07$\pm$0.01 &0.12$\pm$0.02 &0.19$\pm$0.02 &99.31$\pm$0.04 &98.82$\pm$0.20 &98.15$\pm$0.21\\
        \hline \hline
		\multirow{1}{*}{Ours}
		&DLDL~(KL) &\textbf{0.02$\pm$0.01} &\textbf{0.07$\pm$0.01} &\textbf{0.09$\pm$0.01} &\textbf{99.81$\pm$0.04} &\textbf{99.27$\pm$0.08}&\textbf{99.09$\pm$0.09}\\
        \hline
	\end{tabular}
\end{table*}

Unlike \emph{Pointing'04} and \emph{BJUT-3D}, the \emph{AFLW} is a real-world face database. Head pose is coarsely obtained by fitting a mean 3D face with the POSIT algorithm~\cite{dementhon1995model}. The dataset contains about 24k faces in real-world images. We select 23,409 faces to ensure pitch and yaw angles within $[-90^\circ,90^\circ]$.

\textbf{Implementation details.} The head region is provided by bounding box annotations in \emph{Pointing'04} and \emph{AFLW}. The \emph{BJUT-3D} does not contain background regions.  Therefore, we will not perform any preprocessing.

In DLDL, we set $\sigma = 15^\circ$ in \emph{Pointing'04} and $\sigma = 5^\circ$ in \emph{BJUT-3D} for constructing label distributions. For \emph{AFLW}, ground-truth of head pose angles are given as real numbers. Ground-truth~(pitch and yaw) angles are divided from $-90^\circ$ to $+90^\circ$ in steps of $3^\circ$, so we get $61 \times 61 =3,721$ (pitch, yaw) pair category labels. We set $\sigma = 3^\circ$ for \emph{AFLW}. Since the discrete Jeffrey's divergence is used in LDL~\cite{geng2014head}, we implement BFGS-LDL with the Kullback-Leibler divergence. All experiments are performed under the same setting, including data splits, input size and network architecture.

To validate the effectiveness of DLDL for head pose estimation, we use the same baselines as age estimation. Our experiments show that Eq.~\ref{eq-exp} has lower accuracy than Eq.~\ref{eq-max}. Hence, we use Eq.~\ref{eq-max} in this section.

\textbf{Evaluation criteria.} Three types of prediction values are evaluated: pitch, yaw, and pitch+yaw, where pitch+yaw jointly estimates the pitch and yaw angles. Two different measurements are used, which is MAE (Eq.~\ref{eq-mae}) and classification accuracy (\texttt{Acc}). When we treat different poses as different classes, \texttt{Acc} measures the pose class classification accuracy. In particular, the MAE of pitch+yaw is calculated as the Euclidean distance between the predicted (pitch, yaw) pair and the ground-truth  pair; the \texttt{Acc} of pitch+yaw is calculated by regarding each (pitch, yaw) pair as a class. For R-ConvNet, we only report its MAE but not \texttt{Acc}, because its predicted value are continuous real numbers. All methods are tested with 5-fold cross validation for \emph{Pointing'04} and \emph{BJUT-3D} following~\cite{geng2014head}. For \emph{AFLW}, 15,561 face images are randomly chosen for training, and the remaining 7,848 for evaluation. The setup is similar to the recent literature~\cite{sundararajan2015head}~(14,000 images for training and the rest 7,041 images for testing).

\begin{table}
 \centering
 \caption{MAE and Acc ($\%$ of images with $\pm15^\circ$ error) for different methods on the \emph{AFLW} dataset.} \label{table:pose-af}
 \footnotesize
 \setlength{\tabcolsep}{2pt}
 \begin{tabular}{|l|ccc|ccc|}
  \hline
 \multirow{2}{*}{Description}  &\multicolumn{3}{c|}{MAE (lower is better)} &\multicolumn{3}{c|}{Acc (higher is better)}\\
     &Pitch &Yaw &Pitch+Yaw &Pitch &Yaw &Pitch+Yaw\\
    \hline \hline
    AVM~\cite{sundararajan2015head}   &- &16.75 &- &- &60.75 &-\\
   \hline \hline
   BFGS-LDL~(KL)     &7.21  &8.72 &12.69 &90.62 &86.81 &79.80\\
   C-ConvNet~(softmax) &7.87  &9.34 &13.65 &87.75 &83.79 &75.04 \\
  R-ConvNet~($\ell_2$) &6.57  &8.44 &11.88 &92.84 &84.76 &79.56 \\
  R-ConvNet~($\ell_1$) &6.01  &7.07 &10.34 &94.60 &89.62 &85.45 \\
  R-ConvNet~($\epsilon$-ins) &5.96 &7.13 	&10.35 	&94.94 &90.00 &86.21\\
  ConvNet+LS~(KL)     &7.69  &9.10 &13.33 &88.34 &85.00 &76.47 \\
  ConvNet+LD~($\alpha$-div) &6.55 &7.02 &10.77 &92.80 &91.88 &86.14\\
  \hline \hline
  DLDL~(KL) &\textbf{5.75} &\textbf{6.60} &\textbf{9.78} &\textbf{95.41} &\textbf{92.89} &\textbf{89.27}\\
  \hline
 \end{tabular}
\end{table}

\begin{figure*}
    \captionsetup[subfigure]{labelformat=empty}
    \centering
    \subfloat[{(+77$^\circ$,-4$^\circ$)}] {\includegraphics[width= 0.2\columnwidth]{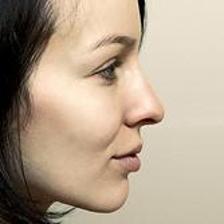}}
    \subfloat[{(-16$^\circ$,-1$^\circ$)}] {\includegraphics[width= 0.2\columnwidth]{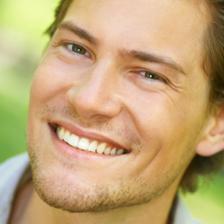}}
    \subfloat[{(-1$^\circ$,-30$^\circ$)}] {\includegraphics[width= 0.2\columnwidth]{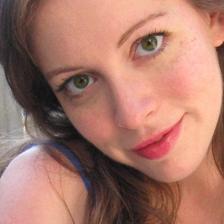}}
    \subfloat[{(+30$^\circ$,+8$^\circ$)}] {\includegraphics[width= 0.2\columnwidth]{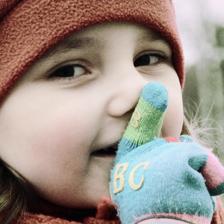}}
    \subfloat[{(+4$^\circ$,-4$^\circ$)}]  {\includegraphics[width= 0.2\columnwidth]{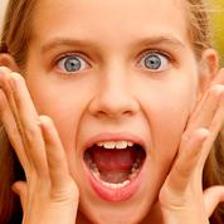}}
    \subfloat[{(-36$^\circ$,+13$^\circ$)}]{\includegraphics[width= 0.2\columnwidth]{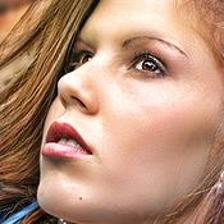}}
    \subfloat[{(-87$^\circ$,-3$^\circ$)}] {\includegraphics[width= 0.2\columnwidth]{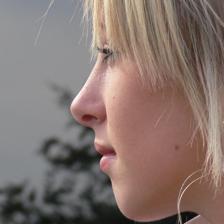}}
    \hspace{2pt} \vrule \hspace{2pt}
    \subfloat[{(-61$^\circ$,-58$^\circ$)}]{\includegraphics[width= 0.2\columnwidth]{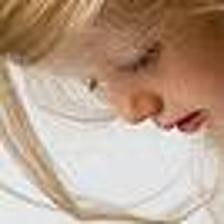}}
    \subfloat[{(+63$^\circ$,+12$^\circ$)}]{\includegraphics[width= 0.2\columnwidth]{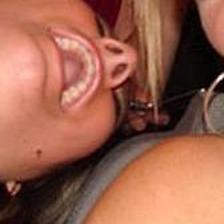}}
    \subfloat[{(+80$^\circ$,-27$^\circ$)}]{\includegraphics[width= 0.2\columnwidth]{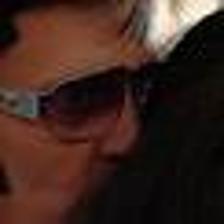}}
    \\ \vspace{-8pt}
    \subfloat[\color{red}{(+75$^\circ$,-3$^\circ$)}]  {\includegraphics[width= 0.2\columnwidth]{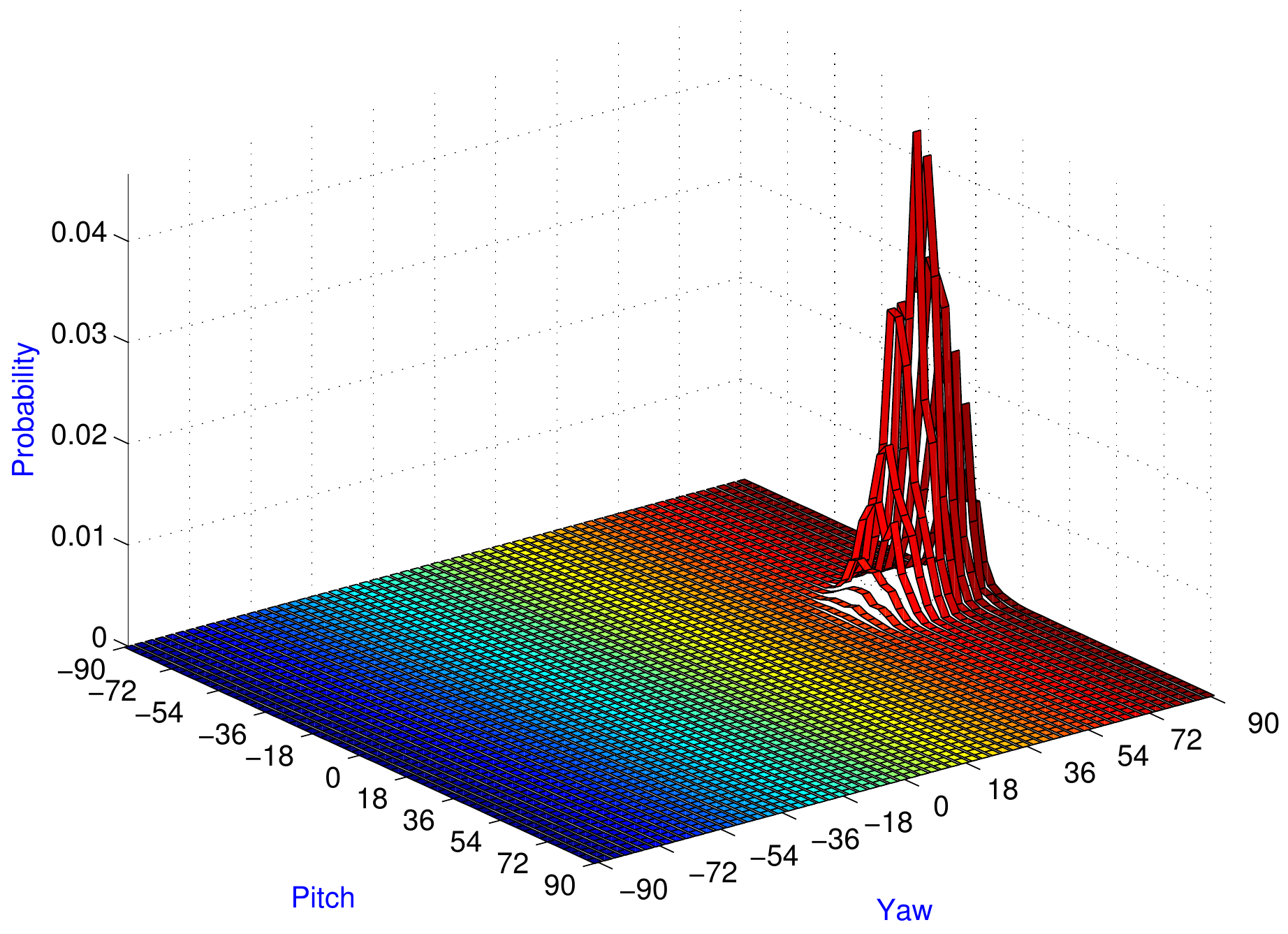}}
    \subfloat[\color{red}{(-15$^\circ$,0$^\circ$)}]   {\includegraphics[width= 0.2\columnwidth]{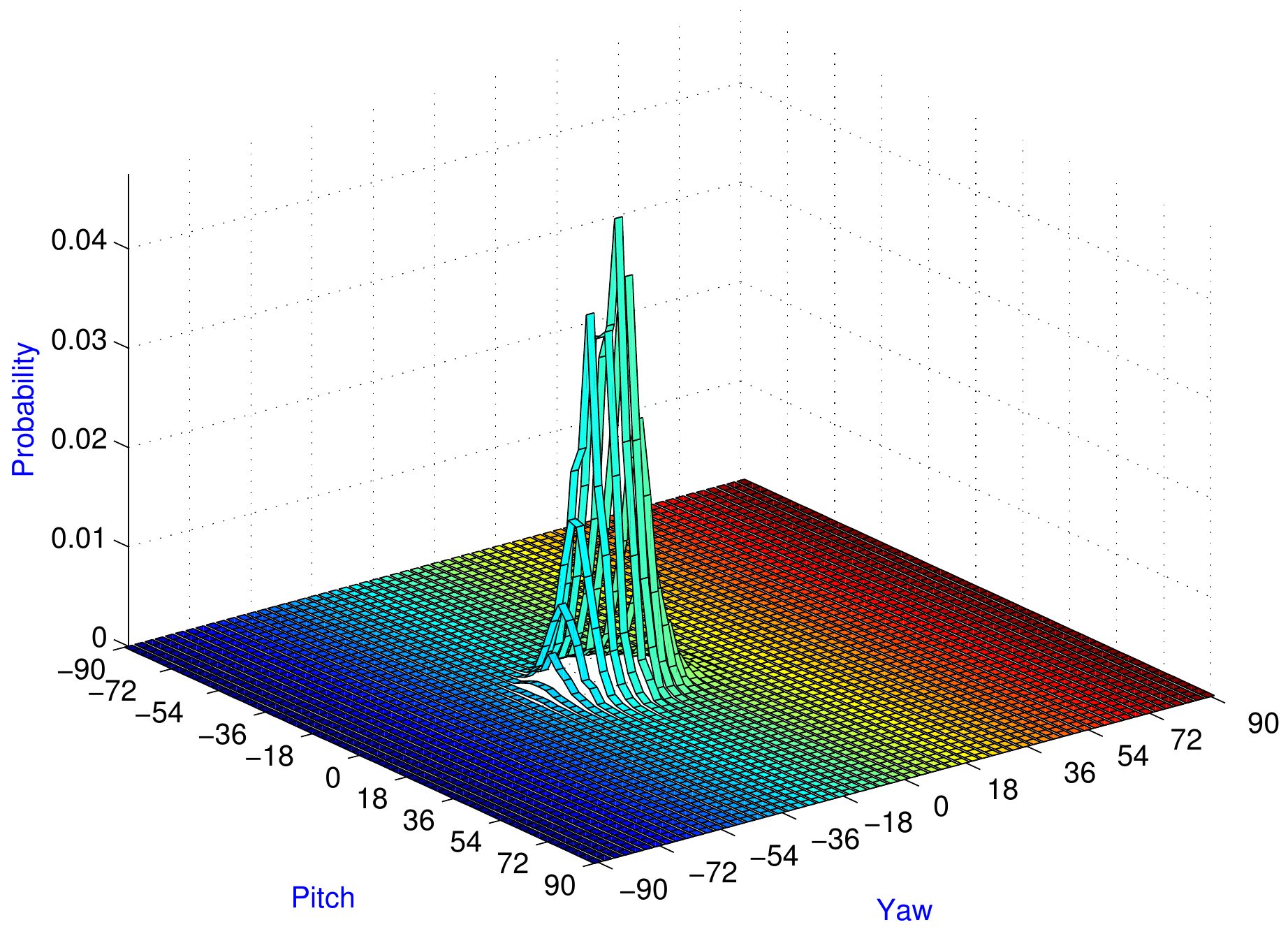}}
    \subfloat[\color{red}{(-3$^\circ$,-27$^\circ$)}]  {\includegraphics[width= 0.2\columnwidth]{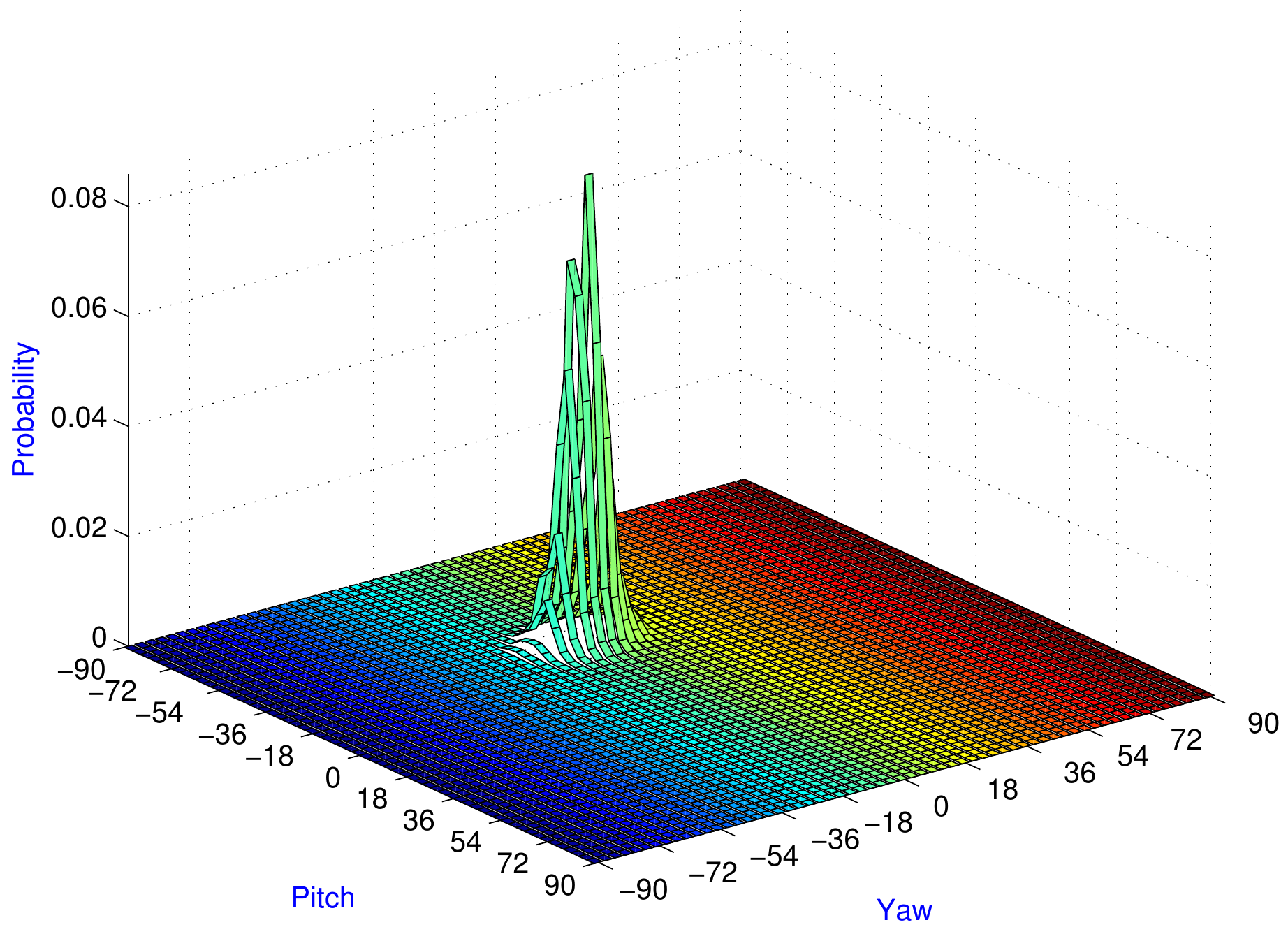}}
    \subfloat[\color{red}{(+27$^\circ$,+6$^\circ$)}]  {\includegraphics[width= 0.2\columnwidth]{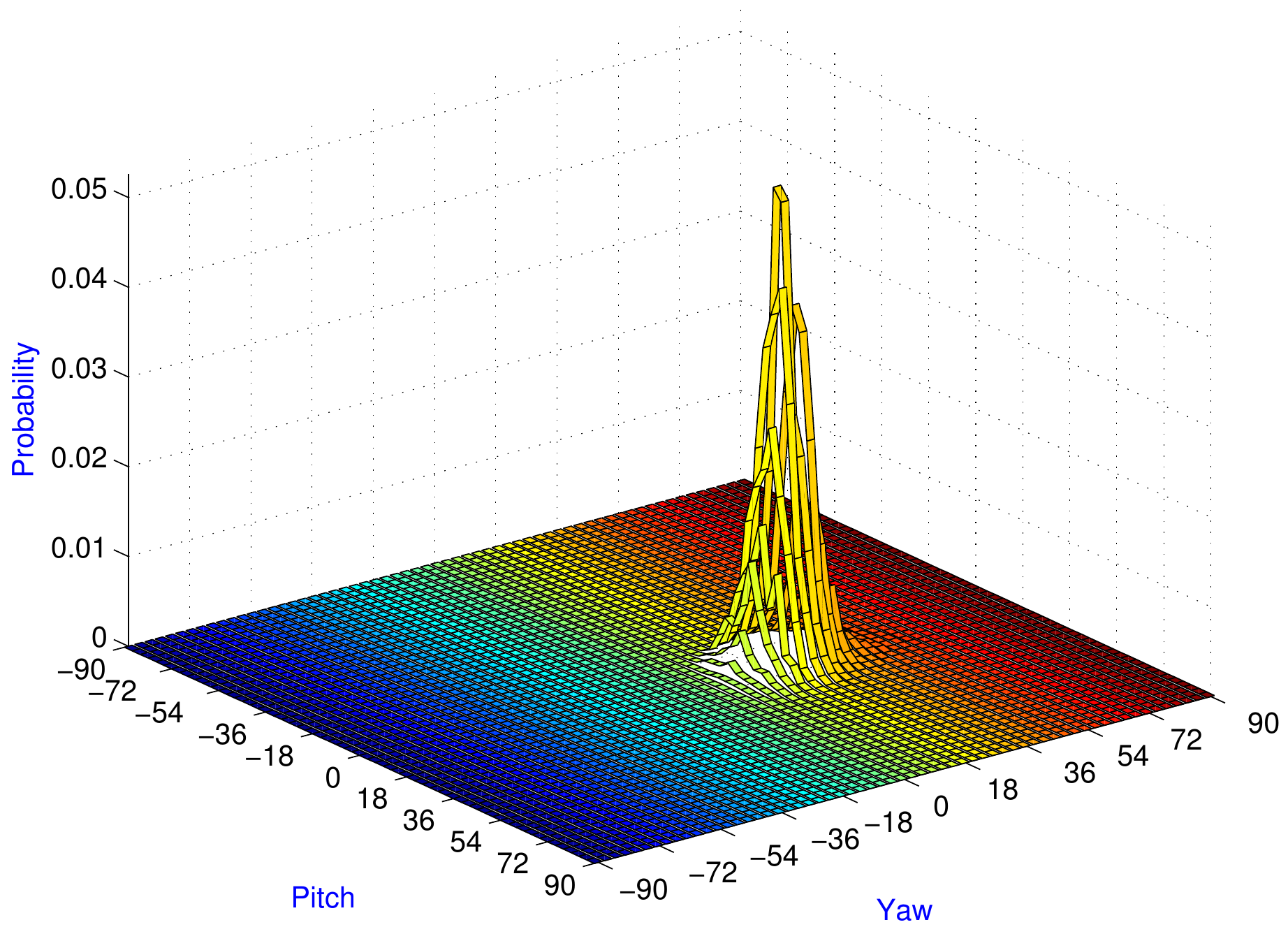}}
    \subfloat[\color{red}{(+6$^\circ$,-3$^\circ)$}]   {\includegraphics[width= 0.2\columnwidth]{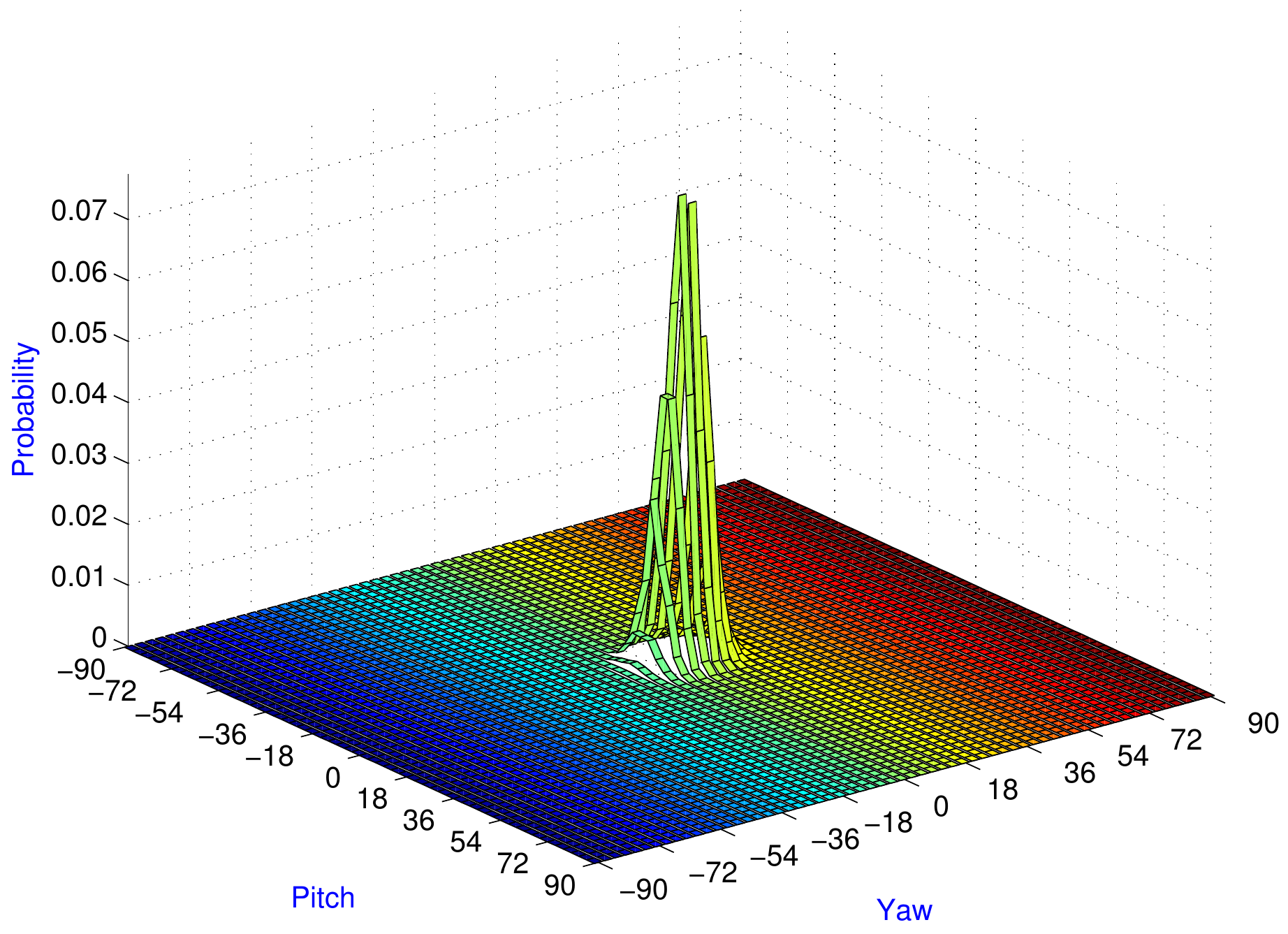}}
    \subfloat[\color{red}{(-39$^\circ$,+15$^\circ)$}] {\includegraphics[width= 0.2\columnwidth]{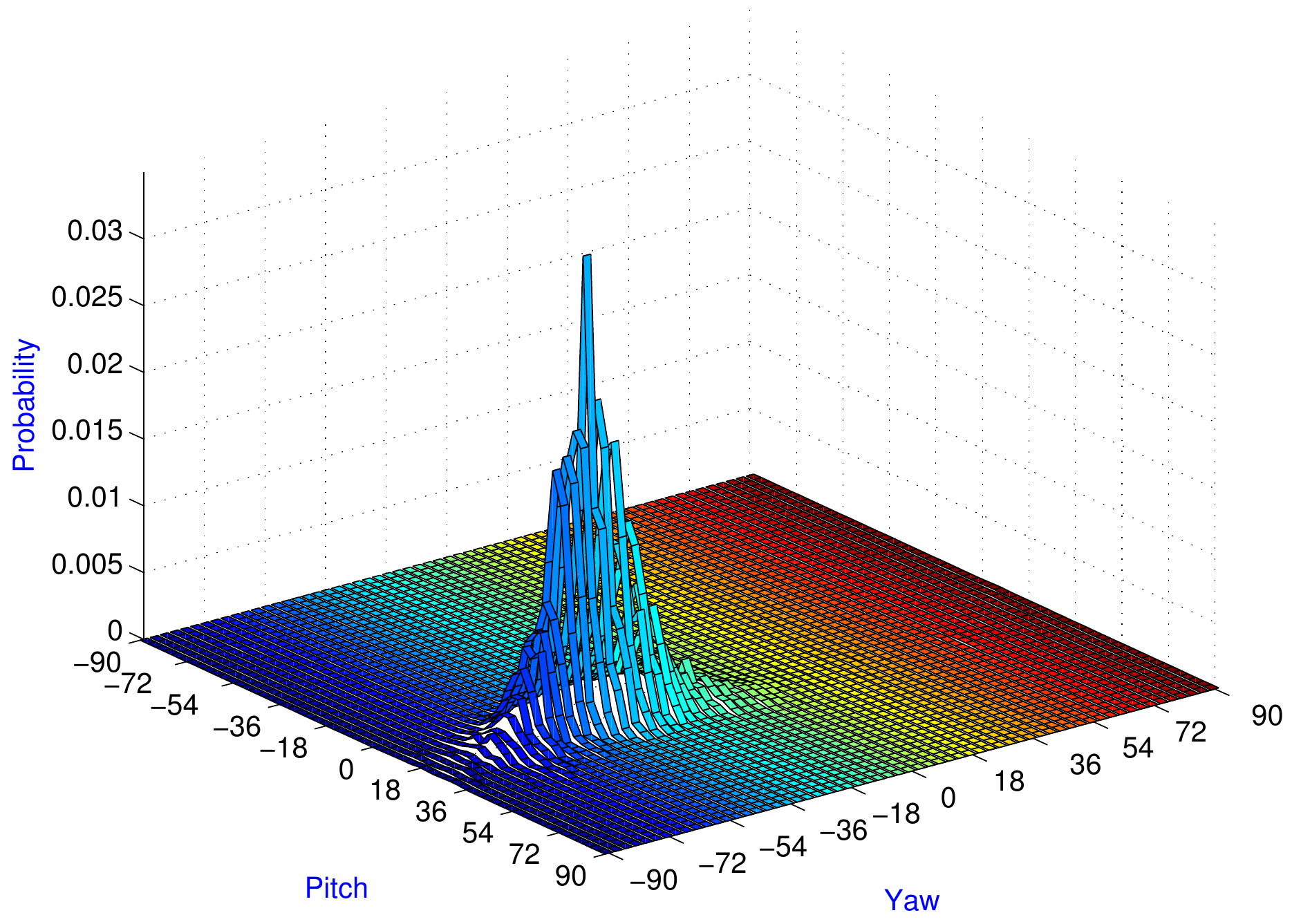}}
    \subfloat[\color{red}{(-87$^\circ$,0$^\circ$)}]   {\includegraphics[width= 0.2\columnwidth]{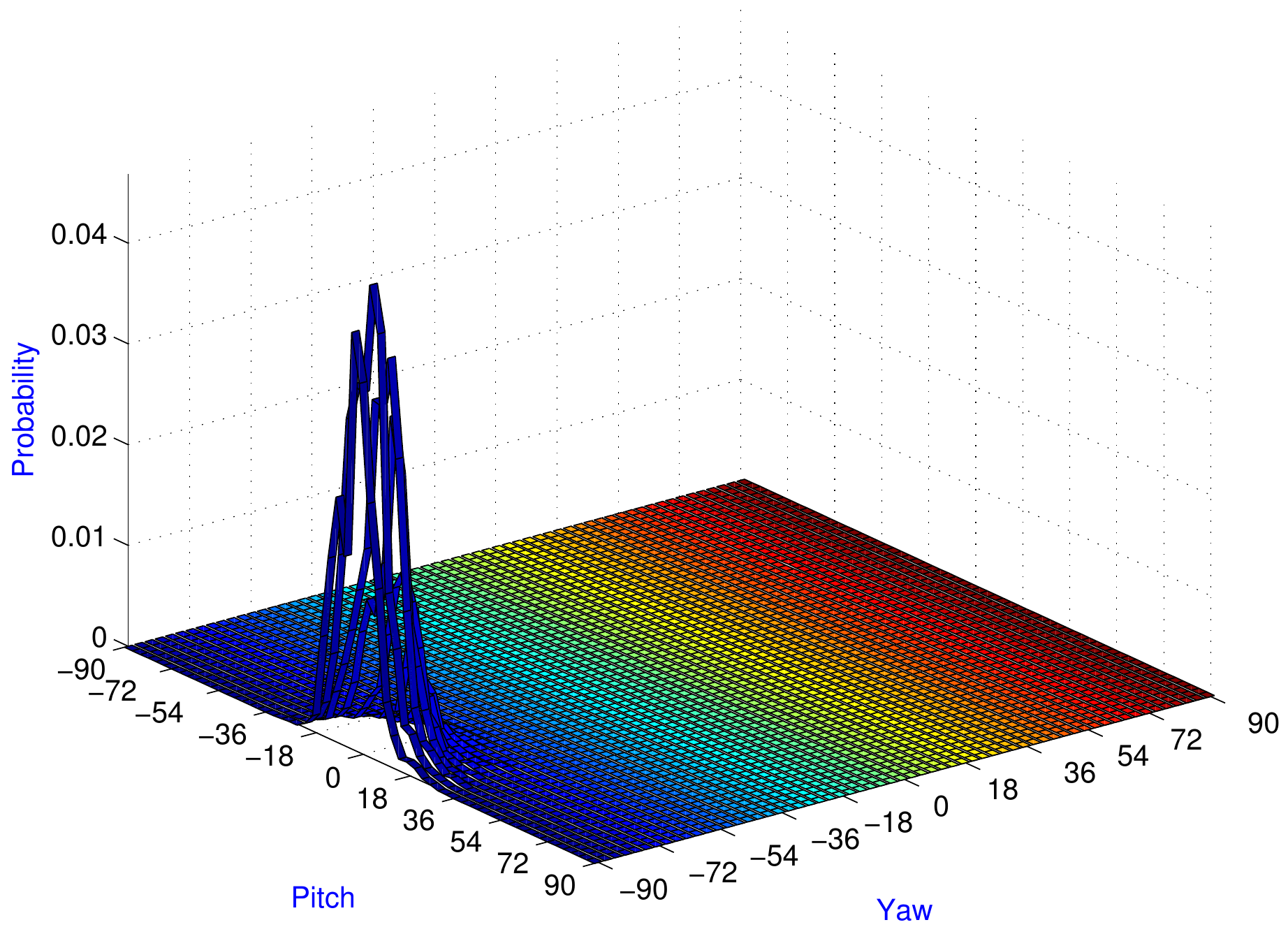}}
    \hspace{2pt} \vrule \hspace{2pt}
    \subfloat[\color{blue}{(-3$^\circ$,-12$^\circ$)}]{\includegraphics[width= 0.2\columnwidth]{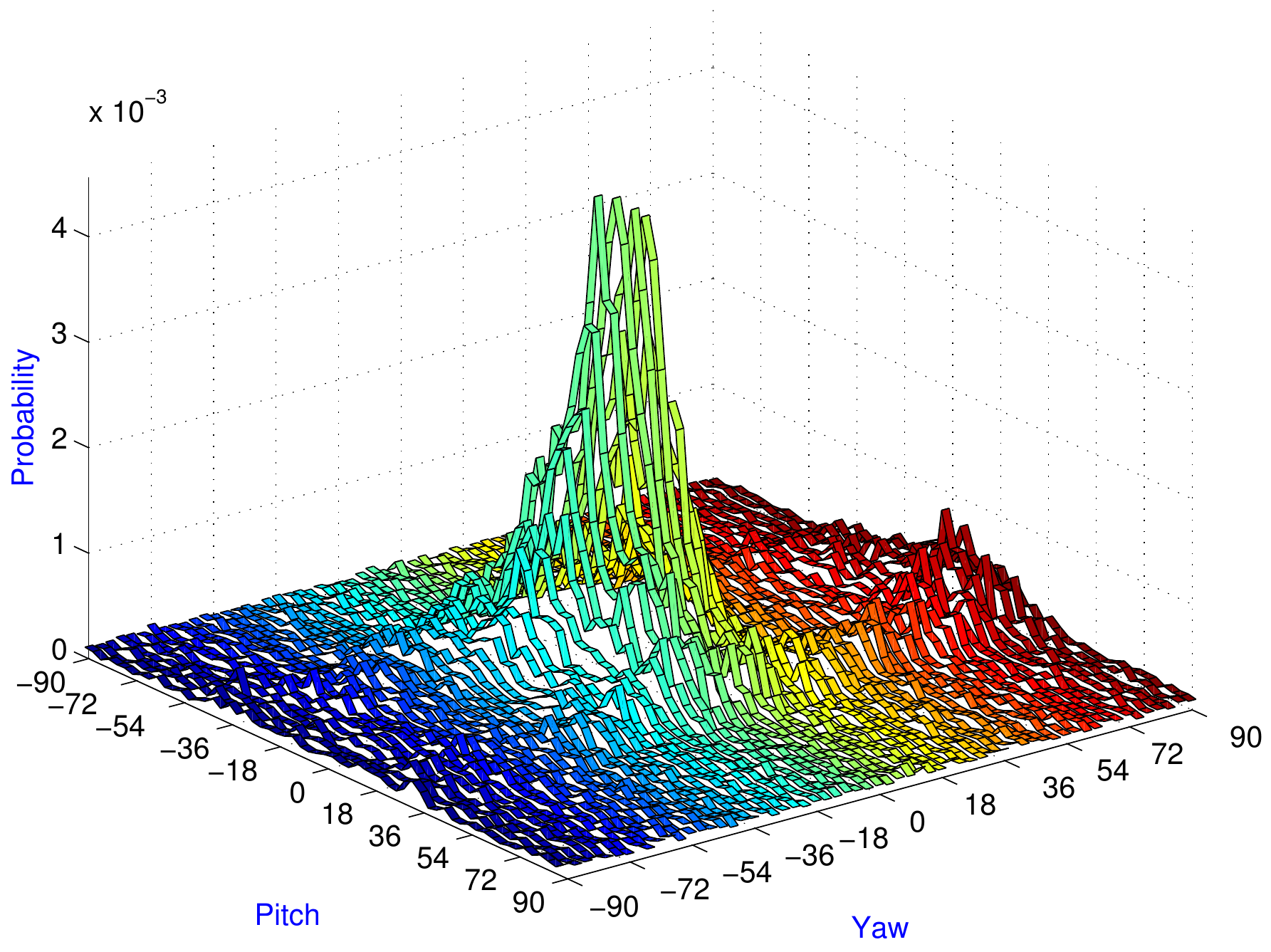}}
    \subfloat[\color{blue}{(+21$^\circ$,+18$^\circ)$}] {\includegraphics[width= 0.2\columnwidth]{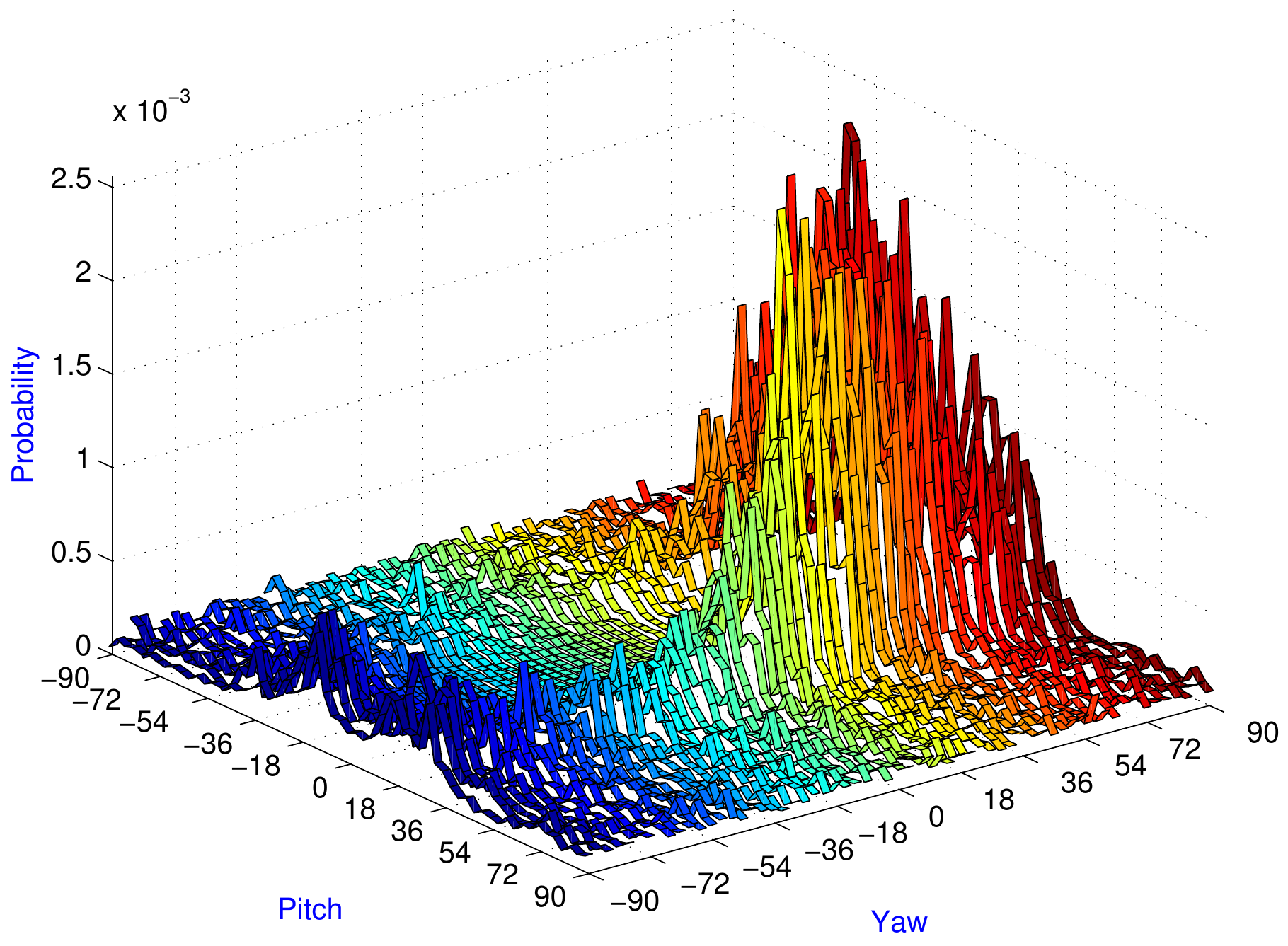}}
    \subfloat[\color{blue}{(+45$^\circ$,-15$^\circ$)}] {\includegraphics[width= 0.2\columnwidth]{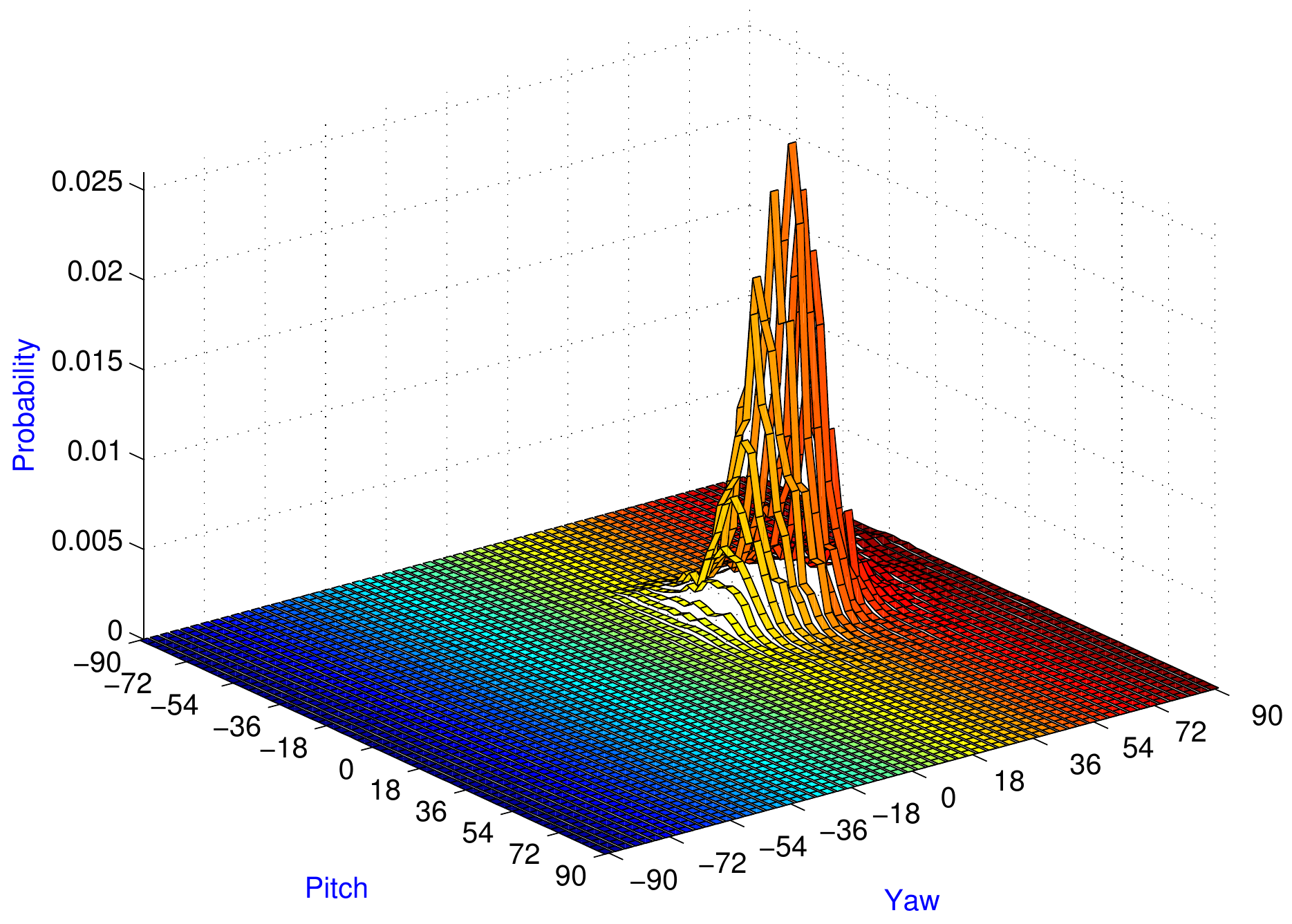}}
 \caption{Examples of face images and DLDL results. The first row shows ten cropped faces from the \emph{AFLW} dataset and their corresponding ground-truth labels (yaw angle, pitch angle). The second row shows their predicted label distributions and predicted head poses. The left seven columns are the good examples and the right three columns are the failure cases.}\label{fig:eg-pose}
\end{figure*}

\textbf{Results.} Tables~\ref{table:pose-po}, \ref{table:pose-bj} and~\ref{table:pose-af} show results on \emph{Pointing'04}, \emph{BJUT-3D} and \emph{AFLW}, respectively. \emph{Pointing'04} is small scale with only 2,790 images. We observe that BFGS-LDL (with hand-crafted features) has much lower MAE and much higher accuracy than deep learning methods C-ConvNet, R-ConvNet and ConvNet+LS. One reasonable conjecture is that C-ConvNet, R-ConvNet and ConvNet+LS are not well-learned with only small number of training images. DLDL, however, successfully learns the head pose. For example, its accuracy for pitch+yaw is 73.15$\%$ (and C-ConvNet is only 42.97$\%$). That is, DLDL is able to perform deep learning with few training images, while C-ConvNet R-ConvNet and ConvNet+LS have failed for this task.

On \emph{BJUT-3D} and \emph{AFLW} which have enough training data, we observe that many deep learning methods show higher performance than BFGS-LDL. DLDL achieves the best performance: it has much lower MAE and higher accuracy than other methods. Another observation is also worth mentioning. Although R-ConvNet is better than C-ConvNet when label is dense such as age estimation and head pose estimation on \emph{AFLW}, it is obviously worse than C-ConvNet on \emph{BJUT-3D} and \emph{pointing'04} for head pose estimation which have sparse labels. In other words, the performance of C-ConvNet and R-ConvNet are not very robust, while the proposed method consistently achieves excellent performance.

Fig.~\ref{fig:aflw-cs} shows the pitch+yaw CS curves on the \emph{AFLW} dataset. There is an obvious gap between DLDL and baseline methods at every error level. Fig.~\ref{fig:eg-pose} shows the predicted label distributions for different head poses on the \emph{AFLW} testing set using the DLDL model. Our approach can estimate head pose with low errors but may fail under some extreme conditions. It is noteworthy that DLDL may produce more incorrect estimations when both yaw and pitch are large~(\emph{e.g.}, $\pm90^\circ$). The reason might be that there are much fewer training examples for large angles than for other angles.

\subsection{Multi-label classification}
\textbf{Datasets.} We evaluate our approach for multi-label classification on the PASCAL VOC dataset~\cite{everingham2010pascal}: PASCAL \emph{VOC2007} and \emph{VOC2012}. There are 9,963 and 22,531 images in them, respectively. Each image is annotated with one or several labels, corresponding to 20 object categories. These images are divided into three subsets including \texttt{TRAIN}, \texttt{VAL} and \texttt{TEST} sets. We train on the \texttt{TRAINVAL} set and evaluate on the \texttt{TEST} set. The evaluation metric is average precision (AP) and mean average precision (mAP), complying with the PASCAL challenge protocols.

We denote our methods as Images-Fine-tuning-DLDL~(IF-DLDL) and Proposals-Fine-tuning-DLDL~(PF-DLDL) when ConvNets are fine-tuned by images and proposals of images, respectively. Details of these two variants are explained later in this section. We compare the proposed approaches with the following methods:
\begin{itemize}
\item \textbf{VGG+SVM~\cite{simonyan2015very}}. This method densely extracted 4,096 dimensional ConvNet features at the penultimate layer of VGG-Nets pre-trained on ImageNet. These features from different scales (smallest image side $Q\in \{256,\allowbreak 384, 512, 640, 768\}$) were aggregated by average pooling. Then, these averaged features from two networks (``Net-D" containing 16 layers and ``Net-E" containing 19 layers) were further fused by stacking. Finally, \cite{simonyan2015very} $\ell_2$ normalized the resulting image features and used these features to train a linear SVM classifier for multi-label classification.

\item \textbf{HCP~\cite{wei2014cnn}}. HCP proposed to solve the multi-label object recognition task by extracting object proposals from the images. The method used image label and square loss to fine-tune a pre-trained ConvNet. Then, BING~\cite{cheng2014bing} or EdgeBoxes~\cite{zitnick2014edge} was used to extract object proposals, which were used to fine-tune the ConvNet again. Finally, scores of these proposals were max-pooled to obtain the prediction. 

\item \textbf{Fev+Lv~\cite{yang2016exp}}. This approach transformed the multi-label object recognition problem into a multi-class multi-instance learning problem. Two views~(label view and feature view) were extracted for each proposal of images. Then, these two views were encoded by a Fisher vector for each image.

\item \textbf{IF-VGG-$\ell_2$} and \textbf{IF-VGG-KL}. We fine-tune the VGG-Nets with square loss and multi-label cross-entropy loss~\cite{gong2013deep} and use them as our IF-DLDL's baselines. They are trained using the same setting.
\end{itemize}

\begin{table}
	\centering
	\caption{Single model classification mAP (in $\%$) on \emph{VOC2007} (\texttt{TRAINVAL}/\texttt{TEST}). The * sign indicates ground-truth bounding box information was used during training.}\label{table:singlevoc07}
	\small
	\begin{tabular}{|@{\,}l@{\,}|@{\,}l@{\,}| *{4}{@{\,}c@{\,}}|}
		\hline
		 \multirow{2}{*}{Methods} &\multirow{2}{*}{Description}        &Net-D   &Net-D &Net-E  &Net-E \\
		         &   &Max  &Avg &Max  &Avg  \\
        \hline\hline
        &Fev+Lv-20-VD*~\cite{yang2016exp}    &90.6 &-    &-    &-     \\
        &HCP-VGG~\cite{wei2015hcp}             &90.9 &-    &-    &-     \\
        \hline
        \multirow{3}{*}{Baselines} 
        &VGG+SVM~\cite{simonyan2015very}         &89.3 &-    &89.3 &-     \\
        &IF-VGG-$\ell_2$ &89.8 &89.5 &89.7 &89.8  \\
        &IF-VGG-KL       &90.0 &90.3 &90.3 &90.2  \\
        \hline
        \multirow{2}{*}{Ours}
        &IF-DLDL         &90.1 &90.5 &90.6 &90.7  \\
        &PF-DLDL         &\textbf{92.3} &\textbf{92.1} &\textbf{92.5} &\textbf{92.2}  \\
        \hline
	\end{tabular}
\end{table}

\begin{table*}
 \centering
 \caption{Comparisons of the classification results (in $\%$) of state-of-the-art approaches on \emph{VOC2007} (\texttt{TRAINVAL}/\texttt{TEST}). * indicates methods using ground-truth bounding box information for training.}\label{table:voc07}
 \footnotesize
 \begin{tabular}{|@{\,}l@{\,}|@{\,}l@{\,}| *{20}{@{\,}c@{\,}}|@{\,}c@{\,}|}
  \hline
  Methods &Description &aero &bike 	&bird &boat &bottle &bus &car &cat &chair &cow &table &dog &horse &mbike &person &plant &sheep &sofa &train &tv &mAP\\
  \hline\hline
  &AGS*~\cite{dong2013sub} &82.2 &83.0 &58.4 &76.1 &56.4 &77.5 &88.8 &69.1 &62.2 &61.8 &64.2 &51.3 &85.4 &80.2 &91.1 &48.1 &61.7 &67.7 &86.3 &70.9 &71.1\\
  &AMM*~\cite{song2011contextualizing} &84.5 &81.5 &65.0 &71.4 &52.2 &76.2 &87.2 &68.5 &63.8 &55.8 &65.8 &55.6 &84.8 &77.0 &91.1 &55.2 &60.0 &69.7 &83.6 &77.0 &71.3\\
  &HCP-2000C~\cite{wei2014cnn} &96.0 &92.1 &93.7 &93.4 &58.7 &84.0 &93.4 &92.0 &62.8 &89.1 &76.3 &91.4 &95.0 &87.8 &93.1 &69.9 &90.3 &68.0 &96.8 &80.6 &85.2\\
  &Fev+Lv-20-VD*~\cite{yang2016exp} &97.9 &97.0 &96.6 &94.6 &73.6 &93.9 &96.5 &95.5 &73.7 &90.3 &82.8 &95.4 &\textbf{97.7} &95.9 &\textbf{98.6} &77.6 &88.7 &78.0 &98.3 &89.0 &90.6\\
  &HCP-VGG~\cite{wei2015hcp} &98.6 &97.1 &98.0 &95.6 &75.3 &94.7 &95.8 &97.3 &73.1 &90.2 &80.0 &97.3 &96.1 &94.9 &96.3 &78.3 &94.7 &76.2 &97.9 &91.5 &90.9\\
  \hline
  \multirow{3}{*}{Baselines}
  &VGG+SVM~\cite{simonyan2015very} &98.9 &95.0 	&96.8 &95.4 &69.7 &90.4	&93.5 &96.0	&74.2 &86.6 &87.8 &96.0 &96.3 &93.1 &97.2 &70.0 &92.1 &80.3 &98.1 &87.0 &89.7\\
  &IF-VGG-$\ell_2$ &98.9 &95.7 &97.3 &95.5 &65.0 &92.8 &93.7 &97.1 &74.2 &90.8 &87.0 &97.1 &97.1 &93.8 &97.0 &70.8 &94.3 &77.8 &98.0 &86.4 &90.0\\
  &IF-VGG-KL &99.1 &95.5	&97.4	&94.9	&68.1	&92.7	&94.3	&97.0	&75.7	&90.3	&89.0	&97.0	&97.6	&94.6	&97.2	&76.3	&93.8	&80.1	&98.2	&87.9	&90.8\\
  \hline\hline
  \multirow{2}{*}{Ours}
  &IF-DLDL     &99.1 &95.8 &97.4 &95.3 &69.2 &93.3 &94.5 &96.6 &76.1 &90.4 &\textbf{89.0} &97.1 &\textbf{97.7} &94.5 &97.7 &76.1 &93.6 &81.9 &98.2 &89.1 &91.1\\
  &PF-DLDL     &\textbf{99.3} 	&\textbf{97.6} 	&\textbf{98.3} 	&\textbf{97.0} 	&\textbf{79.0} 	&\textbf{95.7} 	&\textbf{97.0} 	&\textbf{97.9} 	&\textbf{81.8} 	&\textbf{93.3} 	&88.2 	&\textbf{98.1} 	&96.9 	&\textbf{96.5} 	&98.4 &\textbf{84.8} 	&\textbf{94.9} 	&\textbf{82.7} 	&\textbf{98.5} 	&\textbf{92.8} 	&\textbf{93.4} \\
  \hline
 \end{tabular}
\end{table*}

\begin{table*}
 \centering{
 \caption{Comparisons of the classification results (in $\%$) of state-of-the-art approaches on \emph{VOC2012} (\texttt{TRAINVAL}/\texttt{TEST}). * indicates methods using ground-truth bounding box information for training.}\label{table:voc12}
 \footnotesize
 \begin{tabular}{|@{\,}l@{\,}|@{\,}l@{\,}| *{20}{@{\,}c@{\,}}|@{\,}c@{\,}|}
  \hline
  Methods &Description &aero &bike 	&bird &boat &bottle &bus &car &cat &chair &cow &table &dog &horse &mbike &person &plant &sheep &sofa &train &tv &mAP\\
  \hline\hline
  &NUS-PSL*\cite{dong2013sub} &97.3 &84.2 &80.8 &85.3 &60.8 &89.9 &86.8 &89.3 &75.4 &77.8 &75.1 &83.0 &87.5 &90.1 &95.0 &57.8 &79.2 &73.4 &94.5 &80.7 &82.2\\
  &PRE-1512*\cite{oquab2014learning} &94.6 &82.9 &88.2 &84.1 &60.3 &89.0 &84.4 &90.7 &72.1 &86.8 &69.0 &92.1 &93.4 &88.6 &96.1 &64.3 &86.6 &62.3 &91.1 &79.8 &82.8\\
  &HCP-2000C~\cite{wei2014cnn} &97.5 &84.3 &93.0 &89.4 &62.5 &90.2 &84.6 &94.8 &69.7 &90.2 &74.1 &93.4 &93.7 &88.8 &93.3 &59.7 &90.3 &61.8 &94.4 &78.0 &84.2\\
  &Fev+Lv-20-VD*~\cite{yang2016exp} &98.4 &92.8 &93.4 &90.7 &74.9 &93.2 &90.2 &96.1 &78.2 &89.8 &80.6 &95.7 &96.1 &95.3 &97.5 &73.1 &91.2 &75.4 &97.0 &88.2 &89.4\\
  &HCP-VGG~\cite{wei2015hcp} &99.1 &92.8 &97.4 &94.4 &79.9 &93.6 &89.8 &\textbf{98.2} &78.2 &94.9 &79.8 &97.8 &97.0 &93.8 &96.4 &74.3 &94.7 &71.9 &96.7 &88.6 &90.5\\
  \hline
  \multirow{3}{*}{Baselines}
  &VGG+SVM~\cite{simonyan2015very} &99.0 &89.1 &96.0 &94.1 &74.1 &92.2 &85.3 &97.9 &79.9 &92.0 &83.7 &97.5 &96.5 &94.7 &97.1 &63.7 &93.6 &75.2 &97.4 &87.8 &89.3\\
  &IF-VGG-$\ell_2$  &98.9 &88.4 &96.7 &93.4 &70.7 &92.3 &85.8 &97.7 &77.3 &94.2 &81.2 &97.4 &96.8 &93.7 &96.7 &62.2 &94.1 &70.7 &96.9 &85.8  &88.6\\
  &IF-VGG-KL  &99.0 &89.9 &96.6 &93.7 &74.0 &93.2 &87.3 &97.5 &78.5 &94.7 &83.1 &97.1 &96.9 &94.0 &96.6 &66.9 &94.5 &75.9 &97.4 &87.7  &89.7\\
  \hline\hline
  \multirow{2}{*}{Ours}
  &IF-DLDL    &99.0 &89.7 &96.6 &94.1 &74.8 &93.1 &87.8 &97.6 &79.3 &94.3 &83.4 &97.2 &96.9 &94.0 &97.3 &67.8 &94.2 &76.5 &97.4 &87.8 &89.9\\
  &PF-DLDL &\textbf{99.5} 	&\textbf{94.1} 	&\textbf{97.9} 	&\textbf{95.9} 	&\textbf{81.0} 	&\textbf{94.8} 	&\textbf{93.1} 	&\textbf{98.2} 	&\textbf{82.4} 	&\textbf{96.1} 	&\textbf{84.0} 	&\textbf{98.0} 	&\textbf{97.8} 	&\textbf{95.7} 	&\textbf{97.7} &\textbf{78.9} 	&\textbf{95.5} 	&\textbf{78.0} 	&\textbf{97.8} 	&\textbf{92.2} 	&\textbf{92.4} \\
  \hline
 \end{tabular}}
\end{table*}

\textbf{Implementation details.}
 According to the ground-truth labels, we set different probabilities for all possible labels on PASCAL VOC dataset. In our experiments,
 $p_P= 1$, $p_D =0.3$, $p_N=0$. Finally, similar to label smoothing, a uniform distribution $u_i = \epsilon/20$ is added to $\vec y$, where $\epsilon=0.01$.

\textbf{IF-DLDL}. Following~\cite{simonyan2015very}, each training image is individually rescaled by randomly sampling in the range [256, 512]. We randomly crop $256 \times 256$ patches from these resized images. We also adjust the pooling kernel in the \texttt{pool5} layer from $3\times 3$ to $4 \times 4$. Max-pooling and Avg-pooling are used at \texttt{pool5} to train two ConvNets. We obtain four ConvNet models thought fine-tuning ``Net-D" and ``Net-E". At the prediction stage, the smaller side of each image is scaled to a fixed length $Q \in\{256,320,384,448,512\}$. Each scaled image is fed to the fine-tuned ConvNets to obtain the 20-dim probability outputs. These probability outputs from different scales and different models are averaged to form the final prediction.

\textbf{PF-DLDL}. Following~\cite{wei2015hcp}, we further fine-tune IF-DLDL models with proposals of images to boost performance. For each training image, we employ EdgeBoxes~\cite{zitnick2014edge} to produce a set of proposal bounding boxes which are grouped into $m$ clusters by the normalized cut algorithm~\cite{shi2000normalized}. For each cluster, the top $k$ proposals with higher predictive scores generated by EdgeBoxes are resized into square shapes~(\emph{i.e.}, $256\times 256$). As a result, we can obtain $mk$ proposals for an image. Finally, these $mk$ resized proposals are fed into a fine-tuned IF-DLDL model to obtain prediction scores and these scores are fused by max-pooling to form the prediction distribution of the image. This process can be learned by using an end-to-end way. In our implementation, we set $m=15, k=1$ and $m=15, k=30$ at the training and the prediction stage, respectively.
Similar to IF-DLDL, we also average fuse prediction scores of different models to generate the final prediction.

\textbf{Results.}
In Table~\ref{table:singlevoc07}, we compare single model results (average AP of all classes) on \emph{VOC2007}. Our PF-DLDL defeats all the other methods. Compared with Fev+Lv~\cite{yang2016exp}, 1.7\% improvement can be achieved by PF-DLDL even without using the bounding box annotation. Compared with HCP-VGG~\cite{wei2015hcp}, our PF-DLDL can achieve 92.3\% mAP, which is significantly higher than their 90.9\%. This further indicates that it is very important to learn a label distribution.

Table~\ref{table:voc07} and~\ref{table:voc12} report details of all experimental results on \emph{VOC2007} and \emph{VOC2012}, respectively. It can be seen that IF-DLDL outperforms IF-VGG-$\ell_2$ by 1.1\% for \emph{VOC2007} and 1.3\% for \emph{VOC2012}, which indicates that the KL loss function is more suitable than $\ell_2$ loss for measuring the similarity of two label distributions. Furthermore, IF-DLDL improves IF-VGG-KL for about 0.2--0.3 points in mAP, which suggests that learning a label distribution is beneficial. More importantly, PF-DLDL can achieve 93.4\% for \emph{VOC2007} and 92.4\% for \emph{VOC2012} in mAP when we average fuse output scores of four PF-DLDL models.

Our framework shows good performance especially for scene categories such as ``chair", `table" and ``sofa". Although PF-DLDL significantly outperforms IF-DLDL in mAP, PF-DLDL has higher computational cost than IF-DLDL on both training and testing stages. Since IF-DLDL does not need region proposals or bounding box information, it may be effectively and efficiently implemented for practical multi-label application such as multi-label image retrieval~\cite{lai2016instance}. It is also possible that by adopting new techniques (such as the region proposal method using gated unit in~\cite{zhao2016regional}, which has higher accuracy that ours on VOC tasks), the accuracy of our DLDL methods can be further improved.

\begin{figure*}
\setlength{\tempwidth}{.15\linewidth}
\settoheight{\tempheight}{\includegraphics[width=\tempwidth]{example-image-a}}%
\centering
\subfloat[ChaLearn]{
   \begin{tabular}{c}
    \rowname{\footnotesize{BFGS-LDL}}\includegraphics[height=0.28\columnwidth,keepaspectratio]{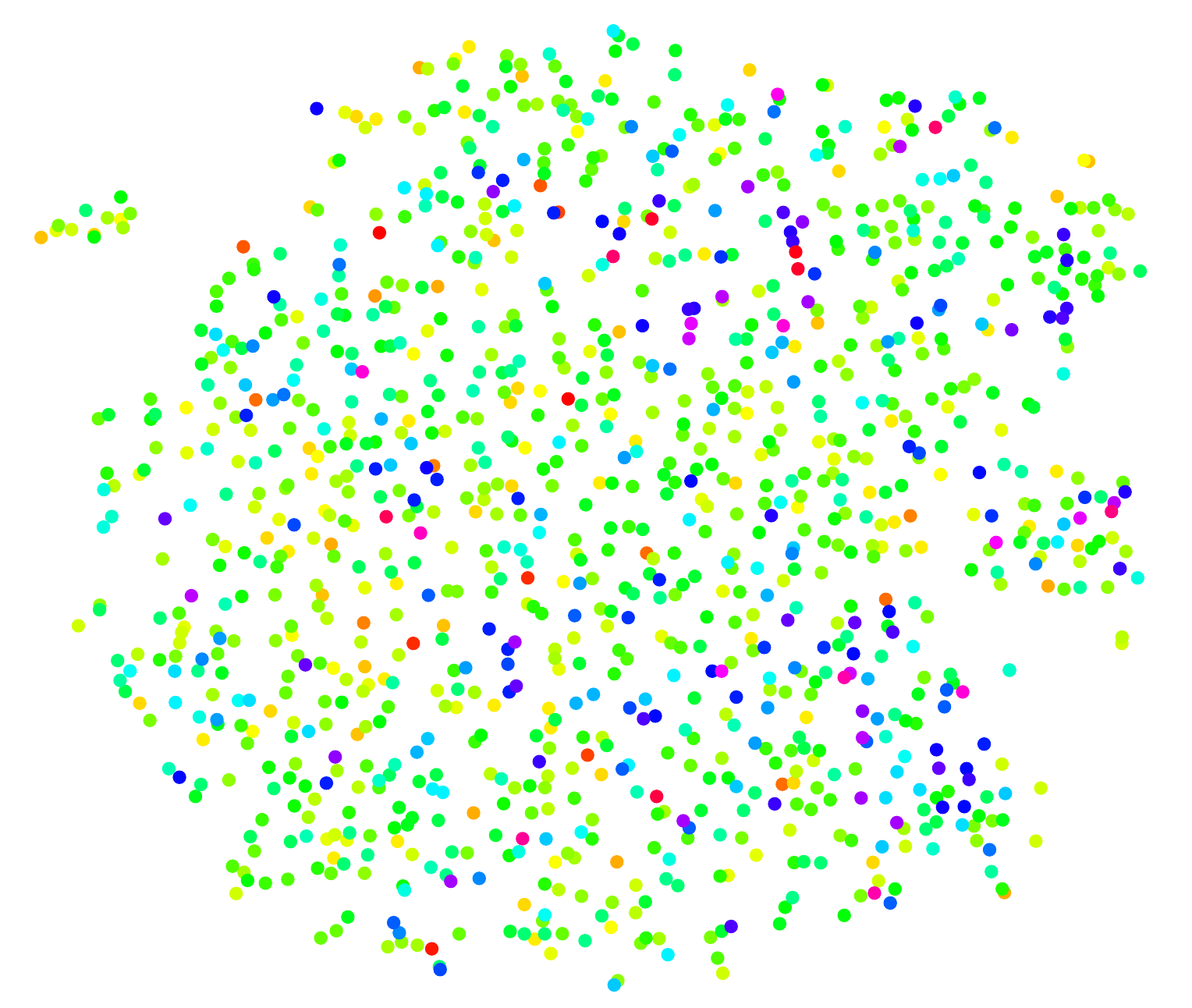} \\
    \rowname{\footnotesize{DLDL}}\includegraphics[height=0.28\columnwidth,keepaspectratio]{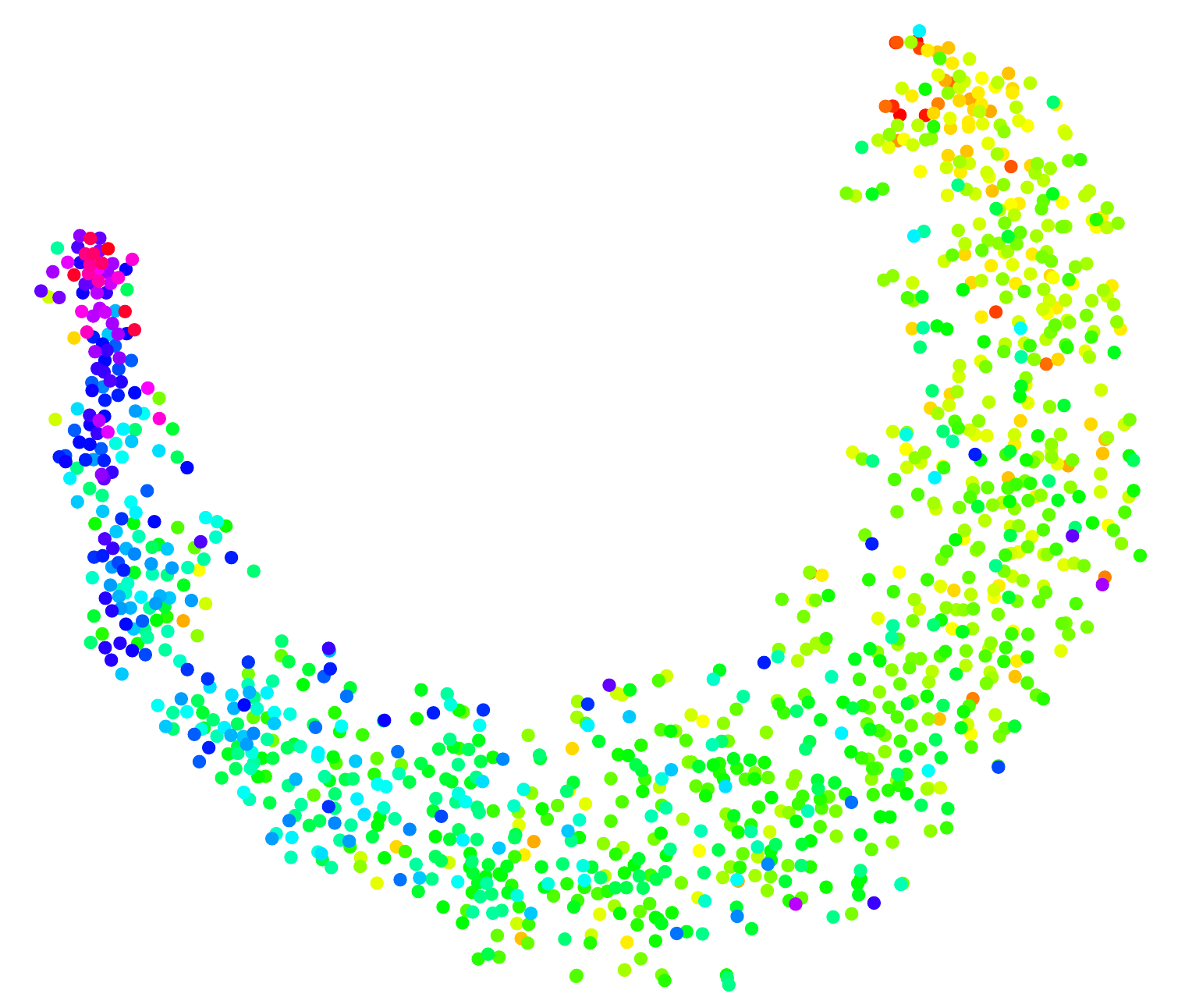}\\
   \end{tabular}
   }
\subfloat[Morph]{   
    \begin{tabular}{c}
    \includegraphics[height=0.28\columnwidth,keepaspectratio]{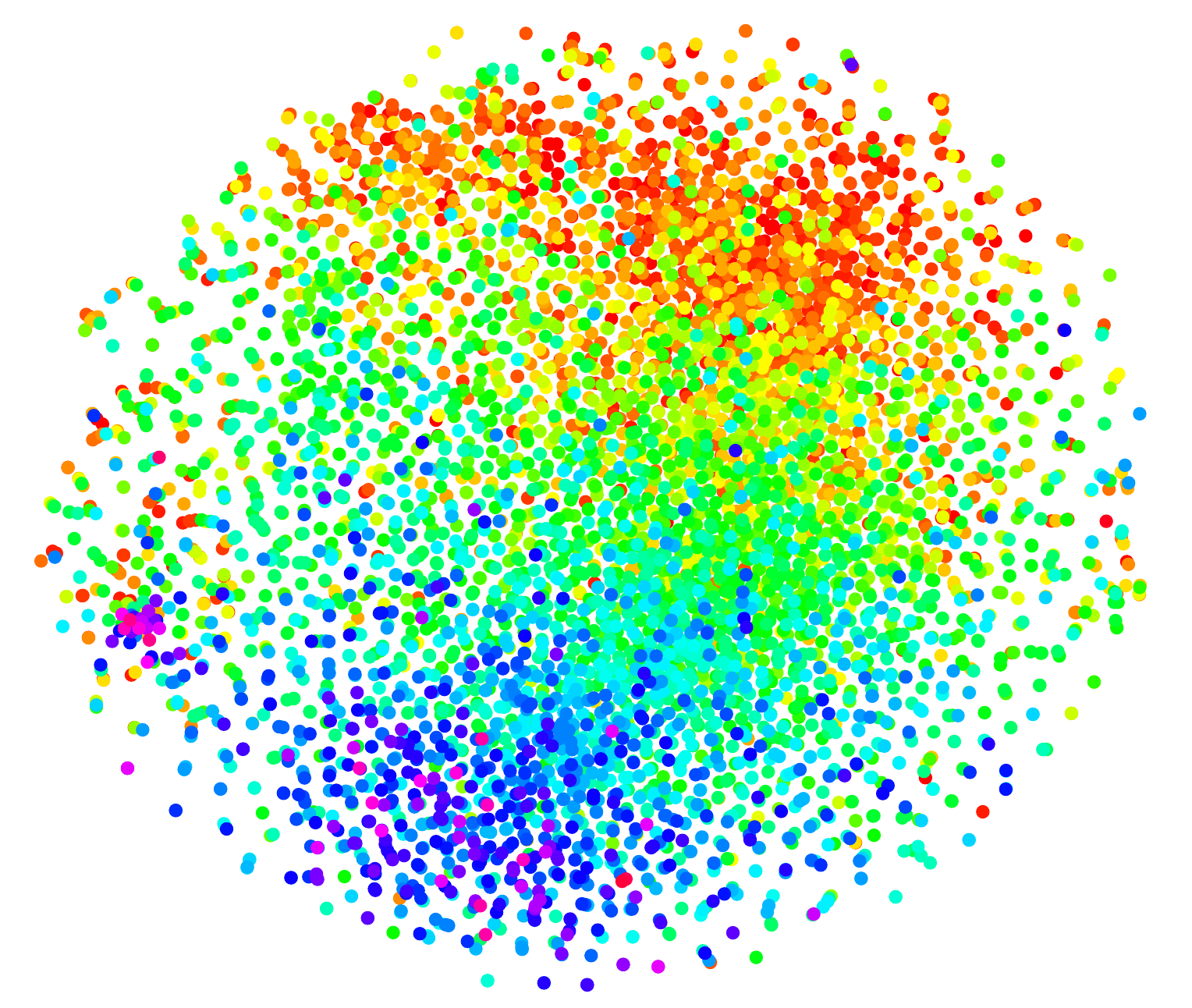} \\
    \includegraphics[height=0.28\columnwidth,keepaspectratio]{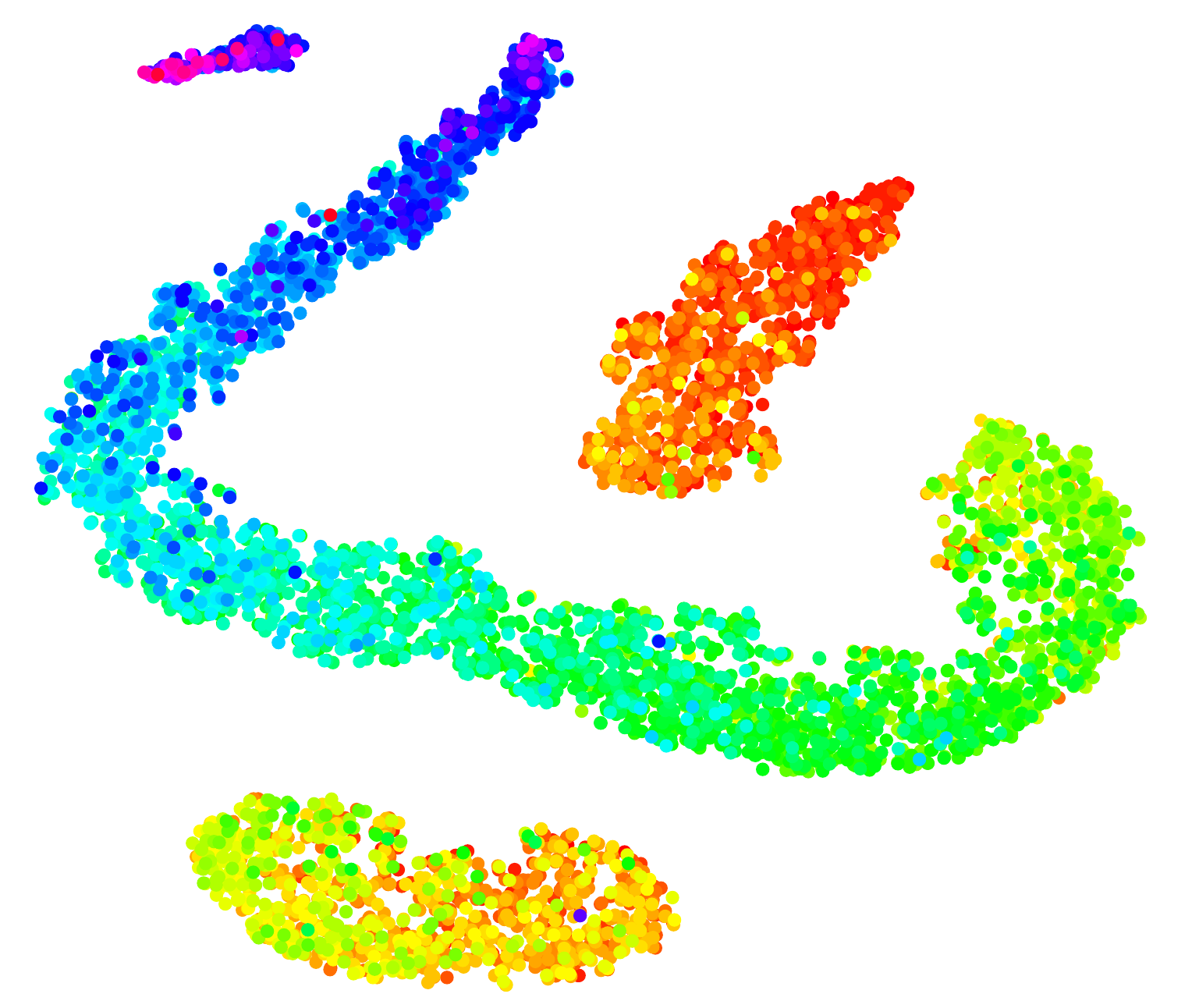}\\
   \end{tabular}
  }
\subfloat[Pointing'04]{   
    \begin{tabular}{c}
    \includegraphics[height=0.28\columnwidth,keepaspectratio]{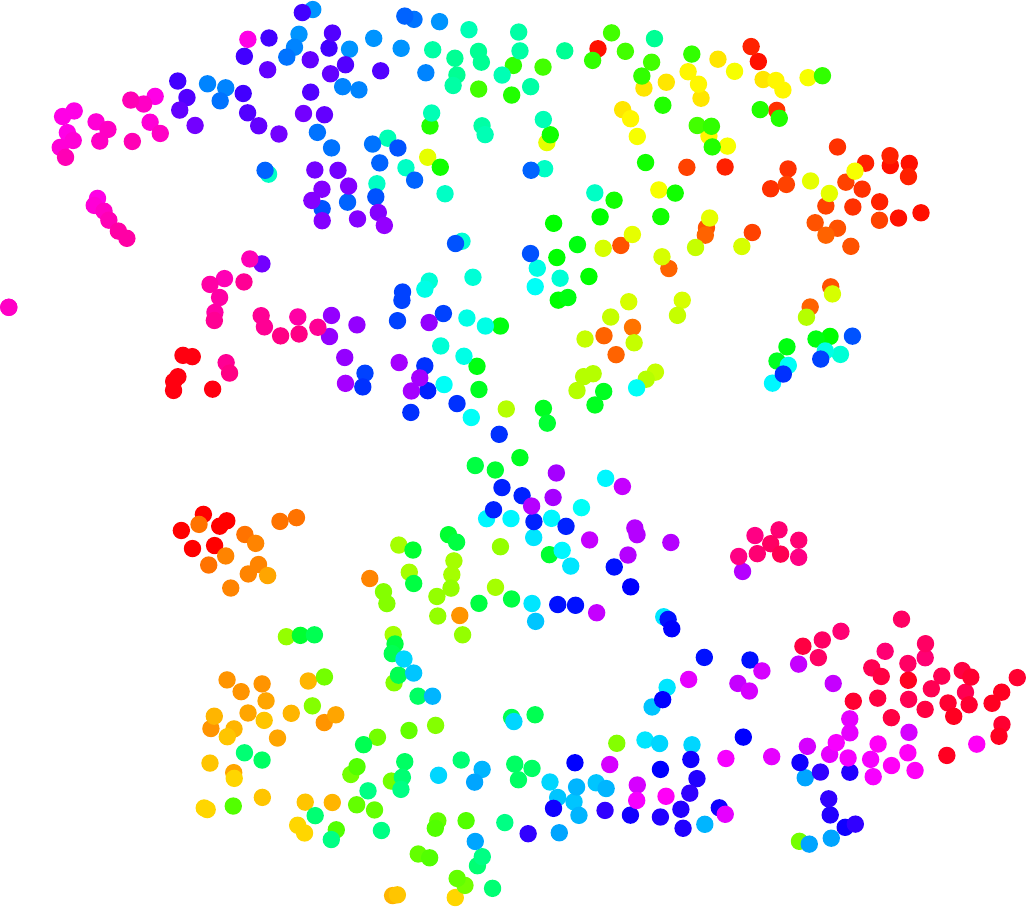} \\
    \includegraphics[height=0.28\columnwidth,keepaspectratio]{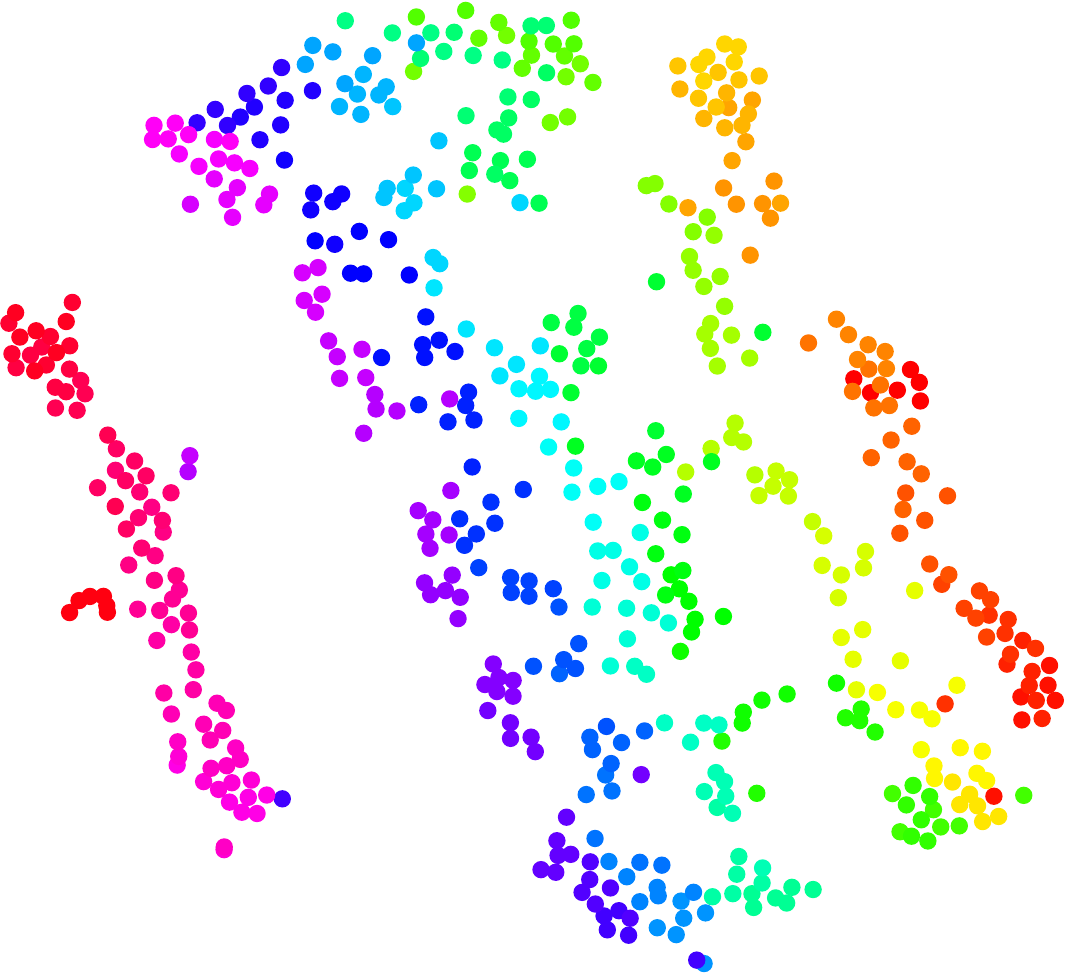}\\
   \end{tabular}
  }      
\subfloat[AFLW pitch]{   
    \begin{tabular}{c}
    \includegraphics[height=0.28\columnwidth,keepaspectratio]{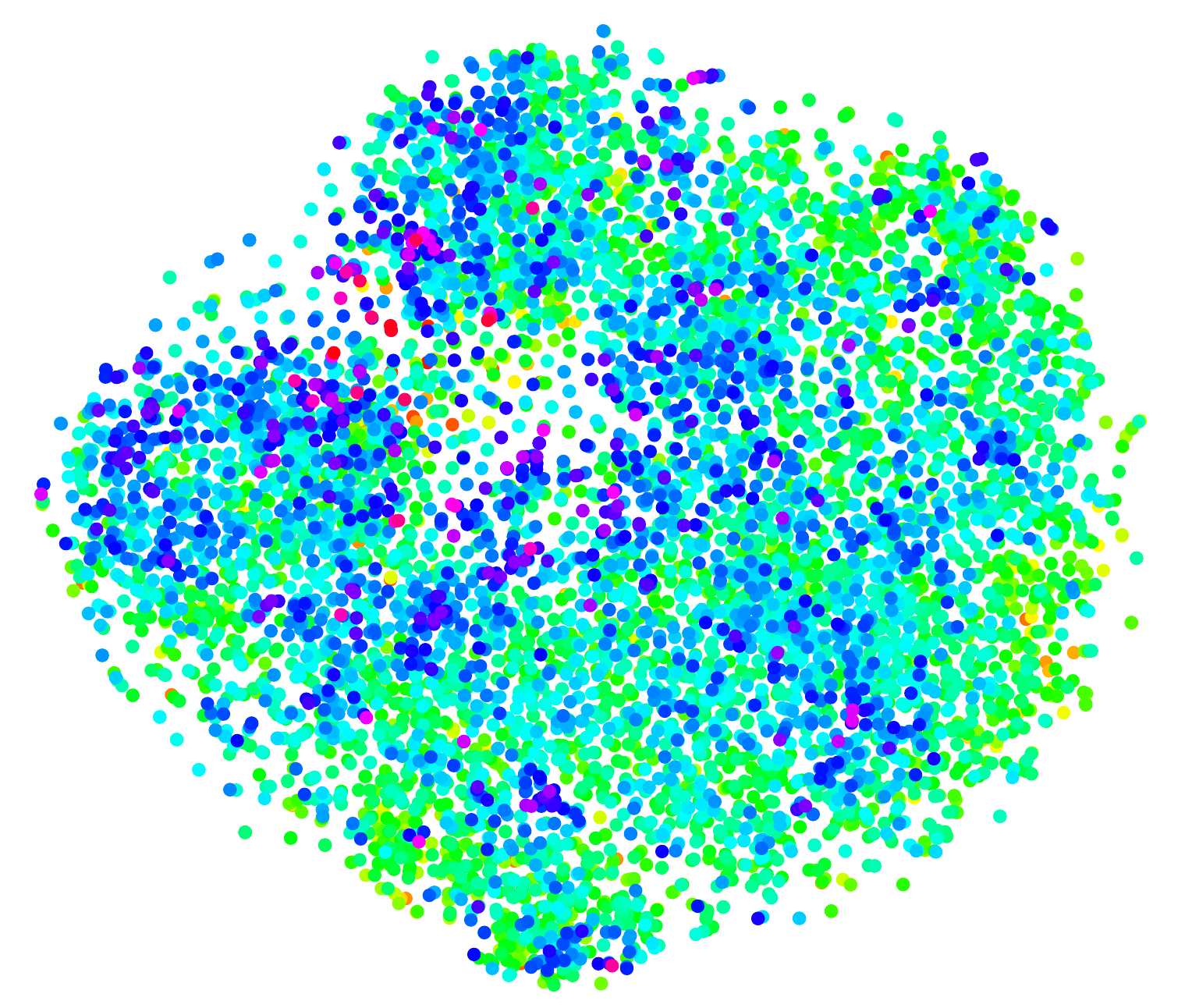} \\
    \includegraphics[height=0.28\columnwidth,keepaspectratio]{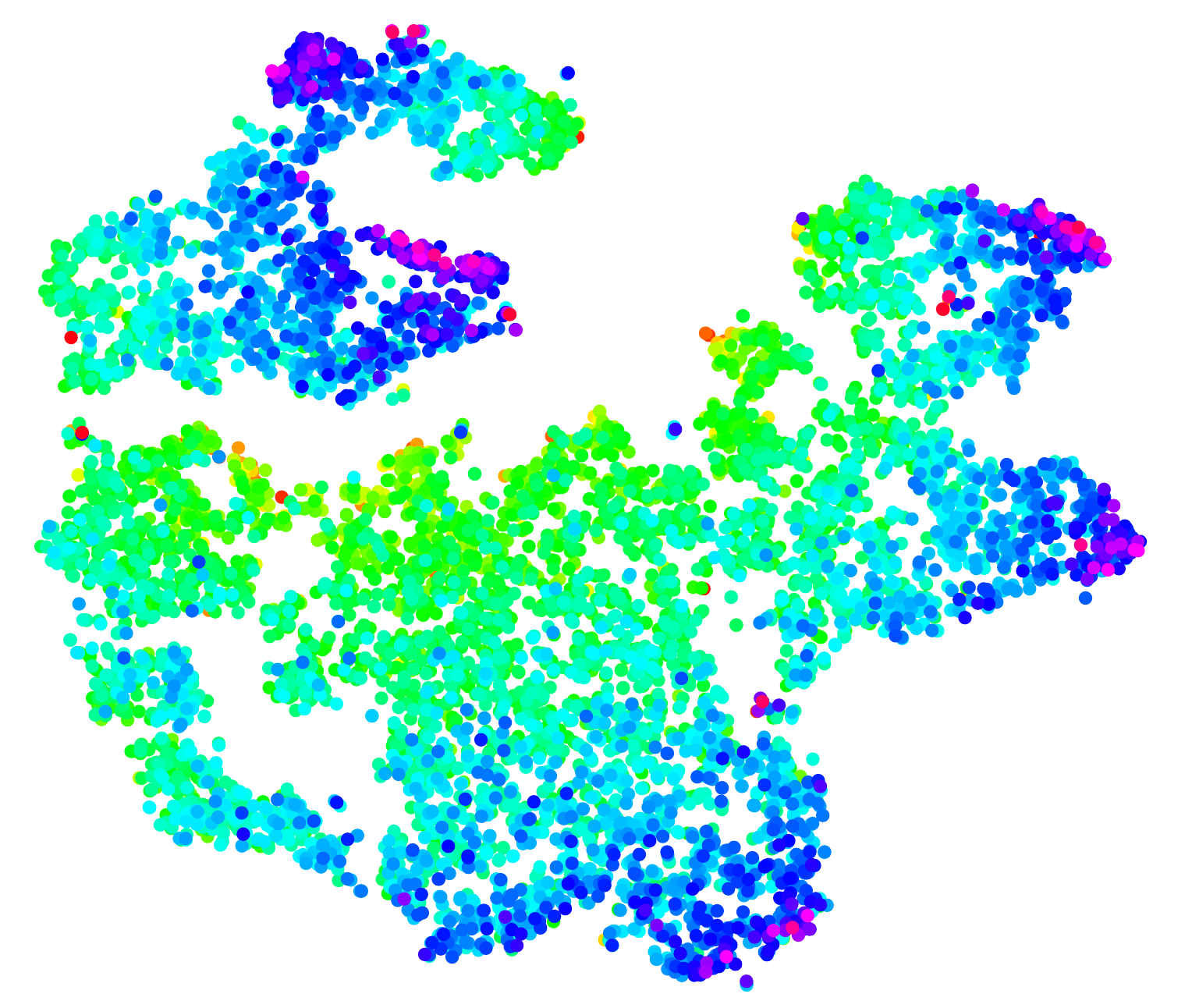}\\
   \end{tabular}
  }
\subfloat[AFLW yaw ]{   
    \begin{tabular}{c}
    \includegraphics[height=0.28\columnwidth,keepaspectratio]{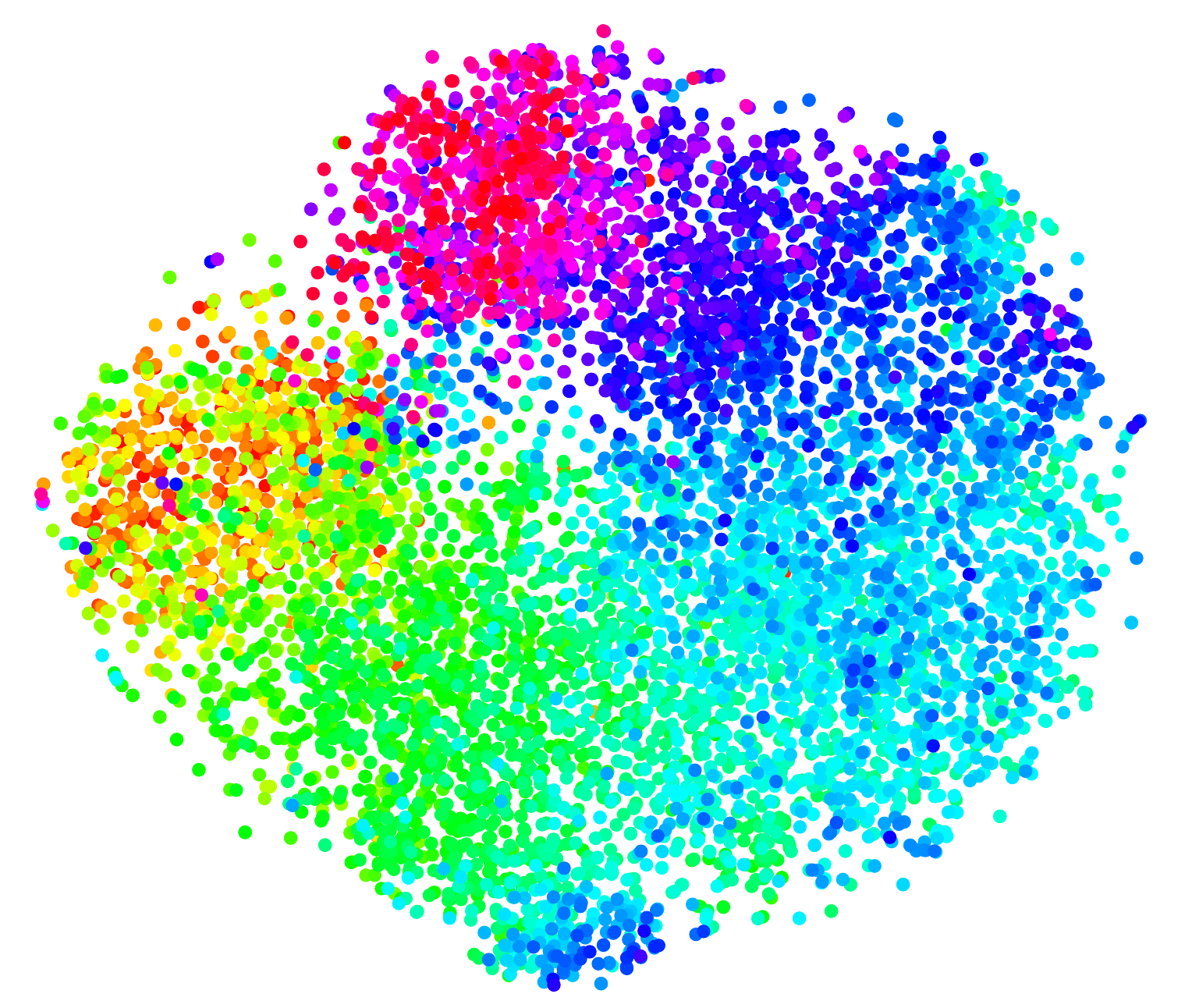} \\
    \includegraphics[height=0.28\columnwidth,keepaspectratio]{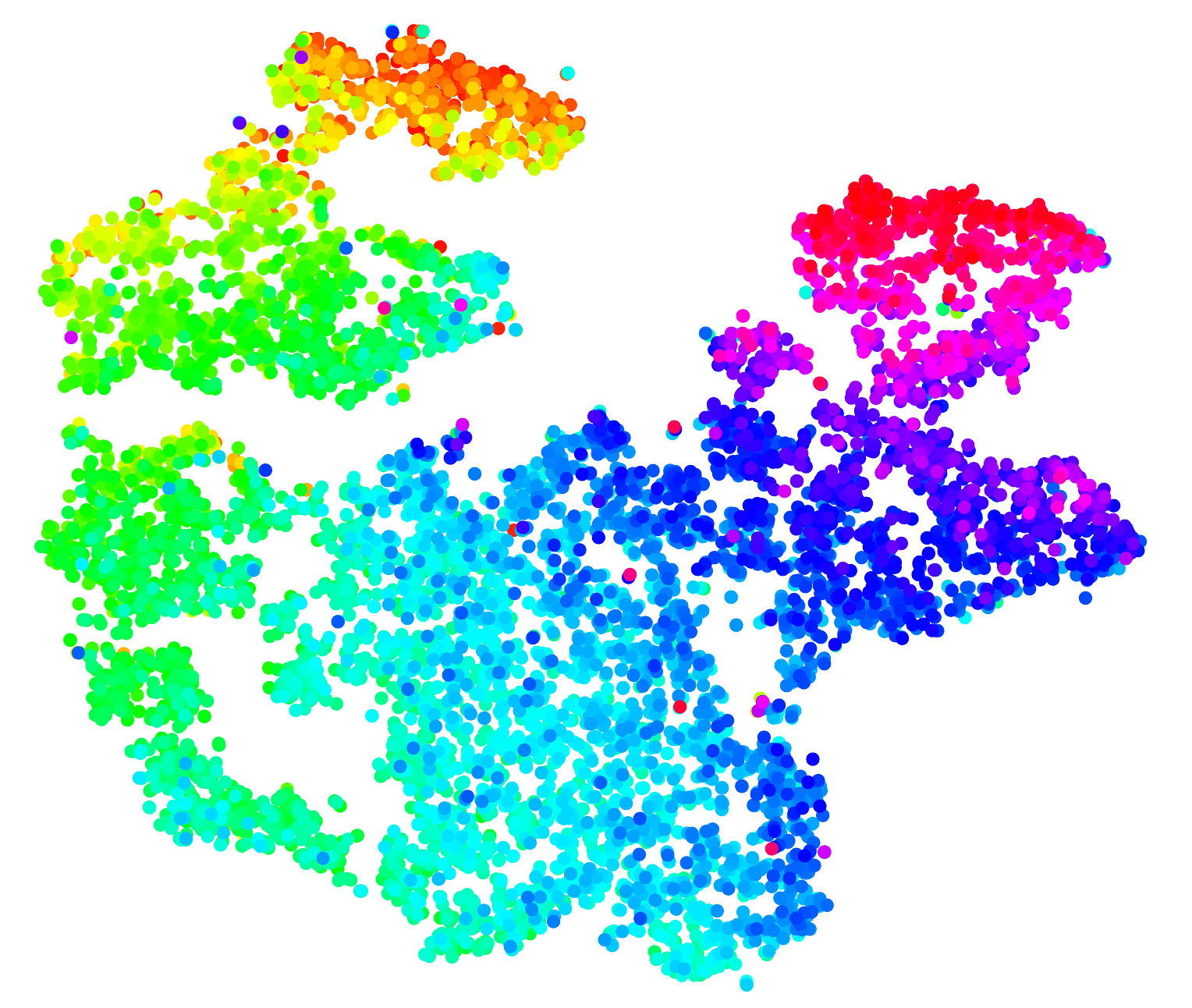}\\
   \end{tabular}
  }
\caption{Visualizations of hand-crafted and DLDL features using the t-SNE algorithm on \emph{Morph}, \emph{ChaLearn} and \emph{AFLW} validation sets. The first row shows the embeddings of hand-crafted features~(BIF or HOG). The second row shows the embeddings of the DLDL features derived from the penultimate fully connected layer of DLDL~(best viewed in color).}\label{fig:vi}
\label{fig:feat_vis}
\end{figure*}

\subsection{Semantic segmentation}

\textbf{Datasets.} We employ the PASCAL \emph{VOC2011} segmentation dataset and the Semantic Boundaries Dataset~(\emph{SBD}) for training the proposed DLDL. There are 2,224 images~(1,112 for training and 1,112 for testing) with pixel labels for 20 semantic categories in \emph{VOC2011}. \emph{SBD} contains 11,355 annotated images~(8,984 for training and 2,371 for testing) from Hariharan \emph{et al.}~\cite{hariharan2011semantic}. Following FCN~\cite{long2015fully}, we train DLDL using the union set (8,825 images) of \emph{SBD} and \emph{VOC2011} training images. We evaluate the proposed approach on \emph{VOC2011}~(1,112) and \emph{VOC2012}~(1,456) test images.

\textbf{Evaluation criteria.} The performance is measured in terms of mean IU~(intersection over union), which is the most widely used metric in semantic segmentation.

We keep the same settings as FCN including training images and model structure. The main change is that we employ KL divergence as the loss function based on label distribution~(Eq.~\ref{eq-ssld2}). Note that although we transform the ground-truth to label distribution in the training process, our evaluation rely only on ground-truth label.

Recently, Conditional Random Field~(CRF) has been broadly used in many state-of-the-art semantic segmentation systems. We optionally employ a fully connected CRF~\cite{KrahenbuhlK11} to refine the predicted category score maps using the default parameters of~\cite{CP2015Semantic}.

\textbf{Results.} Table~\ref{table:ss_test} gives the performance of DLDL-8s and DLDL-8s-CRF on the test images of \emph{VOC2011} and \emph{VOC2012} and compares it to the well-known FCN-8s. DLDL-8s improves the mean IU of FCN-8s form 62.7\% to 64.9\% on \emph{VOC2011}. On \emph{VOC2012}, DLDL-8s leads to an improvement of 2.3 points in mean IU.  DLDL achieves better results than FCN, which suggests it is important to improve the segmentation performance using label ambiguity. In addition, the CRF further improve performance of DLDL-8s, offering a 2.6\% absolute increase in mean IU both on \emph{VOC2011} and \emph{VOC2012}.

\begin{table}
	\centering
	\caption{Comparisons of DLDL and FCN on the PASCAL \emph{VOC2011} and \emph{VOC2012} test sets.}\label{table:ss_test}
	\small
	\begin{tabular}{|l|cc|}
		\hline
		\multirow{2}{*}{Methods}  &mean IU &mean IU \\
           &VOC2011 test  &VOC2012 test \\
        \hline\hline
        FCN-8s~\cite{long2015fully}  &62.7 &62.2 \\
        DLDL-8s &64.9 &64.5 \\
        DLDL-8s+CRF &\textbf{67.6} &\textbf{67.1} \\
        \hline
	\end{tabular}
\end{table}

\begin{figure}
	\centering
	\captionsetup[subfloat]{labelformat=empty}
	\captionsetup{position=top}
    \subfloat[{Image}]{
    \begin{tabular}{@{\,}c@{\,}}
    {\includegraphics[width= 0.185\columnwidth,keepaspectratio]{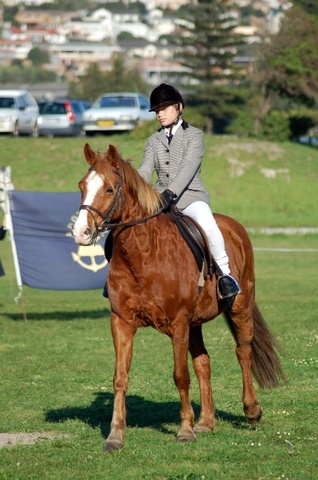}}\\
    \includegraphics[width= 0.185\columnwidth,keepaspectratio]{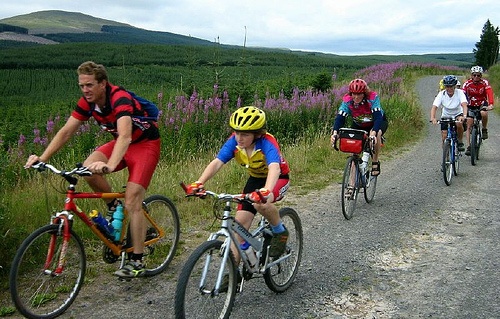}\\
    \includegraphics[width= 0.185\columnwidth,keepaspectratio]{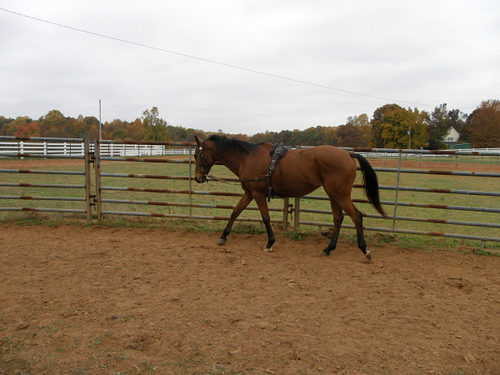}\\
    \includegraphics[width= 0.185\columnwidth,keepaspectratio]{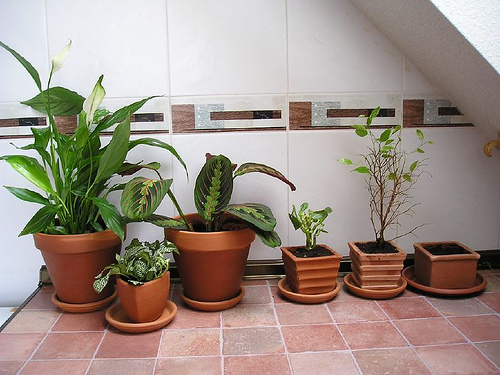}
    \end{tabular}
    }
    \hspace{-12pt}
    \subfloat[{FCN-8s}~\cite{long2015fully}]{
    \begin{tabular}{@{\,}c@{\,}}
	{\includegraphics[width= 0.185\columnwidth,keepaspectratio]{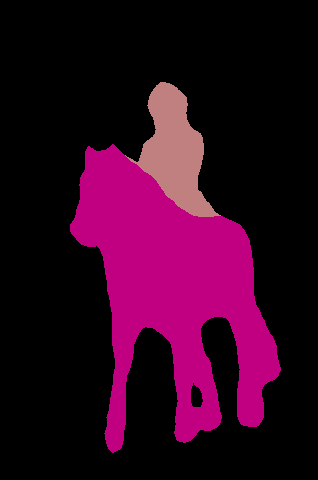}}\\
     \includegraphics[width= 0.185\columnwidth,keepaspectratio]{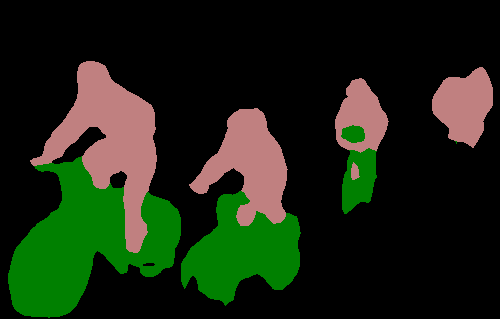}\\
     \includegraphics[width= 0.185\columnwidth,keepaspectratio]{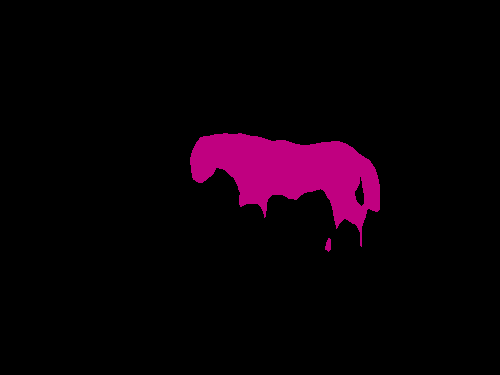}\\
     \includegraphics[width= 0.185\columnwidth,keepaspectratio]{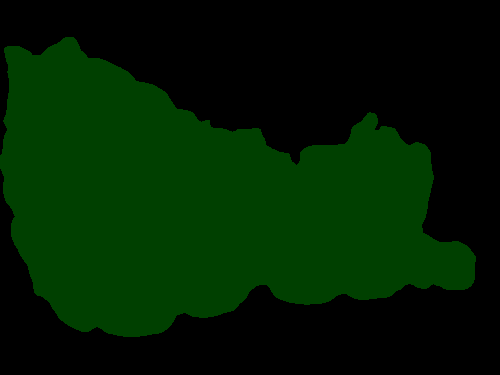}
    \end{tabular}
    }
    \hspace{-12pt}
    \subfloat[{DLDL-8s}]{
    \begin{tabular}{@{\,}c@{\,}}
	\includegraphics[width= 0.185\columnwidth,keepaspectratio]{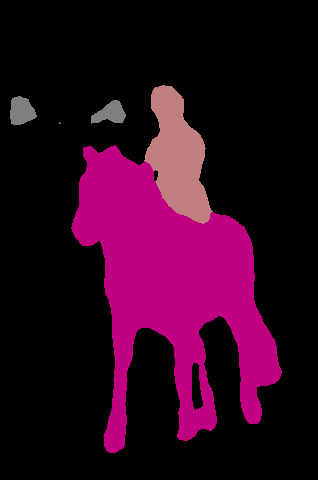}\\
    \includegraphics[width= 0.185\columnwidth,keepaspectratio]{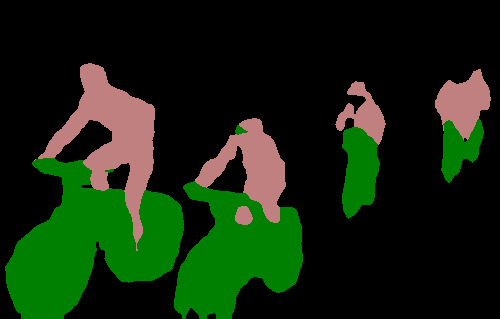}\\
    \includegraphics[width= 0.185\columnwidth,keepaspectratio]{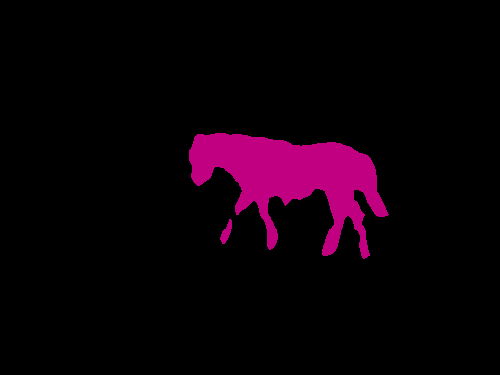}\\
    \includegraphics[width= 0.185\columnwidth,keepaspectratio]{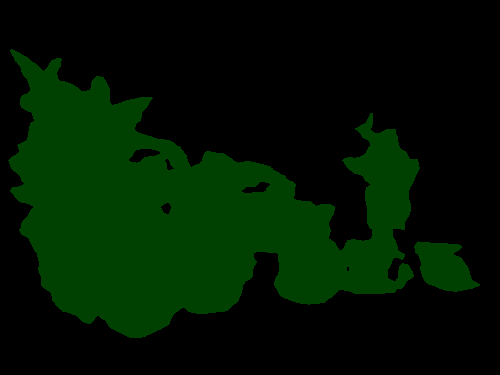}
    \end{tabular}
    }
    \hspace{-12pt}
    \subfloat[DLDL-8s+CRF]{
    \begin{tabular}{@{\,}c@{\,}}
	\includegraphics[width= 0.185\columnwidth,keepaspectratio]{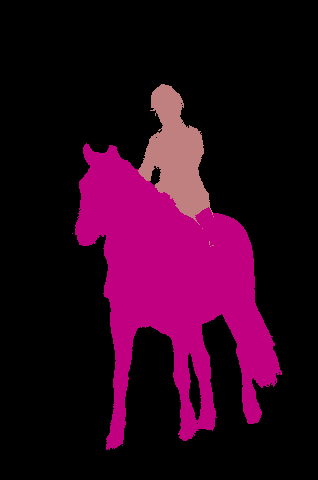}\\
    \includegraphics[width= 0.185\columnwidth,keepaspectratio]{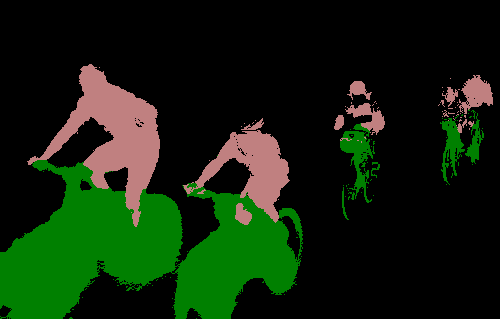}\\
    \includegraphics[width= 0.185\columnwidth,keepaspectratio]{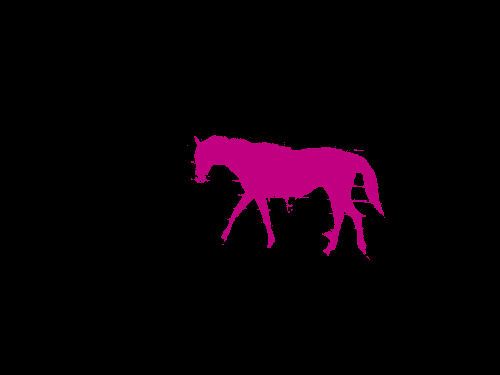}\\
    \includegraphics[width= 0.185\columnwidth,keepaspectratio]{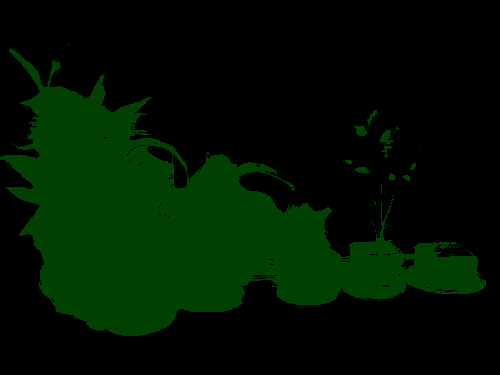}
    \end{tabular}
    }
    \hspace{-12pt}
    \subfloat[{Ground-truth}]{
    \begin{tabular}{@{\,}c@{\,}}
    {\includegraphics[width= 0.185\columnwidth,keepaspectratio]{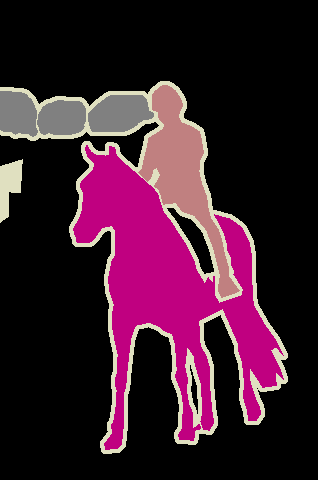}}\\
	\includegraphics[width= 0.185\columnwidth,keepaspectratio]{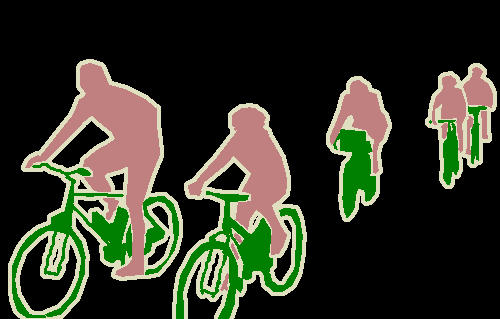}\\
    \includegraphics[width= 0.185\columnwidth,keepaspectratio]{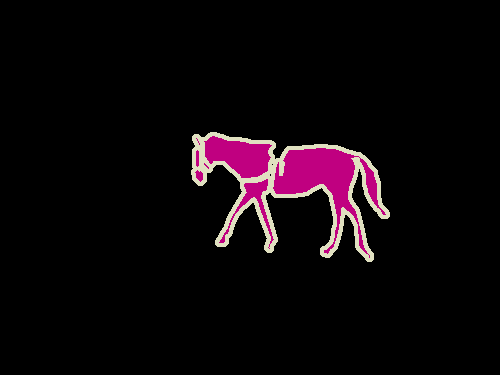}\\
    \includegraphics[width= 0.185\columnwidth,keepaspectratio]{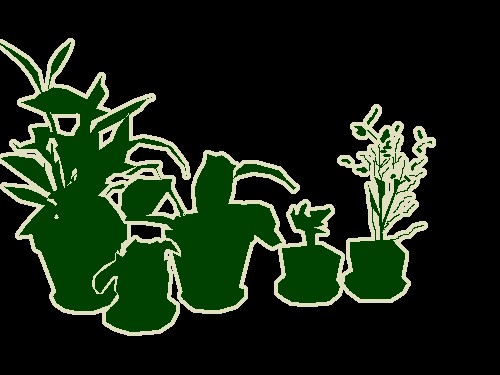}
    \end{tabular}
    }
	\caption{Semantic segmentation examples using FCN-8s, DLDL-8s and DLDL-8s-CRF on PASCAL \emph{VOC2011} validation set.}
	\label{fig:eg-ss}
\end{figure}

Fig.~\ref{fig:eg-ss} shows four semantic segmentation examples from the \emph{VOC2011} validation images using FCN-8s, DLDL-8s and DLDL-8s-CRF. We can see that DLDL-8s can successfully segment some small objects~(\emph{e.g.}, car and bicycle) and particularly improve the segmentation of object boundaries~(\emph{e.g.}, horse's leg and plant's leaves), but FCN-8s does not. DLDL-8s may fail, \emph{e.g.}, it sees a flowerpot as a potted plant in the fourth row in Fig.~\ref{fig:eg-ss}. Furthermore, compared to DLDL-8s, DLDL-8s-CRF is able to refine coarse pixel-level label predictions to produce sharp boundaries and fine-grained segmentations~(\emph{e.g.}, plant's leaves).

\begin{figure*}
    \centering
	\subfloat[ChaLearn]
	{\includegraphics[width= 0.25\textwidth]{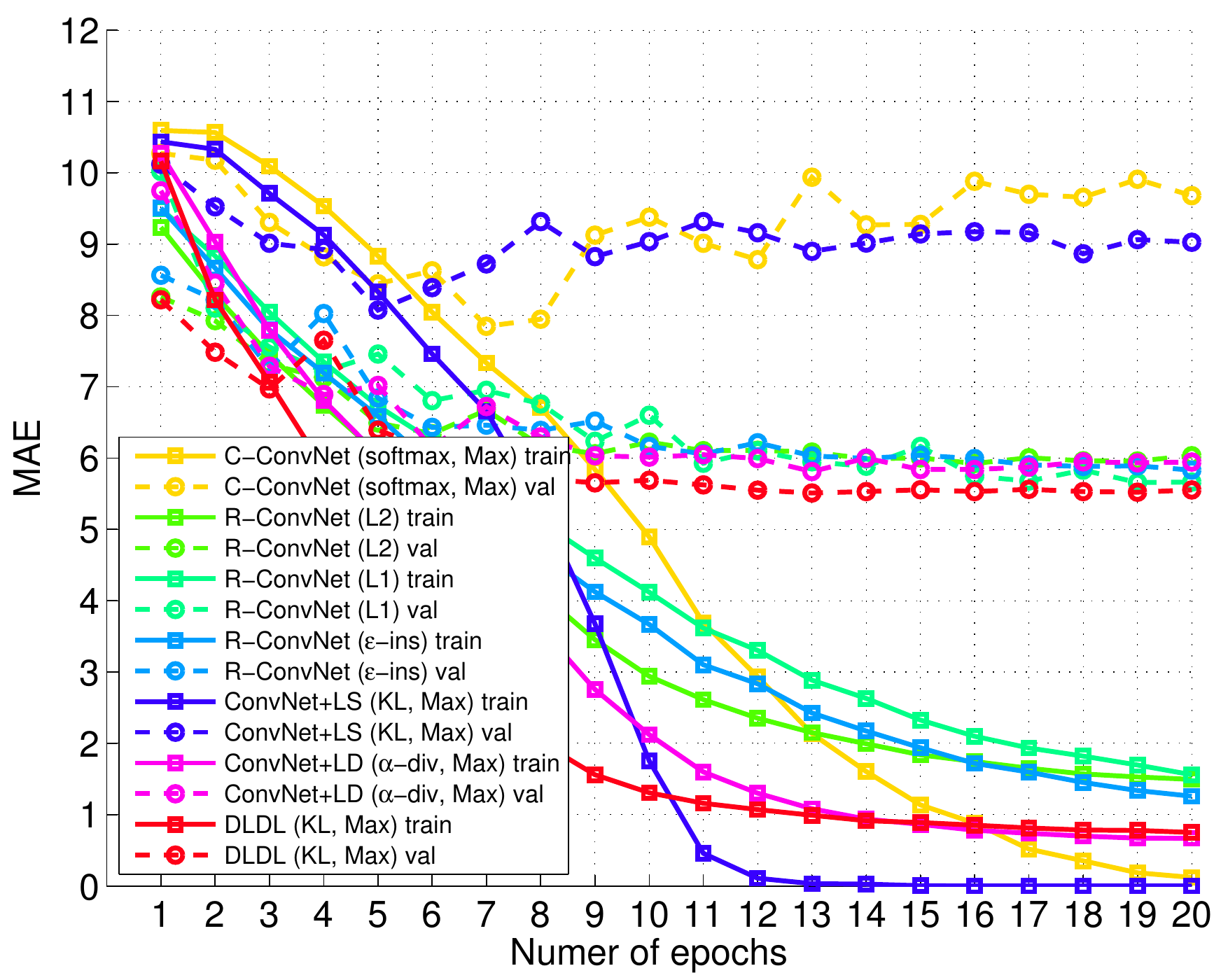}\label{fig:chaleaarn-epoch}}
    \subfloat[Morph]
    {\includegraphics[width= 0.25\textwidth]{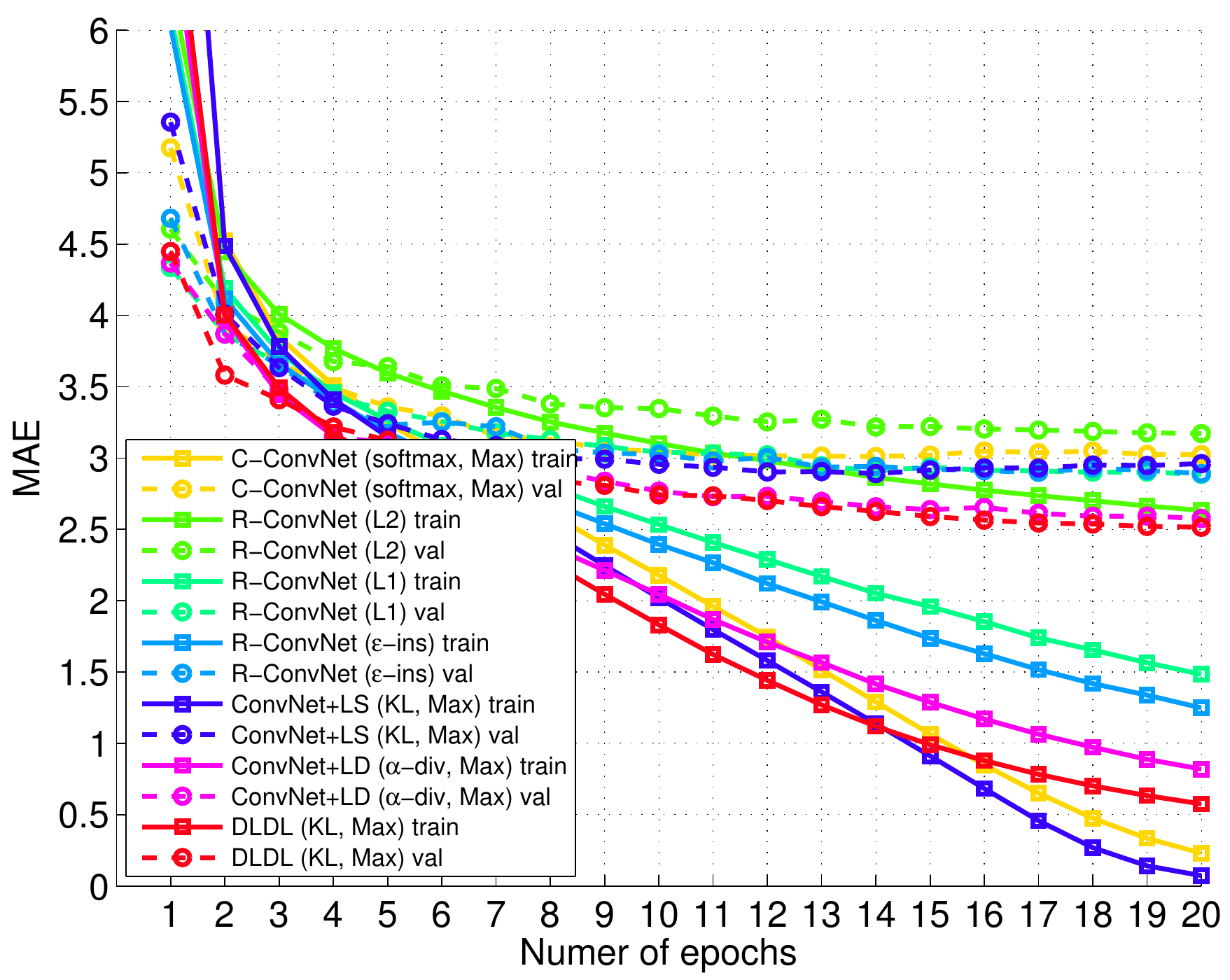}\label{fig:morph-epoch}}
	\subfloat[BJUT-3D]
	{\includegraphics[width= 0.25\textwidth]{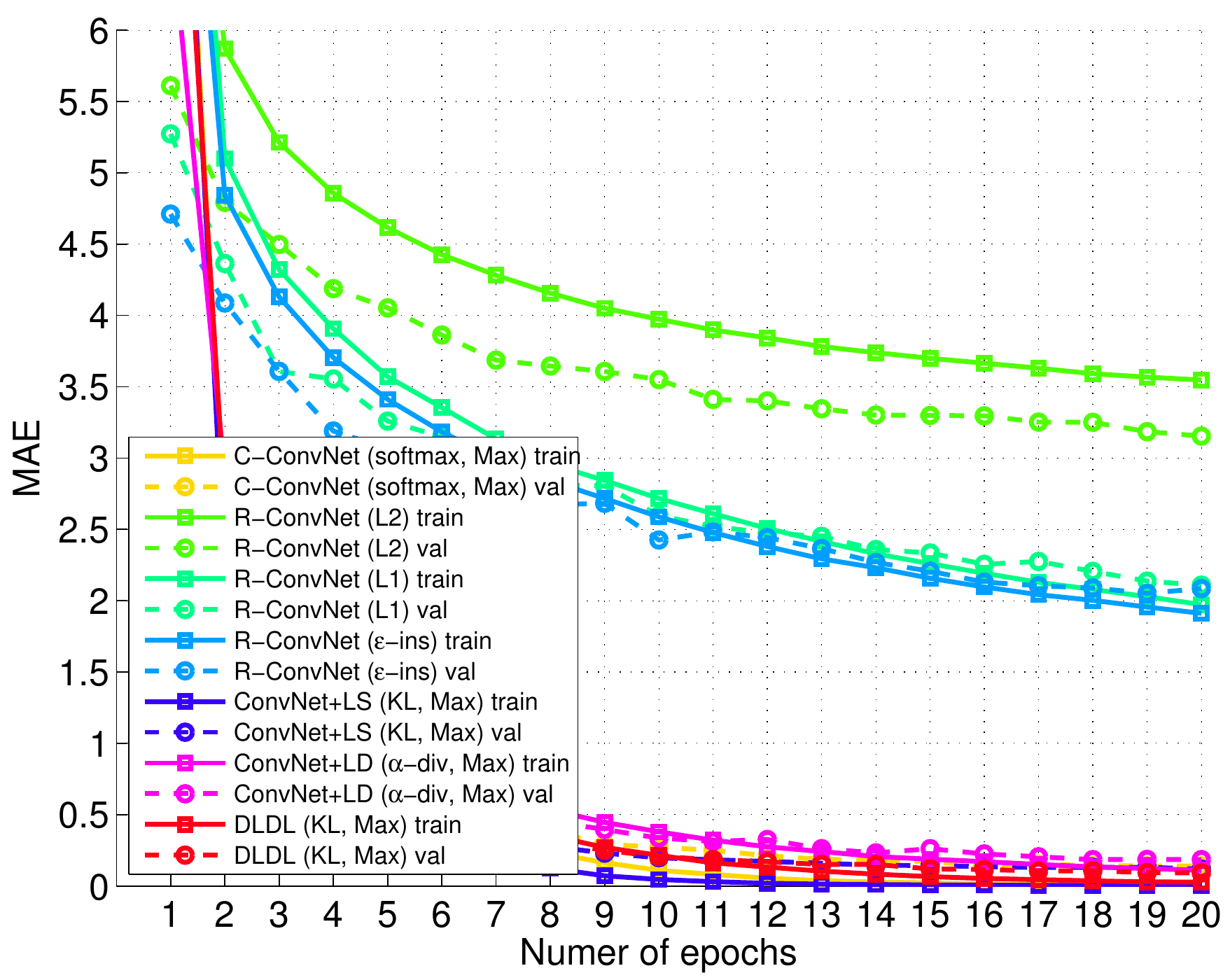} \label{fig:bjut-epoch}}
	\subfloat[AFLW]
	{\includegraphics[width= 0.25\textwidth]{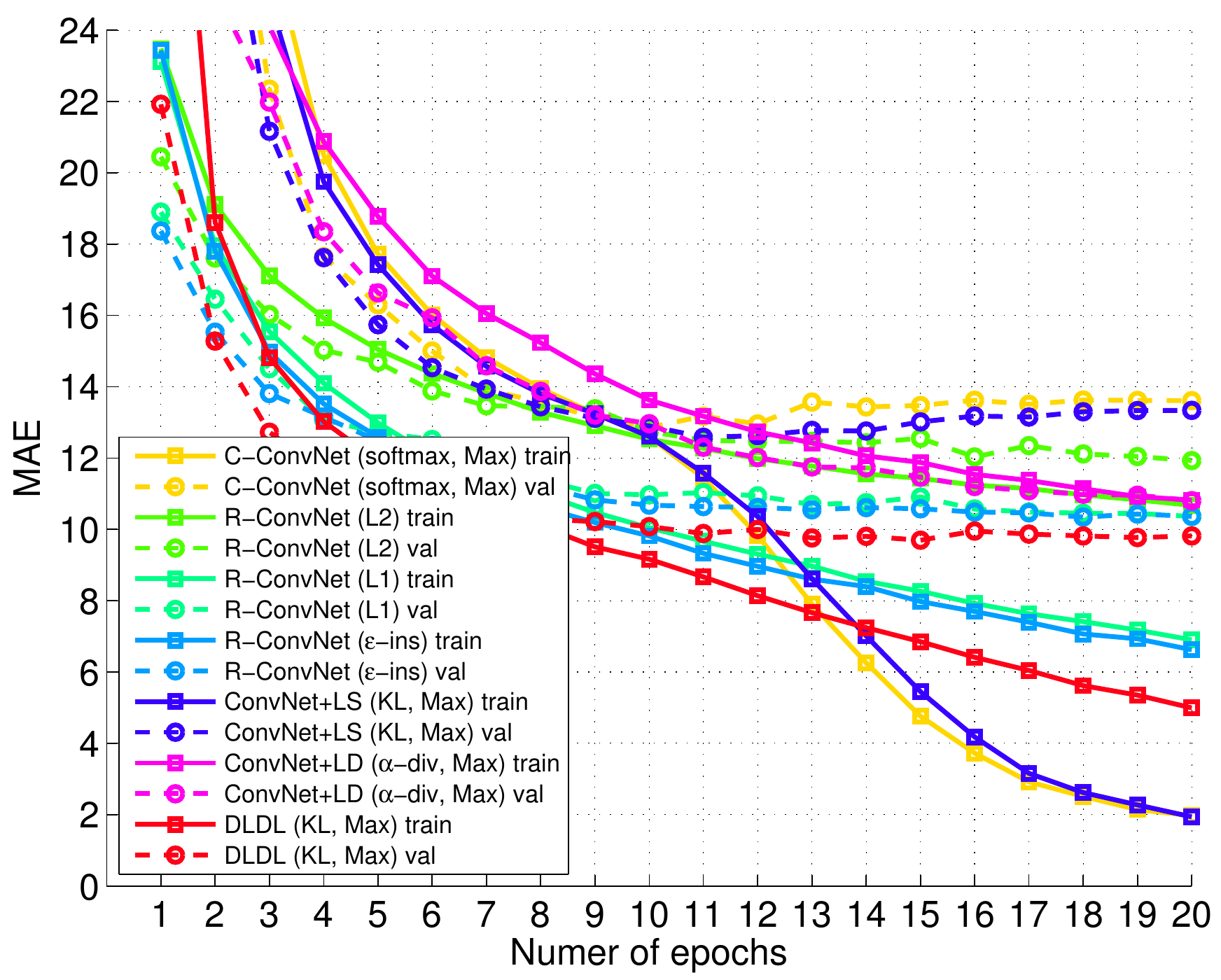} \label{fig:aflw-epoch}}
	\caption{ Comparisons of training and validation MAE of DLDL and all baseline methods on the \emph{ChaLearn}, \emph{Morph}, \emph{BJUT-3D} and \emph{AFLW} datasets (lower is better, best viewed in color).} \label{fig:mae-epoch}
\end{figure*}

\section{Discussions}\label{sec:dis}
In this section, we try to understand the generalization performance of DLDL through feature visualization, and to analyze why DLDL can achieve high accuracy with limited training data. In addition, a study of the hyper-parameter is also provided.

\textbf{Feature visualization.}
We visualize the model features in a low-dimensional space. Early layers learn low-level features~(\emph{e.g.}, edge and corner) and latter layers learn high level features~(\emph{e.g.}, shapes and objects) in a deep ConvNet~\cite{zeiler2014visualizing}. Hence, we extract the penultimate layer features~(4,096-dimensional) on \emph{Morph}, \emph{ChaLearn}, \emph{Pointing'04} and \emph{AFLW} validation sets. To obtain the 2-dimensional embeddings of the extracted high dimensional features, we employ a popular dimension reduction algorithm t-SNE~\cite{van2008vis}. The low-dimensional embeddings of validation images from the above four datasets are shown in Fig.~\ref{fig:vi}. The first row shows the 2-dim embeddings of hand-crafted features (BIF for \emph{Morph} and \emph{Chalearn}, HOG for \emph{Pointing'04} and \emph{AFLW}) and the second row shows that of the DLDL features. These figures are colored by their semantic category. It can be observed that clear semantic clusterings (old or young for age datasets, left or right, up or down for head pose datasets) appear in deep features but do not in hand-crafted features.

\textbf{Reduce over-fitting.}
DLDL can effectively reduce over-fitting when the training set is small. This effect can be explained by the label ambiguity. Considering an input sample $X$ with one single label $l$. In traditional deep ConvNet, $y_l=1$ and $y_k=0$ for all $k\neq l$. In DLDL, the label distribution $\vec y$ contains many non zeros elements. The diversity of labels helps reduce over-fitting. Moreover, the objective function~(Eq.~\ref{eq-of}) of DLDL can be rewritten as
 \begin{equation}
	T =  -({y_l}\ln{\hat y_l} + \sum_{k\neq l} {y_k}\ln{\hat y_k}) \,. \label{eq-rof}
\end{equation}
 In Eq.~\ref{eq-rof}, the first term is the tradition ConvNet loss function. The second term maximize the log-likelihood of the ambiguous labels. Unlike existing data augmentation techniques such as random cropping on the images, DLDL augments data on the label side.

 In Fig.~\ref{fig:mae-epoch}, MAE is shown as a function of the number of epochs on two age datasets~(\emph{ChaLearn} and \emph{Morph}) and two head pose datasets (\emph{BJUT-3D} and \emph{AFLW}). On \emph{ChaLearn} and \emph{AFLW}, C-ConveNet~(softmax) achieves the lowest training MAE, but produces the highest validation MAE. In particular, the validation MAE  increases after the 8th epoch on \emph{ChaLearn}. Similar phenomenon is observed on \emph{AFLW}. This fact shows that over-fitting happens in C-ConvNet when the number of training images is small. Although there are 15,561 training images in \emph{AFLW}, each category contains on averagely 4 training images since there are 3,721 categories.

\textbf{Accelerate convergence.}
 We further analyze the convergence performance of DLDL, C-ConvNet and R-ConvNet. We can observe that the training MAE is reduced very slowly at the beginning of training using C-ConvNet and R-ConveNet in many cases as shown in Fig.~\ref{fig:mae-epoch}. On the contrary, the MAE of DLDL reduces quickly.

\textbf{Robust performance.}
 One notable observation is that C-ConvNet and R-ConveNet is unstable. Fig.~\ref{fig:bjut-epoch} shows the MAE for pitch+yaw, a complicated estimation of the joint distribution. This is a very sparse label set because the interval of adjacent class~(pitch or yaw) is $10^\circ$. R-ConvNet has difficulty in estimating this output, yielding errors that are roughly 20 times higher than DLDL and C-ConvNet. On the other hand, C-ConvNet easily fall into over-fitting when there are not enough training data~(\emph{e.g}, Fig.~\ref{fig:chaleaarn-epoch} and Fig.~\ref{fig:aflw-epoch}). The proposed DLDL is more amenable to small datasets or sparse labels than C-ConvNet and R-ConvNet.

\begin{figure}
	\centering
	\includegraphics[width= 0.48\columnwidth]{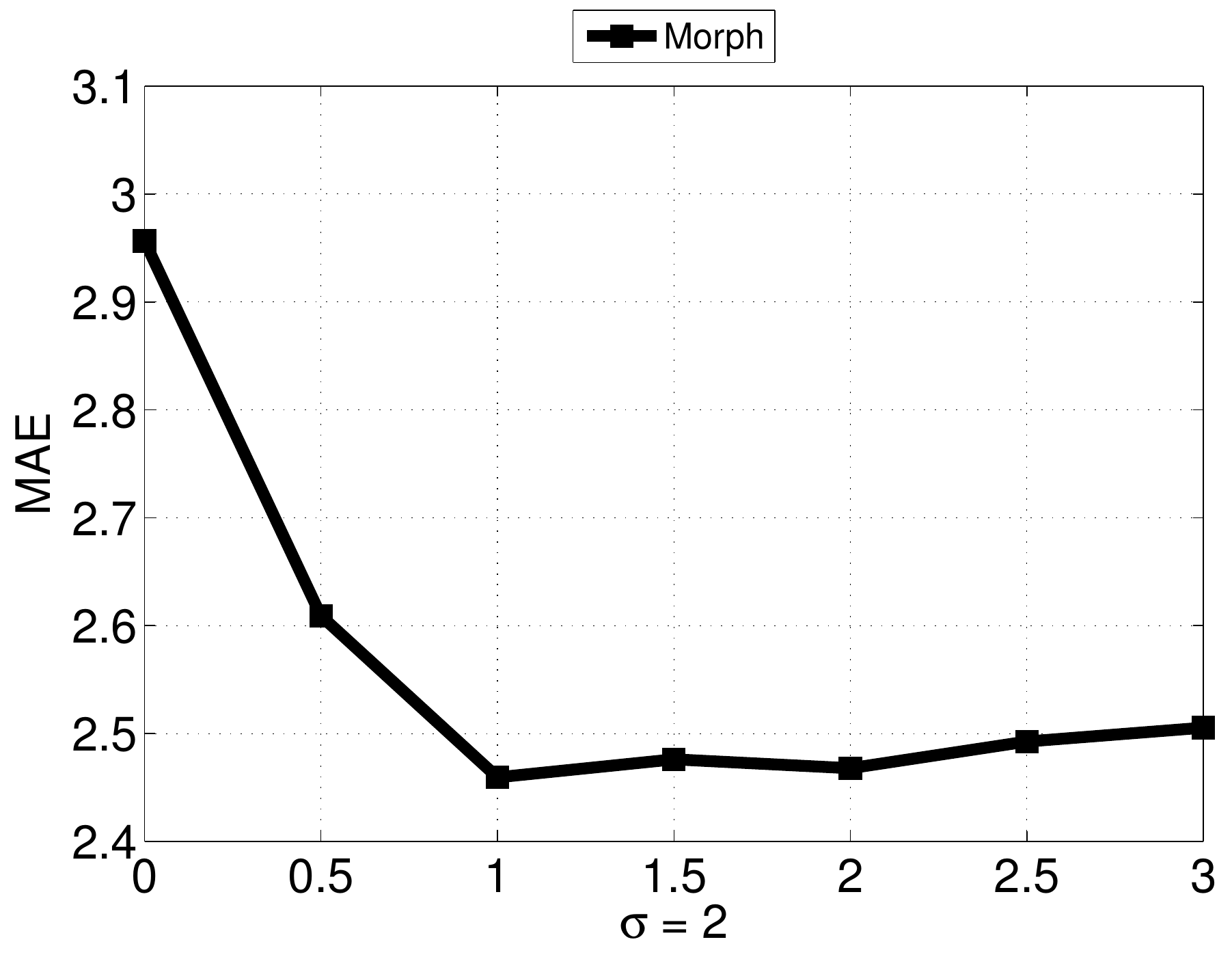}
	\includegraphics[width= 0.48\columnwidth]{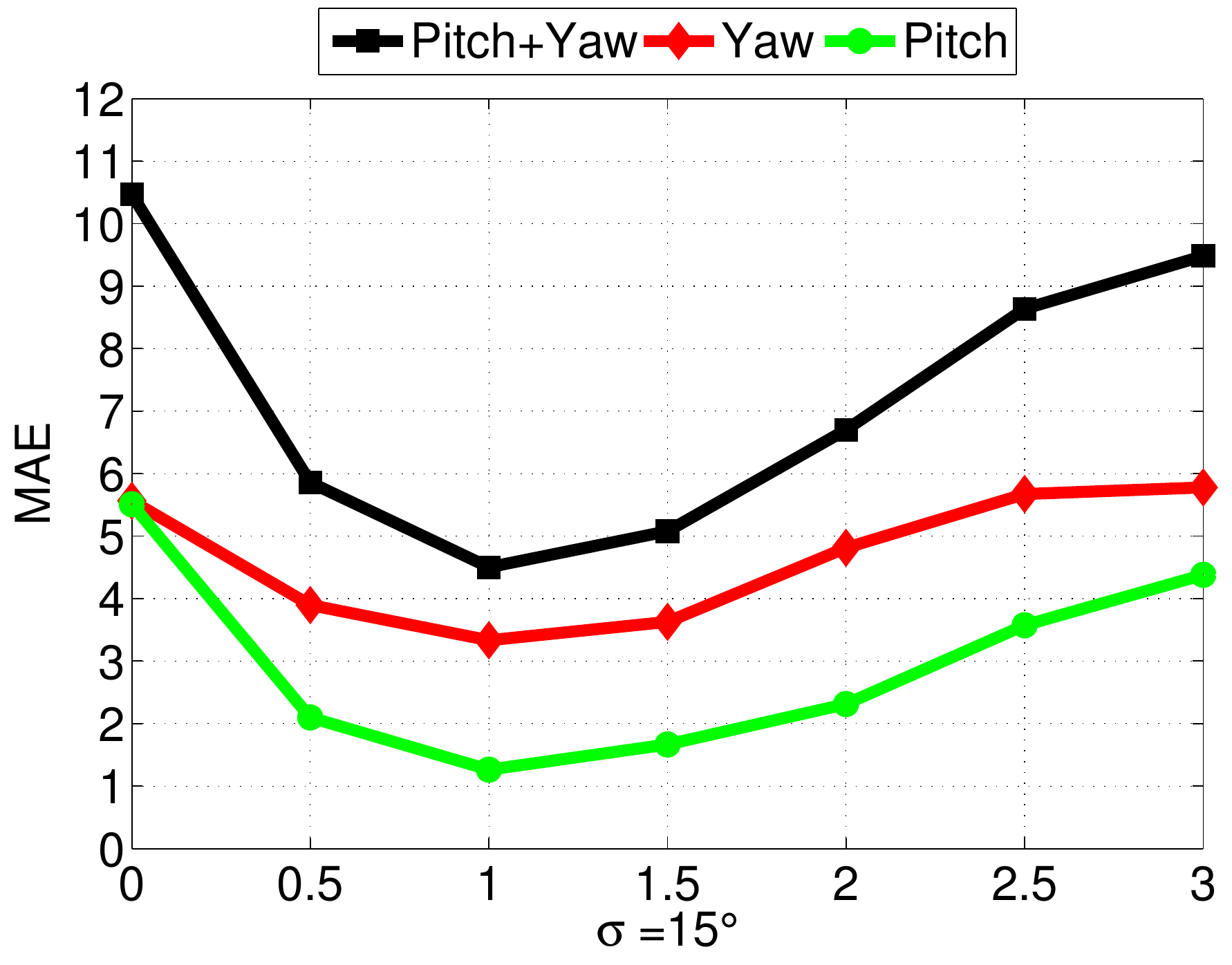}
	\caption{The performance (MAE) of DLDL with different label distributions (different parameter $\sigma$). The left figure is for the \emph{Morph} dataset, while the right figure is for the \emph{Pointing'04} dataset (lower is better).}
	\label{fig:pa}
\end{figure}

\textbf{Analyze the hyper-parameter.}
DLDL's performance may be affected by the label distribution. Here, we take age estimation~(\emph{Morph}) and head pose estimation~(\emph{Pointing'04}) for examples. $\sigma$ is a common hyper-parameter in these tasks if it is not provided in the ground-truth. We have empirically set $\sigma=2$ in \emph{Morph}, and $\sigma = 15^\circ$ in \emph{Pointing'04} in our experiments. In order to study the impact of $\sigma$, we test DLDL with different $\sigma$ values, changing from 0 to 3$\sigma$ with 0.5$\sigma$ interval. Fig.~\ref{fig:pa} shows the MAE performance on \emph{Morph} and \emph{Pointing'04} with different $\sigma$. We can see that a proper $\sigma$ is important for low MAE. But generally speaking, a $\sigma$ value that is close to the interval between neighboring labels is a good choice. Because the shape of all curves are V-shape like, it is also very convenient to find an optimal $\sigma$ value using the cross-validation strategy.

\section{Conclusion}\label{sec:co}

We observe that current deep ConvNets cannot successfully learn good models when there are not enough training data and/or the labels are ambiguous. We propose DLDL, a deep label distribution learning framework to solve this issue by exploiting label ambiguity. In DLDL, each image is labeled by a label distribution, which can utilize label ambiguity in both feature learning and classifier learning. DLDL consistently improves the network training process in our experiments, by preventing it from over-fitting when the training set is small. We empirically showed that DLDL produces robust and competitive performances than traditional classification or regression deep models on several popular visual recognition tasks.

However, constructing a reasonable label distribution is still challenging due to the diversity of label space for different recognition tasks. It is an interesting direction to extend DLDL to more recognition problems by constructing different label distributions.

\bibliographystyle{IEEEtran}
\bibliography{egbib}

\begin{thebibliography}{10}
\providecommand{\url}[1]{#1}
\csname url@samestyle\endcsname
\providecommand{\newblock}{\relax}
\providecommand{\bibinfo}[2]{#2}
\providecommand{\BIBentrySTDinterwordspacing}{\spaceskip=0pt\relax}
\providecommand{\BIBentryALTinterwordstretchfactor}{4}
\providecommand{\BIBentryALTinterwordspacing}{\spaceskip=\fontdimen2\font plus
\BIBentryALTinterwordstretchfactor\fontdimen3\font minus
  \fontdimen4\font\relax}
\providecommand{\BIBforeignlanguage}[2]{{%
\expandafter\ifx\csname l@#1\endcsname\relax
\typeout{** WARNING: IEEEtran.bst: No hyphenation pattern has been}%
\typeout{** loaded for the language `#1'. Using the pattern for}%
\typeout{** the default language instead.}%
\else
\language=\csname l@#1\endcsname
\fi
#2}}
\providecommand{\BIBdecl}{\relax}
\BIBdecl

\bibitem{krizhevsky2012imagenet}
A.~Krizhevsky, I.~Sutskever, and G.~E. Hinton, ``{ImageNet} classification with
  deep convolutional neural networks,'' in \emph{Advances in Neural Information
  Processing Systems}, 2012, pp. 1097--1105.

\bibitem{girshick2014rich}
R.~Girshick, J.~Donahue, T.~Darrell, and J.~Malik, ``Rich feature hierarchies
  for accurate object detection and semantic segmentation,'' in
  \emph{Proceedings of the IEEE Conference on Computer Vision and Pattern
  Recognition}, 2014, pp. 580--587.

\bibitem{long2015fully}
J.~Long, E.~Shelhamer, and T.~Darrell, ``Fully convolutional networks for
  semantic segmentation,'' in \emph{Proceedings of the IEEE Conference on
  Computer Vision and Pattern Recognition}, 2015, pp. 3431--3440.

\bibitem{geng2013facial}
X.~Geng, C.~Yin, and Z.-H. Zhou, ``Facial age estimation by learning from label
  distributions,'' \emph{IEEE Transactions on Pattern Analysis and Machine
  Intelligence}, vol.~35, no.~10, pp. 2401--2412, 2013.

\bibitem{kong2015head}
S.~G. Kong and R.~O. Mbouna, ``Head pose estimation from a {2D} face image
  using {3D} face morphing with depth parameters,'' \emph{IEEE Transactions on
  Image Processing}, vol.~24, no.~6, pp. 1801--1808, 2015.

\bibitem{everingham2010pascal}
M.~Everingham, L.~Van~Gool, C.~K. Williams, J.~Winn, and A.~Zisserman, ``The
  {PASCAL} visual object classes {(VOC)} challenge,'' \emph{International
  Journal of Computer Vision}, vol.~88, no.~2, pp. 303--338, 2010.

\bibitem{geng2014head}
X.~Geng and Y.~Xia, ``Head pose estimation based on multivariate label
  distribution,'' in \emph{Proceedings of the IEEE Conference on Computer
  Vision and Pattern Recognition}, 2014, pp. 1837--1842.

\bibitem{xing2016logistic}
C.~Xing, X.~Geng, and H.~Xue, ``Logistic boosting regression for label
  distribution learning,'' in \emph{Proceedings of the IEEE Conference on
  Computer Vision and Pattern Recognition}, 2016, pp. 4489--4497.

\bibitem{he35data}
Z.~He, X.~Li, Z.~Zhang, F.~Wu, X.~Geng, Y.~Zhang, M.-H. Yang, and Y.~Zhuang,
  ``Data-dependent label distribution learning for age estimation,'' \emph{IEEE
  Transactions on Image Processing}, 2017, to be published, doi:
  10.1109/TIP.2017.2655445.

\bibitem{he2015deep}
K.~He, X.~Zhang, S.~Ren, and J.~Sun, ``Deep residual learning for image
  recognition,'' in \emph{Proceedings of the IEEE Conference on Computer Vision
  and Pattern Recognition}, 2016, pp. 770--778.

\bibitem{belagiannis2015robust}
V.~Belagiannis, C.~Rupprecht, G.~Carneiro, and N.~Navab, ``Robust optimization
  for deep regression,'' in \emph{Proceedings of the IEEE International
  Conference on Computer Vision}, 2015, pp. 2830--2838.

\bibitem{fanelli2011real}
G.~Fanelli, J.~Gall, and L.~Van~Gool, ``Real time head pose estimation with
  random regression forests,'' in \emph{Proceedings of the IEEE Conference on
  Computer Vision and Pattern Recognition}, 2011, pp. 617--624.

\bibitem{lu2015cost}
J.~Lu, V.~E. Liong, and J.~Zhou, ``Cost-sensitive local binary feature learning
  for facial age estimation,'' \emph{IEEE Transactions on Image Processing},
  vol.~24, no.~12, pp. 5356--5368, 2015.

\bibitem{ahn2015real}
B.~Ahn, J.~Park, and I.~S. Kweon, ``Real-time head orientation from a monocular
  camera using deep neural network,'' in \emph{Asian Conference on Computer
  Vision}, 2015, pp. 82--96.

\bibitem{sun2013deep}
Y.~Sun, X.~Wang, and X.~Tang, ``Deep convolutional network cascade for facial
  point detection,'' in \emph{Proceedings of the IEEE Conference on Computer
  Vision and Pattern Recognition}, 2013, pp. 3476--3483.

\bibitem{simonyan2015very}
K.~Simonyan and A.~Zisserman, ``Very deep convolutional networks for
  large-scale image recognition,'' in \emph{Proceedings of International
  Conference on Learning Representations}, 2015, pp. 1--14.

\bibitem{Parkhi15}
O.~M. Parkhi, A.~Vedaldi, and A.~Zisserman, ``Deep face recognition,'' in
  \emph{Proceedings of the British Machine Vision Conference}, 2015, p.~6.

\bibitem{szegedy2015rethinking}
C.~Szegedy, V.~Vanhoucke, S.~Ioffe, J.~Shlens, and Z.~Wojna, ``Rethinking the
  inception architecture for computer vision,'' in \emph{Proceedings of the
  IEEE Conference on Computer Vision and Pattern Recognition}, 2016, pp.
  2818--2826.

\bibitem{zeiler2014visualizing}
M.~D. Zeiler and R.~Fergus, ``Visualizing and understanding convolutional
  networks,'' in \emph{European Conference on Computer Vision}, 2014, pp.
  818--833.

\bibitem{he2015delving}
K.~He, X.~Zhang, S.~Ren, and J.~Sun, ``Delving deep into rectifiers: Surpassing
  human-level performance on imagenet classification,'' in \emph{Proceedings of
  the IEEE International Conference on Computer Vision}, 2015, pp. 1026--1034.

\bibitem{vedaldi15matconvnet}
A.~Vedaldi and K.~Lenc, ``{MatConvNet}: Convolutional neural networks for
  {MATLAB},'' in \emph{Proceedings of the 23rd {ACM} International Conference
  on Multimedia}, 2015, pp. 689--692.

\bibitem{ricanek2006morph}
K.~Ricanek~Jr and T.~Tesafaye, ``Morph: {A} longitudinal image database of
  normal adult age-progression,'' in \emph{International Conference on
  Automatic Face and Gesture Recognition}, 2006, pp. 341--345.

\bibitem{escalera2015chalearn}
S.~Escalera, J.~Fabian, P.~Pardo, X.~Bar{\'o}, J.~Gonzalez, H.~J. Escalante,
  D.~Misevic, U.~Steiner, and I.~Guyon, ``Chalearn looking at people 2015:
  Apparent age and cultural event recognition datasets and results,'' in
  \emph{Proceedings of the IEEE International Conference on Computer Vision
  Workshops}, 2015, pp. 1--9.

\bibitem{minka2005divergence}
T.~Minka, ``Divergence measures and message passing,'' Microsoft Research,
  Tech. Rep. MSR-TR-2005-173, 2005.

\bibitem{geng2016label}
X.~Geng, ``Label distribution learning,'' \emph{IEEE Transactions on Knowledge
  and Data Engineering}, vol.~28, no.~7, pp. 1734--1748, 2016.

\bibitem{mathias2014face}
M.~Mathias, R.~Benenson, M.~Pedersoli, and L.~Van~Gool, ``Face detection
  without bells and whistles,'' in \emph{European Conference on Computer
  Vision}, 2014, pp. 720--735.

\bibitem{chang2015learning}
K.-Y. Chang and C.-S. Chen, ``A learning framework for age rank estimation
  based on face images with scattering transform,'' \emph{IEEE Transactions on
  Image Processing}, vol.~24, no.~3, pp. 785--798, 2015.

\bibitem{yi2015age}
D.~Yi, Z.~Lei, and S.~Z. Li, ``Age estimation by multi-scale convolutional
  network,'' in \emph{Asian Conference on Computer Vision}, 2015, pp. 144--158.

\bibitem{huerta2015deep}
I.~Huerta, C.~Fern{\'a}ndez, C.~Segura, J.~Hernando, and A.~Prati, ``A deep
  analysis on age estimation,'' \emph{Pattern Recognition Letters}, vol.~68,
  pp. 239--249, 2015.

\bibitem{rothe2015dex}
R.~Rothe, R.~Timofte, and L.~Gool, ``{DEX: Deep EXpectation} of apparent age
  from a single image,'' in \emph{Proceedings of the IEEE International
  Conference on Computer Vision Workshops}, 2015, pp. 252--257.

\bibitem{rothe2016deep}
R.~Rothe, R.~Timofte, and L.~Van~Gool, ``Deep expectation of real and apparent
  age from a single image without facial landmarks,'' \emph{International
  Journal of Computer Vision}, pp. 1--14, 2016, doi:10.1007/s11263-016-0940-36.

\bibitem{gourier2004estimating}
N.~Gourier, D.~Hall, and J.~L. Crowley, ``Estimating face orientation from
  robust detection of salient facial structures,'' in \emph{FG Net Workshop on
  Visual Observation of Deictic Gestures}, 2004, pp. 1--9.

\bibitem{baocai2009bjut}
B.~Yin, Y.~Sun, C.~Wang, and Y.~Ge, ``{BJUT-3D} large scale {3D} face database
  and information processing,'' \emph{Journal of Computer Research and
  Development}, vol.~46, no.~6, pp. 1009--1018, 2009.

\bibitem{tugraz:icg:lrs}
M.~Koestinger, P.~Wohlhart, P.~M. Roth, and H.~Bischof, ``Annotated facial
  landmarks in the wild: A large-scale, real-world database for facial landmark
  localization,'' in \emph{Proceedings of the IEEE International Conference on
  Computer Vision Workshops}, 2011, pp. 2144--2151.

\bibitem{dementhon1995model}
D.~F. Dementhon and L.~S. Davis, ``Model-based object pose in 25 lines of
  code,'' \emph{International Journal of Computer Vision}, vol.~15, no. 1--2,
  pp. 123--141, 1995.

\bibitem{sundararajan2015head}
K.~Sundararajan and D.~Woodard, ``Head pose estimation in the wild using
  approximate view manifolds,'' in \emph{Proceedings of the IEEE Conference on
  Computer Vision and Pattern Recognition Workshops}, 2015, pp. 50--58.

\bibitem{wei2014cnn}
Y.~Wei, W.~Xia, J.~Huang, B.~Ni, J.~Dong, Y.~Zhao, and S.~Yan, ``{CNN:}
  single-label to multi-label,'' \emph{CoRR, abs:1406.5726}, 2014.

\bibitem{cheng2014bing}
M.-M. Cheng, Z.~Zhang, W.-Y. Lin, and P.~Torr, ``{BING:} binarized normed
  gradients for objectness estimation at 300fps,'' in \emph{Proceedings of the
  IEEE Conference on Computer Vision and Pattern Recognition}, 2014, pp.
  3286--3293.

\bibitem{zitnick2014edge}
C.~L. Zitnick and P.~Doll{\'a}r, ``Edge boxes: Locating object proposals from
  edges,'' in \emph{European Conference on Computer Vision}, 2014, pp.
  391--405.

\bibitem{yang2016exp}
H.~Yang, J.~T. Zhou, Y.~Zhang, B.-B. Gao, J.~Wu, and J.~Cai, ``Exploit bounding
  box annotations for multi-label object recognition,'' in \emph{Proceedings of
  the IEEE Conference on Computer Vision and Pattern Recognition}, 2016, pp.
  280--288.

\bibitem{gong2013deep}
Y.~Gong, Y.~Jia, T.~Leung, A.~Toshev, and S.~Ioffe, ``Deep convolutional
  ranking for multilabel image annotation,'' \emph{CoRR, abs:1312.4894}, 2013.

\bibitem{wei2015hcp}
Y.~Wei, W.~Xia, M.~Lin, J.~Huang, B.~Ni, J.~Dong, Y.~Zhao, and S.~Yan, ``{HCP}:
  A flexible {CNN} framework for multi-label image classification,'' \emph{IEEE
  Transactions on Pattern Analysis and Machine Intelligence}, vol.~38, no.~9,
  pp. 1901--1907, 2015.

\bibitem{dong2013sub}
J.~Dong, W.~Xia, Q.~Chen, J.~Feng, Z.~Huang, and S.~Yan, ``Subcategory-aware
  object classification,'' in \emph{Proceedings of the IEEE Conference on
  Computer Vision and Pattern Recognition}, 2013, pp. 827--834.

\bibitem{song2011contextualizing}
Z.~Song, Q.~Chen, Z.~Huang, Y.~Hua, and S.~Yan, ``Contextualizing object
  detection and classification,'' in \emph{Proceedings of the IEEE Conference
  on Computer Vision and Pattern Recognition}, 2011, pp. 1585--1592.

\bibitem{oquab2014learning}
M.~Oquab, L.~Bottou, I.~Laptev, and J.~Sivic, ``Learning and transferring
  mid-level image representations using convolutional neural networks,'' in
  \emph{Proceedings of the IEEE Conference on Computer Vision and Pattern
  Recognition}, 2014, pp. 1717--1724.

\bibitem{shi2000normalized}
J.~Shi and J.~Malik, ``Normalized cuts and image segmentation,'' \emph{IEEE
  Transactions on Pattern Analysis and Machine Intelligence}, vol.~22, no.~8,
  pp. 888--905, 2000.

\bibitem{lai2016instance}
H.~Lai, P.~Yan, X.~Shu, Y.~Wei, and S.~Yan, ``{Instance-Aware} hashing for
  multi-label image retrieval,'' \emph{IEEE Transactions on Image Processing},
  vol.~25, no.~6, pp. 2469--2479, 2016.

\bibitem{zhao2016regional}
R.-W. Zhao, J.~Li, Y.~Chen, J.-M. Liu, Y.-G. Jiang, and X.~Xue, ``Regional
  gating neural networks for multi-label image classification,'' in
  \emph{Proceedings of the British Machine Vision Conference}, vol.~6, 2016.

\bibitem{hariharan2011semantic}
B.~Hariharan, P.~Arbel{\'a}ez, L.~Bourdev, S.~Maji, and J.~Malik, ``Semantic
  contours from inverse detectors,'' in \emph{Proceedings of the IEEE
  International Conference on Computer Vision}, 2011, pp. 991--998.

\bibitem{KrahenbuhlK11}
P.~Kr{\"{a}}henb{\"{u}}hl and V.~Koltun, ``Efficient inference in fully
  connected {CRFs} with gaussian edge potentials,'' in \emph{Advances in Neural
  Information Processing Systems}, 2011, pp. 109--117.

\bibitem{CP2015Semantic}
L.-C. Chen, G.~Papandreou, I.~Kokkinos, K.~Murphy, and A.~L. Yuille, ``Semantic
  image segmentation with deep convolutional nets and fully connected {CRFs},''
  in \emph{Proceedings of International Conference on Learning
  Representations}, 2015.

\bibitem{van2008vis}
L.~Van~der Maaten and G.~Hinton, ``Visualizing data using t-{SNE},''
  \emph{Journal of Machine Learning Research}, vol.~9, no. Nov, pp. 2579--2605,
  2008.

\end{thebibliography}
\newpage
\begin{IEEEbiography}[{\includegraphics[width=1in,height=1.25in,clip,keepaspectratio]{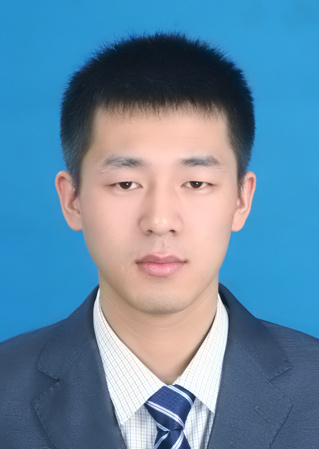}}]{Bin-Bin Gao}
received the B.S. and M.S. degrees in applied mathematics in 2010 and 2013, respectively. He is currently pursuing 
the Ph.D. degree in the Department of Computer Science and Technology, Nanjing University, China. 
His research interests include computer vision and machine learning.
\end{IEEEbiography}
\begin{IEEEbiography}[{\includegraphics[width=1in,height=1.25in,clip,keepaspectratio]{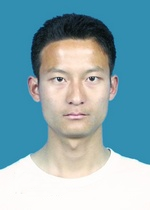}}]{Chao Xing}
received the B.S. degree in software engineering from Southeast University, China, in 2014. 
He is currently a postgraduate student in the School of Computer Science and Engineering at Southeast University, 
China. His research interests include pattern recognition, machine learning, and data mining.
\end{IEEEbiography}
\begin{IEEEbiography}[{\includegraphics[width=1in,height=1.25in,clip,keepaspectratio]{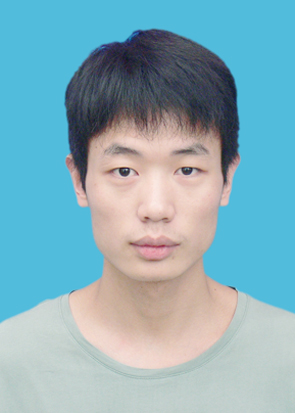}}]{Chen-Wei Xie}
received his B.S. degree from Southeast University, China, in 2015. He is currently a postgraduate student in 
the Department of Computer Science and Technology, Nanjing University, China. His research interests include 
computer vision and machine learning.
\end{IEEEbiography}
\begin{IEEEbiography}[{\includegraphics[width=1in,height=1.25in,clip,keepaspectratio]{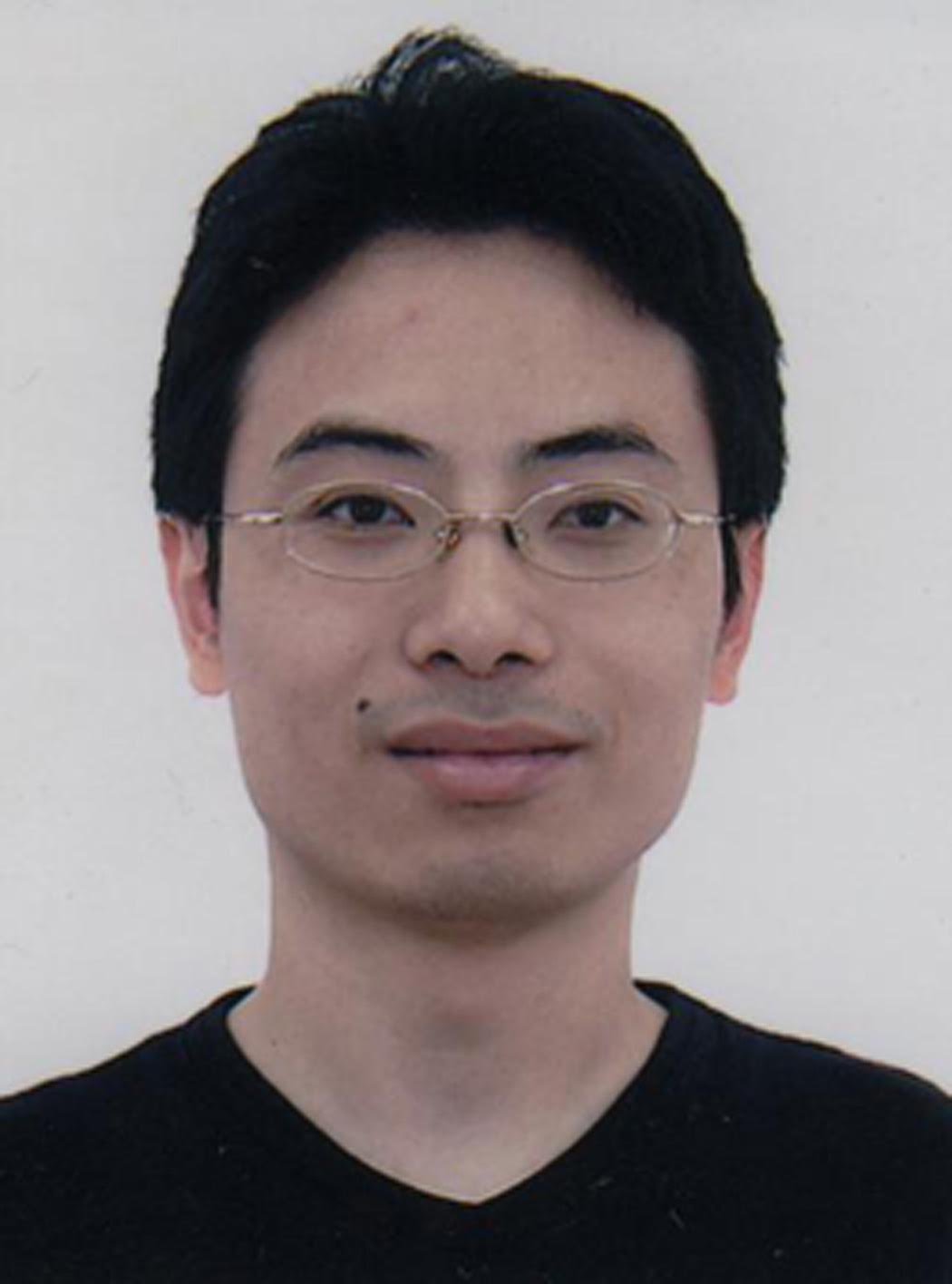}}]{Jianxin Wu}
(M'09) received the B.S. and M.S. degrees in computer science from Nanjing University, and the Ph.D. degree in computer
science from the Georgia Institute of Technology. He was an Assistant Professor with the Nanyang Technological University, 
Singapore. He is currently a Professor with the Department of Computer Science and Technology, Nanjing University, China,
and is associated with the National Key Laboratory for Novel Software Technology, China. His current research interests 
include computer vision and machine learning. He has served as an Area Chair for CVPR 2017 and ICCV 2015, a Senior PC Member 
for AAAI 2017 and AAAI 2016, and an Associate Editor of \emph{Pattern Recognition Journal}.
\end{IEEEbiography}
\begin{IEEEbiography}[{\includegraphics[width=1in,height=1.25in,clip,keepaspectratio]{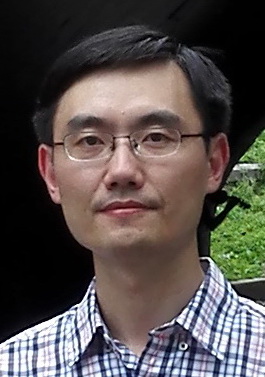}}]{Xin Geng}
(M'13) received the B.S. and M.S. degrees in computer science from Nanjing University, China, in 2001 
and 2004, respectively, and the Ph.D degree from Deakin University, Australia in 2008. He joined the School 
of Computer Science and Engineering at Southeast University, China, in 2008, and is currently a professor 
and vice dean of the school. He has authored over 50 refereed papers, and he holds five patents in these areas. 
His research interests include pattern recognition, machine learning, and computer vision.  
\end{IEEEbiography}
\vfill
\end{document}